\documentclass[10pt,twocolumn,letterpaper]{article}

\usepackage{times}
\usepackage{epsfig}
\usepackage{graphicx}
\usepackage{amsmath}
\usepackage{amssymb}
\usepackage{epstopdf}
\usepackage{dblfloatfix}
\usepackage{indentfirst}
\usepackage{mathptmx}
\usepackage{sectsty}

\usepackage{caption}
\usepackage{subcaption}

\usepackage[pagebackref=true,breaklinks=true,letterpaper=true,colorlinks,bookmarks=false]{hyperref}

\newsavebox\CBox
\def\textBF#1{\sbox\CBox{#1}\resizebox{\wd\CBox}{\ht\CBox}{\textbf{#1}}}

\newcommand{\eg}{{\emph{e.g.}}}

\newcommand{\etal}{{\emph{et al.}\ }}

\newcommand{\etc}{{\emph{etc}}}

\usepackage[top = 2.4cm, bottom = 2.8cm, left = 1.8cm, right = 2.3cm]{geometry}

\sectionfont{\fontsize{12}{15}\selectfont}

\begin{document}

\title{Rendering Natural Camera Bokeh Effect with Deep Learning\vspace{4mm}}

\author{Andrey Ignatov\\
{\tt\small andrey@vision.ee.ethz.ch}
\and
Jagruti Patel\\
{\tt\small patelj@student.ethz.ch}\vspace{3.2mm}
\\
ETH Zurich, Switzerland\\
\vspace{-1.8mm}
\and
Radu Timofte\\
{\tt\small timofter@vision.ee.ethz.ch}
}
\date{}
\maketitle
\thispagestyle{empty}

\begin{abstract}

\noindent\textit{Bokeh is an important artistic effect used to highlight the main object of interest on the photo by blurring all out-of-focus areas. While DSLR and system camera lenses can render this effect naturally, mobile cameras are unable to produce shallow depth-of-field photos due to a very small aperture diameter of their optics. Unlike the current solutions simulating bokeh by applying Gaussian blur to image background, in this paper we propose to learn a realistic shallow focus technique directly from the photos produced by DSLR cameras. For this, we present a large-scale bokeh dataset consisting of 5K shallow / wide depth-of-field image pairs captured using the Canon 7D DSLR with 50mm f/1.8 lenses. We use these images to train a deep learning model to reproduce a natural bokeh effect based on a single narrow-aperture image. The experimental results show that the proposed approach is able to render a plausible non-uniform bokeh even in case of complex input data with multiple objects. The dataset, pre-trained models and codes used in this paper are available on the project website\footnote{\,\,\url{https://people.ee.ethz.ch/~ihnatova/pynet-bokeh.html}}.}

\end{abstract}

\section{Introduction}

\begin{figure}[t!]
\centering
\includegraphics[width=0.85\linewidth]{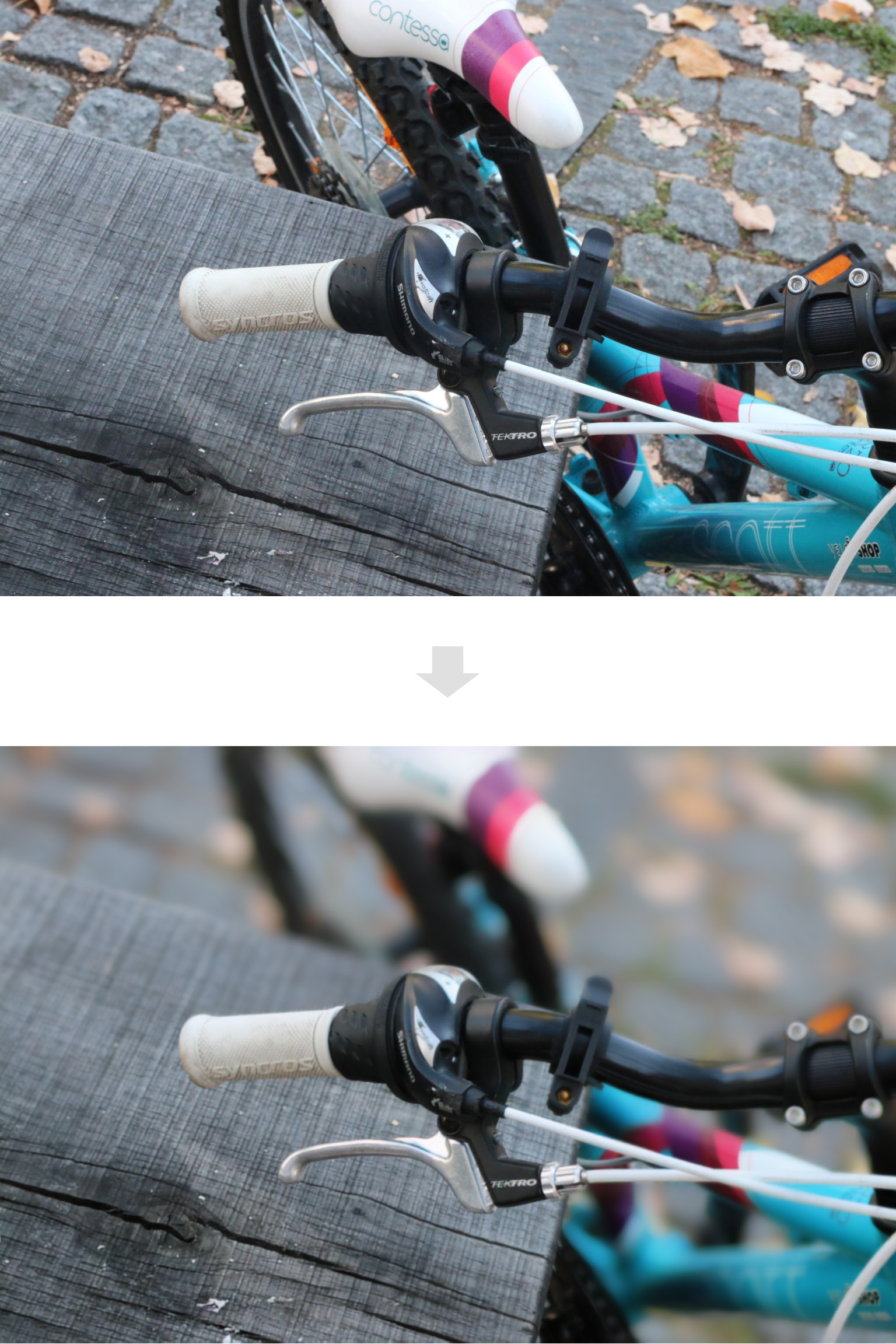}
\vspace{3.4mm}
\caption{The original shallow depth-of-field image and the image produced with our method.}
\label{fig:Examples}
\vspace{-3.7mm}
\end{figure}

Bokeh effect is a very popular photography technique used to make the subject in the shot stand out sharply against a blurred background (Fig.~\ref{fig:Examples}). It is achieved by focusing the camera on the selected area or object and shooting the photo with a wide aperture lens. This produces a shallow depth-of-field image where only objects located within a narrow image plane are visible clearly, while all other parts of the image are blurred. This effect is often leading to very pleasing visual results, and besides that allows to wash out unnecessary, distracting or unattractive background details, which is especially useful in case of mobile photography. In order to get good bokeh, fast lenses with a large aperture are needed, which makes this effect unattainable for mobile cameras with compact optics and tiny sensors. As a result, bokeh effect can only be simulated computationally on smartphones and other devices with small mobile cameras.

\begin{figure*}[t!]
\centering
\setlength{\tabcolsep}{1pt}
\resizebox{\linewidth}{!}
{
\begin{tabular}{cccccc}
Wide Depth-of-field & Shallow Depth-of-field & Wide Depth-of-field & Shallow Depth-of-field & Wide Depth-of-field & Shallow Depth-of-field\\
    \includegraphics[width=0.24\linewidth]{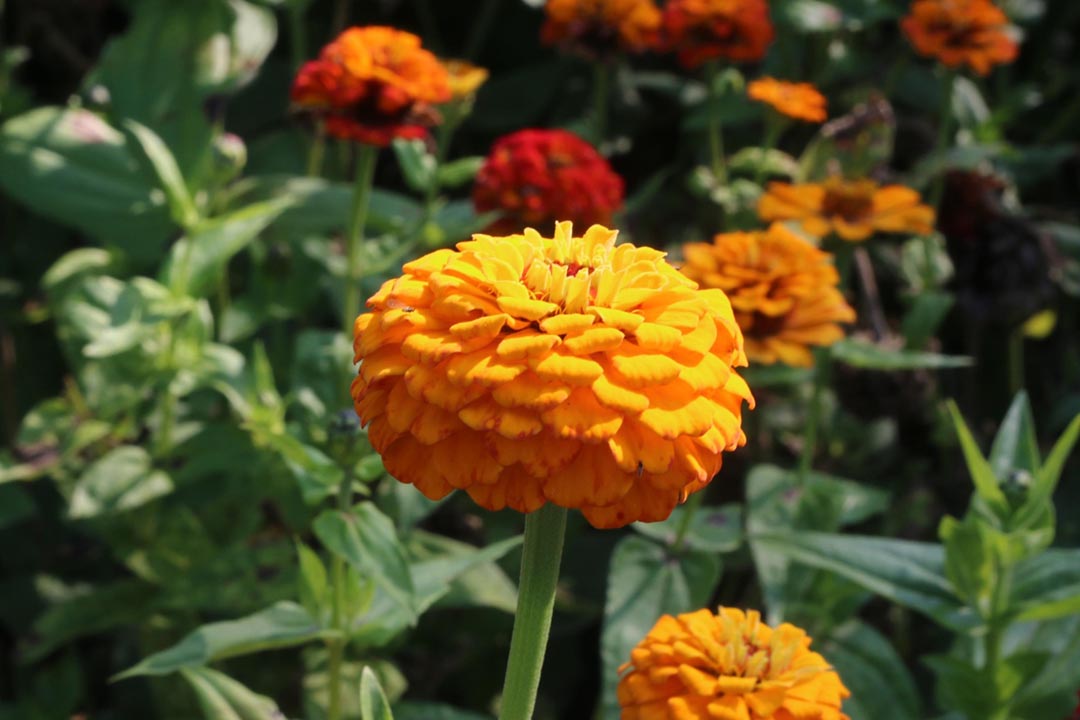}&
    \includegraphics[width=0.24\linewidth]{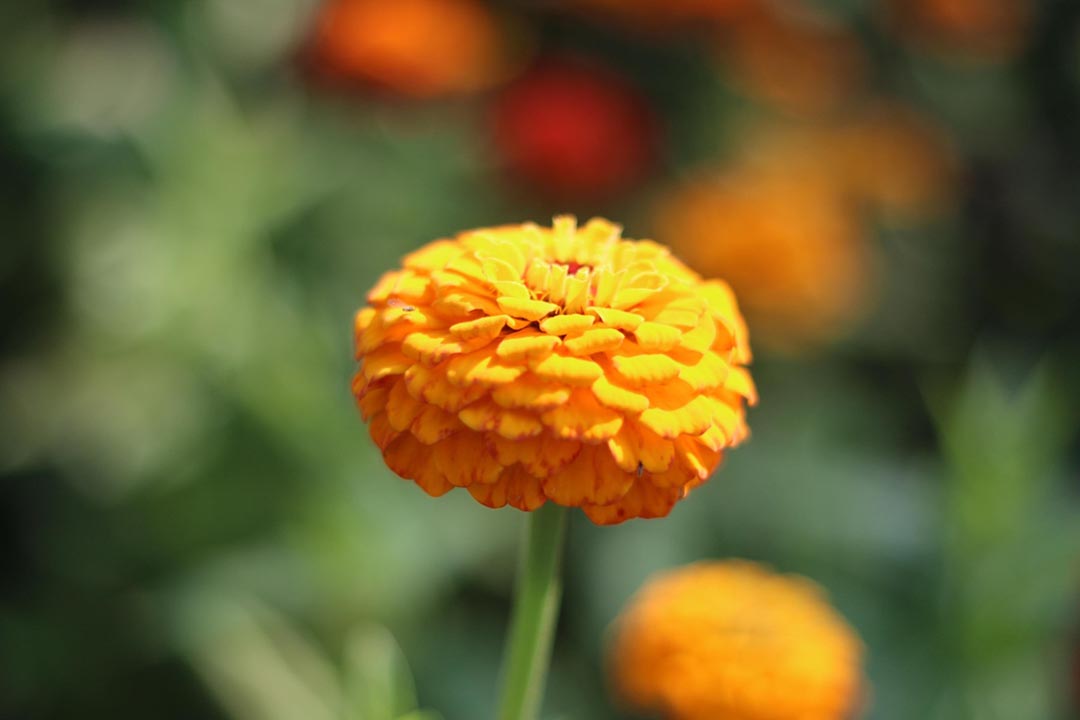}& \hspace{0.8mm}
    \includegraphics[width=0.24\linewidth]{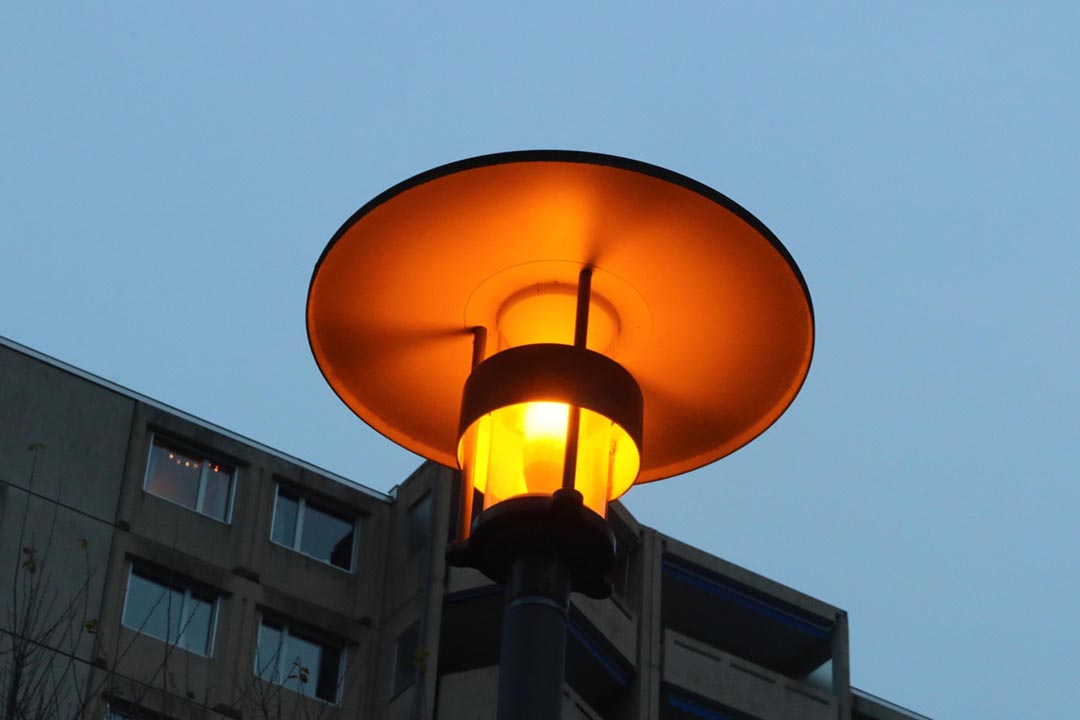}&
    \includegraphics[width=0.24\linewidth]{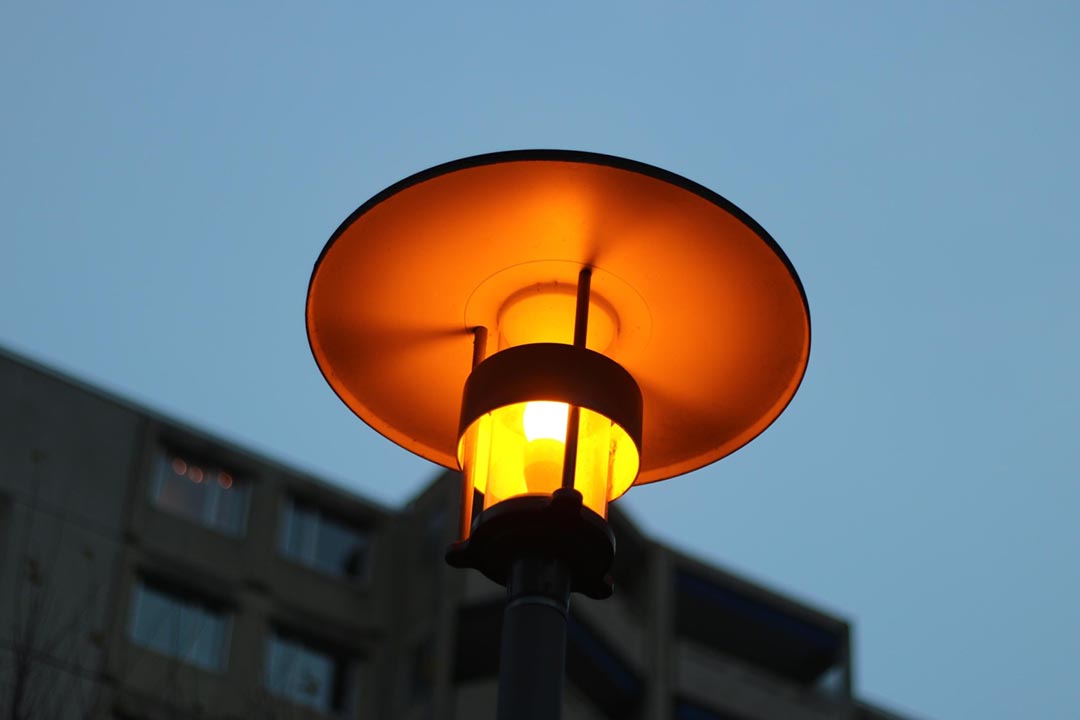}& \hspace{0.8mm}
    \includegraphics[width=0.24\linewidth]{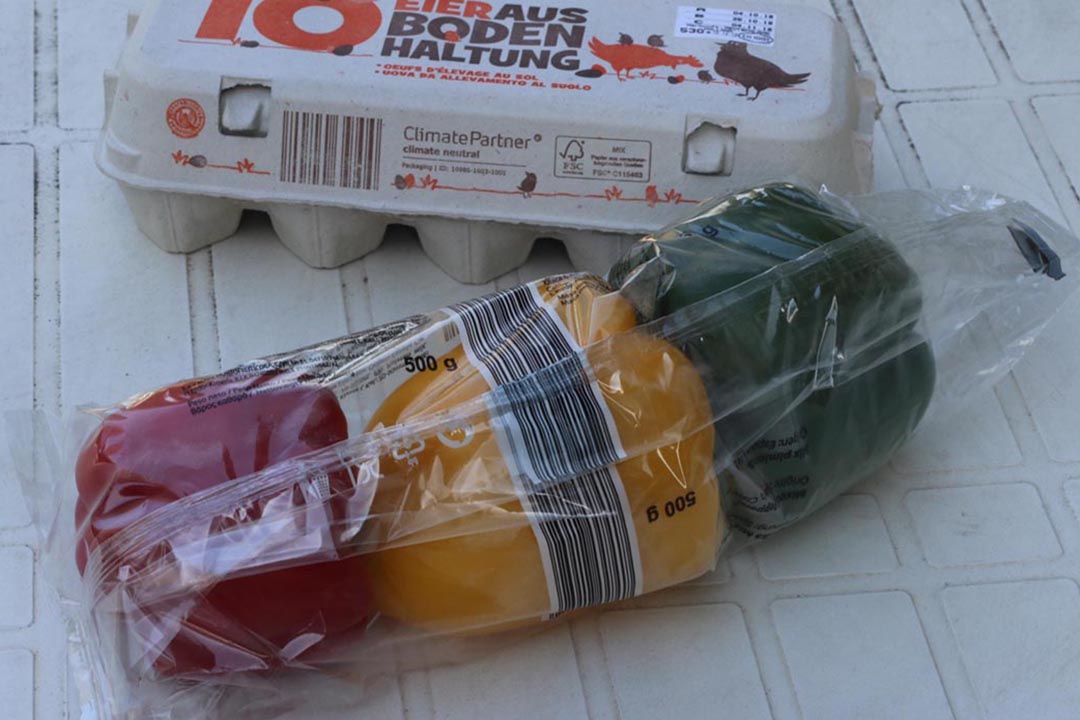}&
    \includegraphics[width=0.24\linewidth]{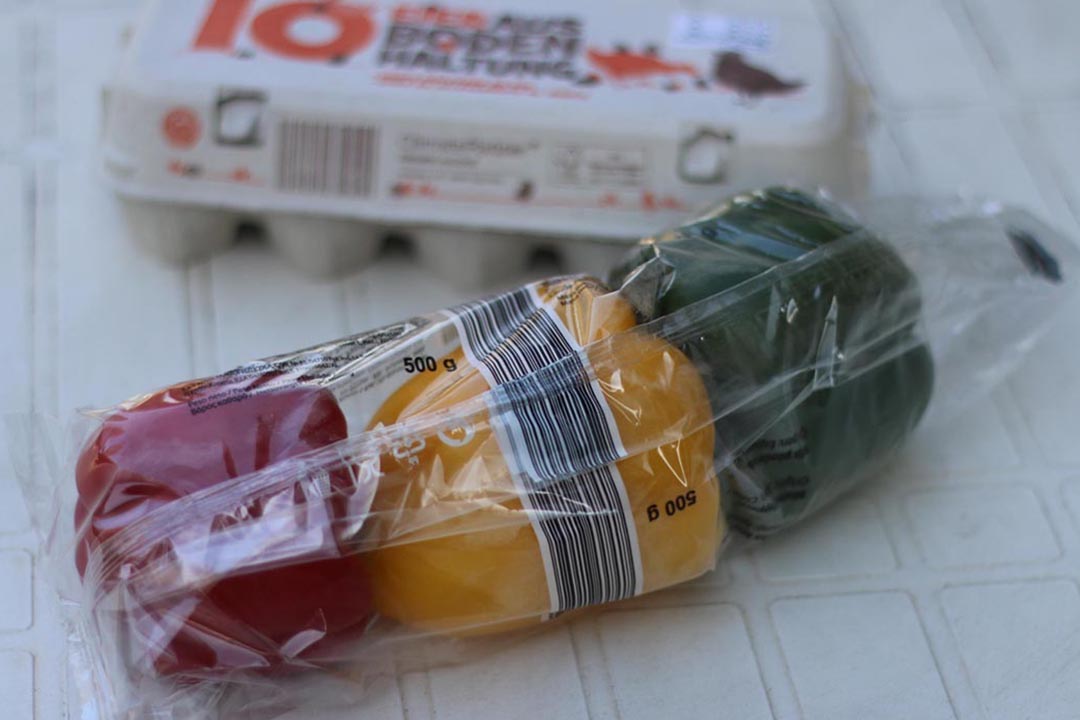} \\
    \includegraphics[width=0.24\linewidth]{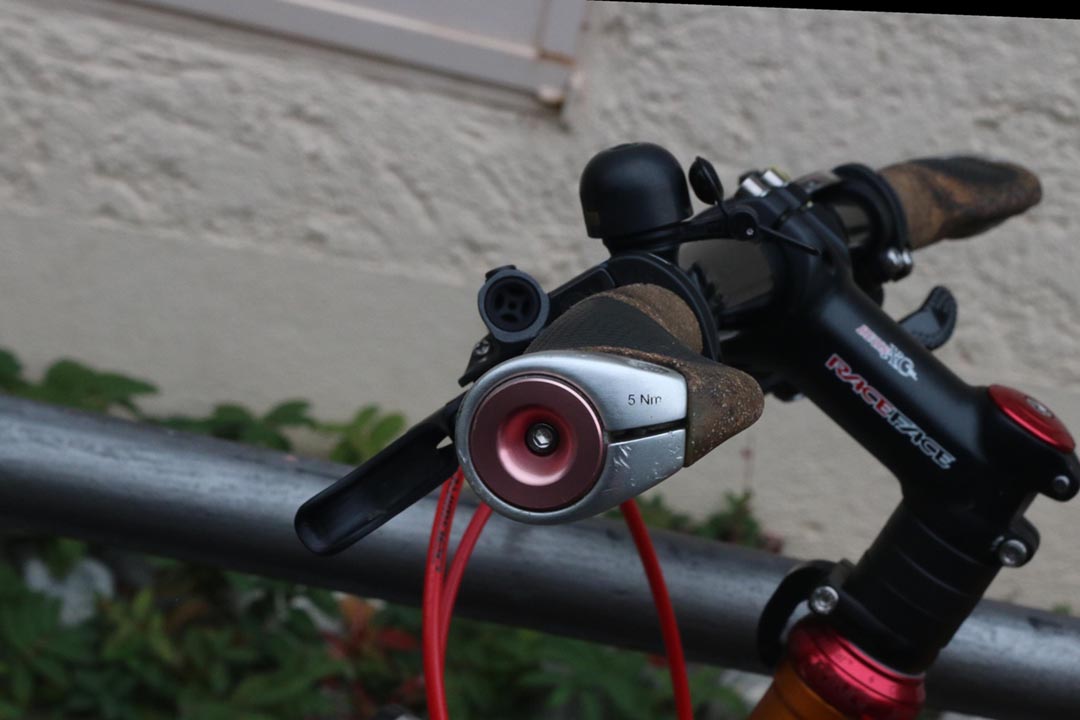}&
    \includegraphics[width=0.24\linewidth]{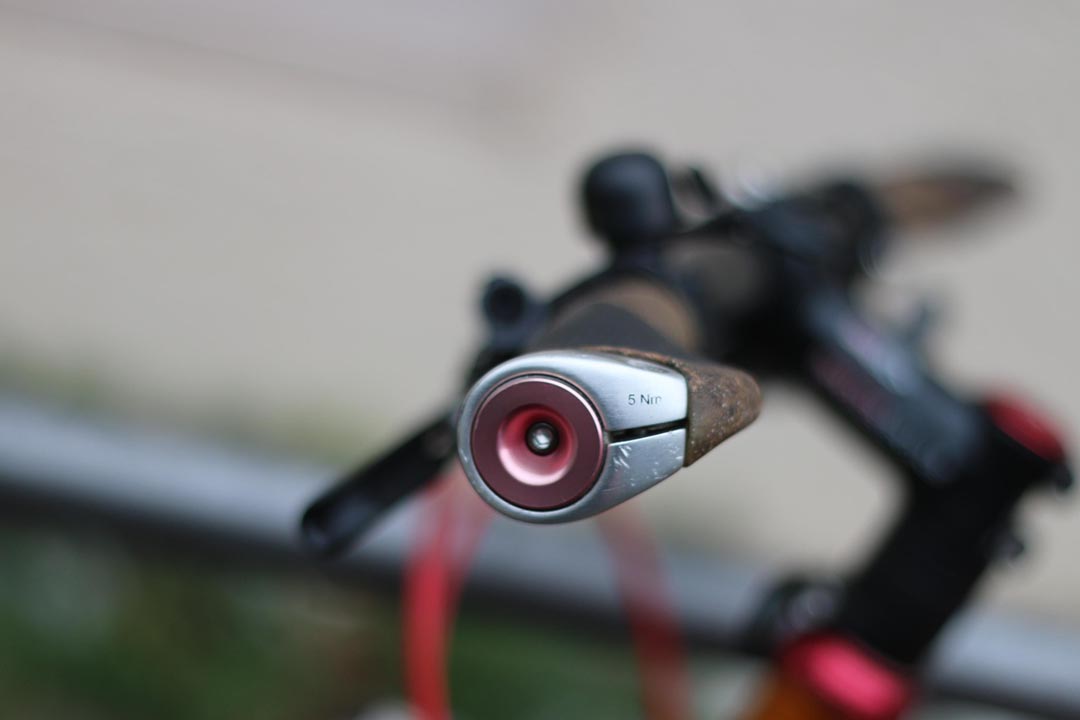}& \hspace{0.8mm}
    \includegraphics[width=0.24\linewidth]{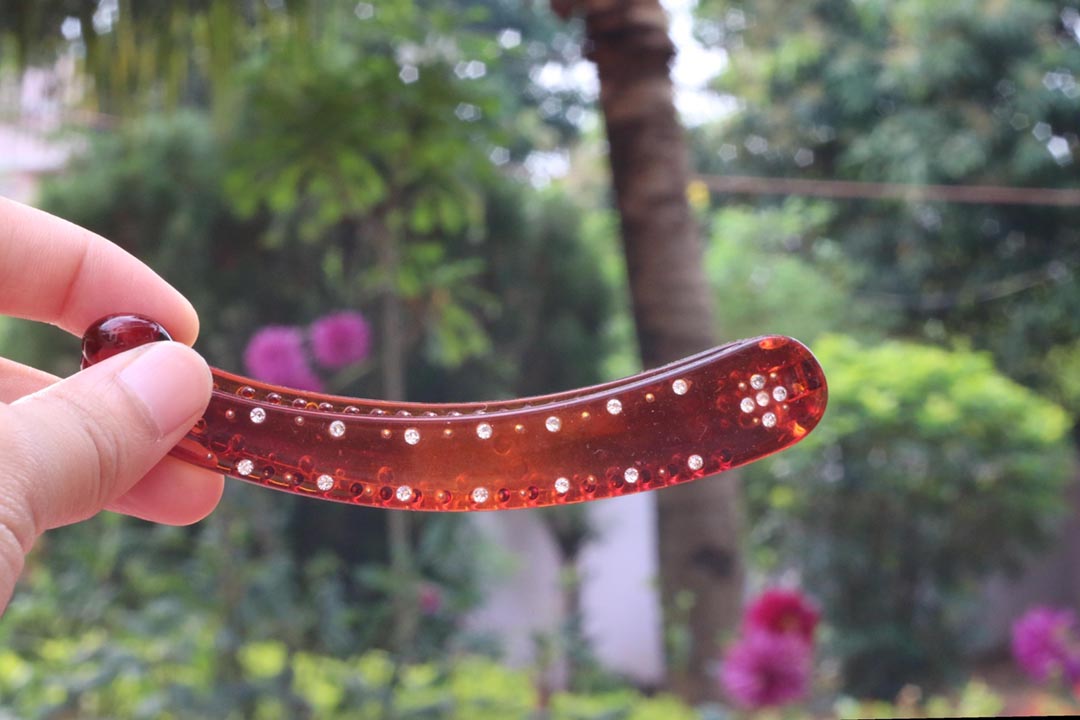}&
    \includegraphics[width=0.24\linewidth]{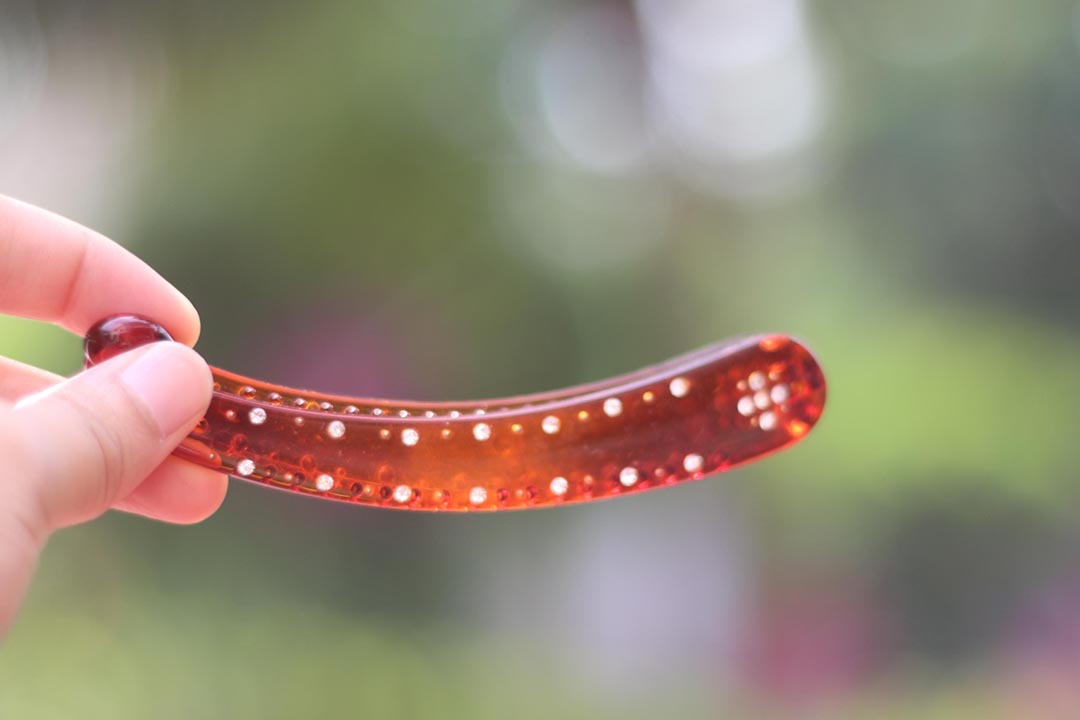}& \hspace{0.8mm}
    \includegraphics[width=0.24\linewidth]{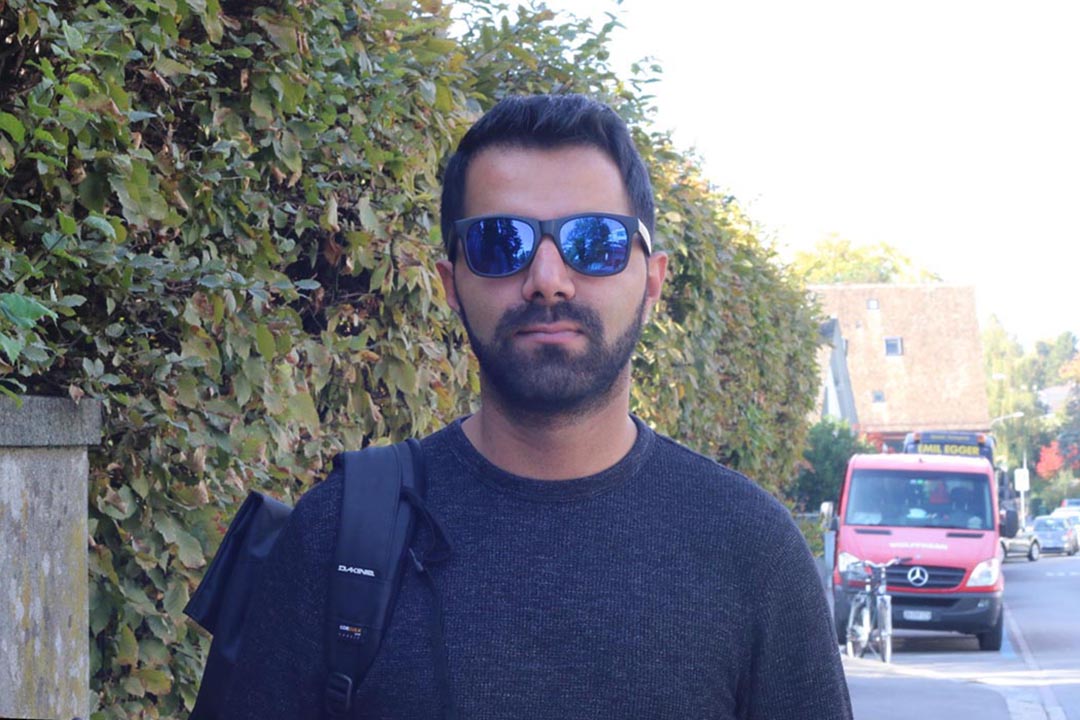}&
    \includegraphics[width=0.24\linewidth]{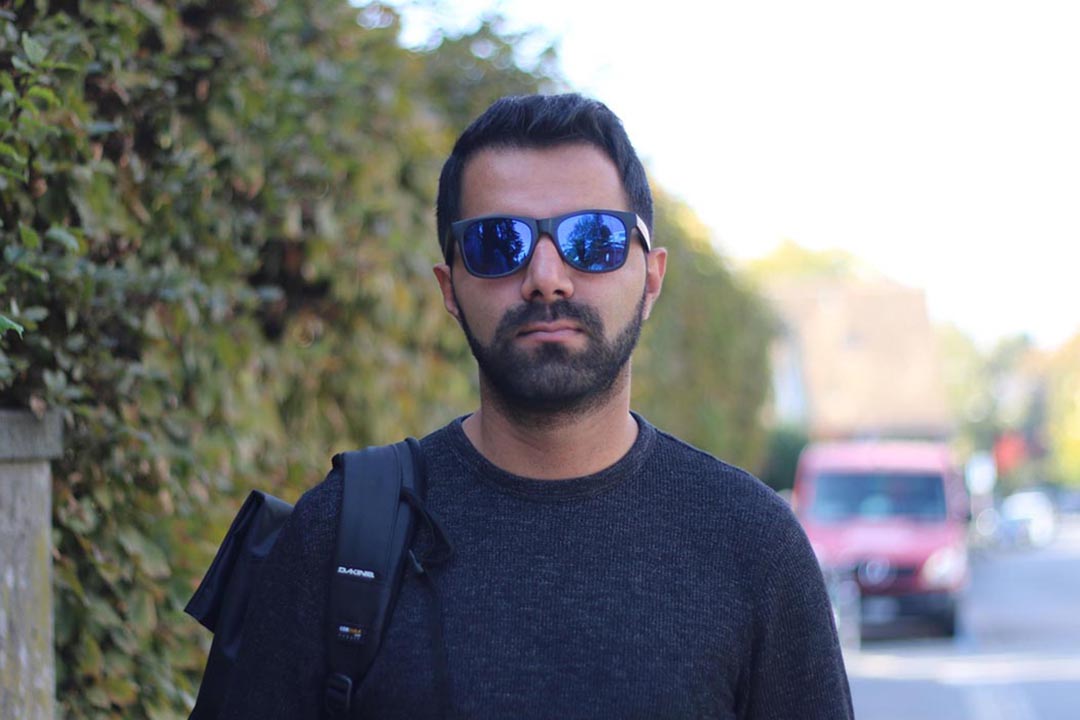} \\
    \includegraphics[width=0.24\linewidth]{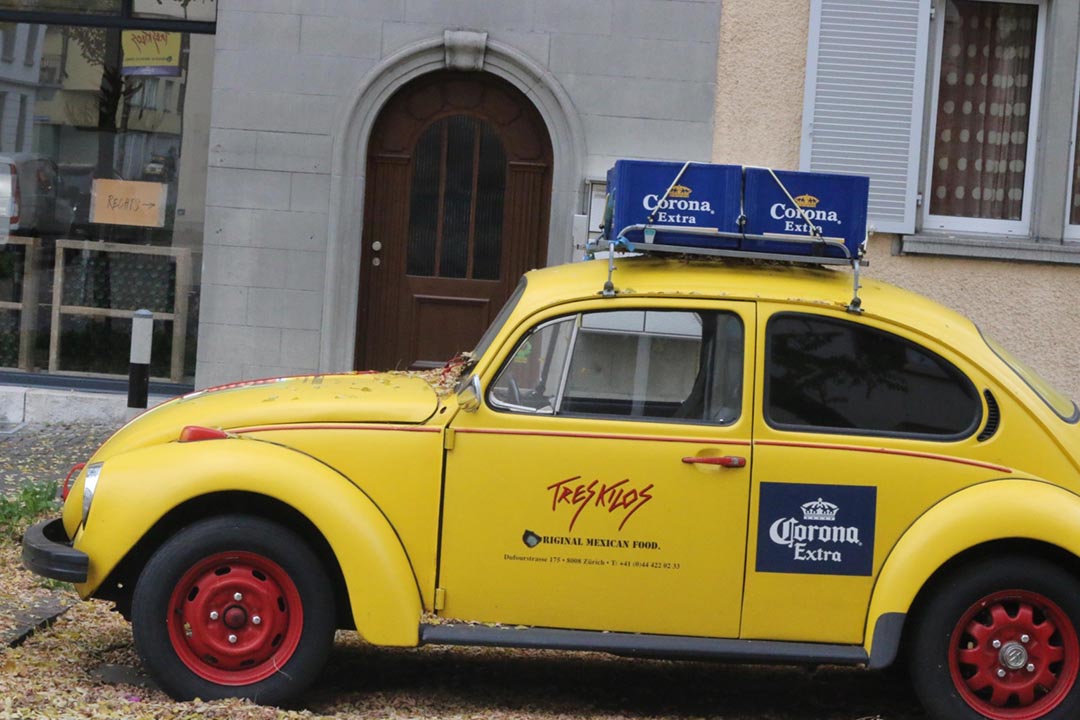}&
    \includegraphics[width=0.24\linewidth]{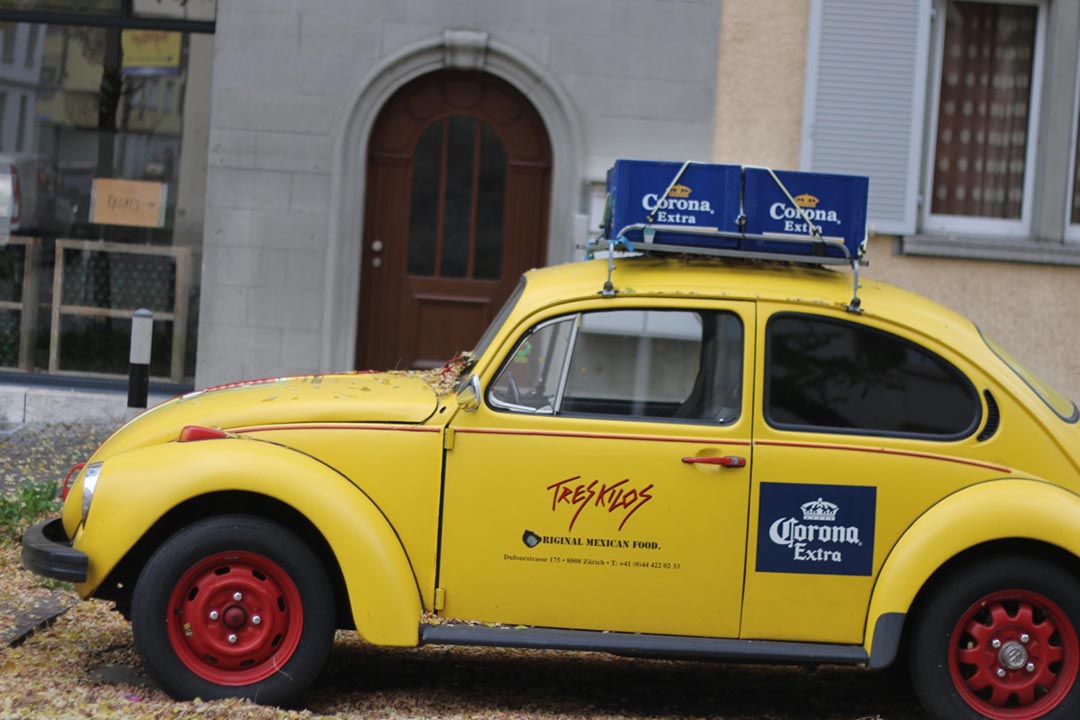}& \hspace{0.8mm}
    \includegraphics[width=0.24\linewidth]{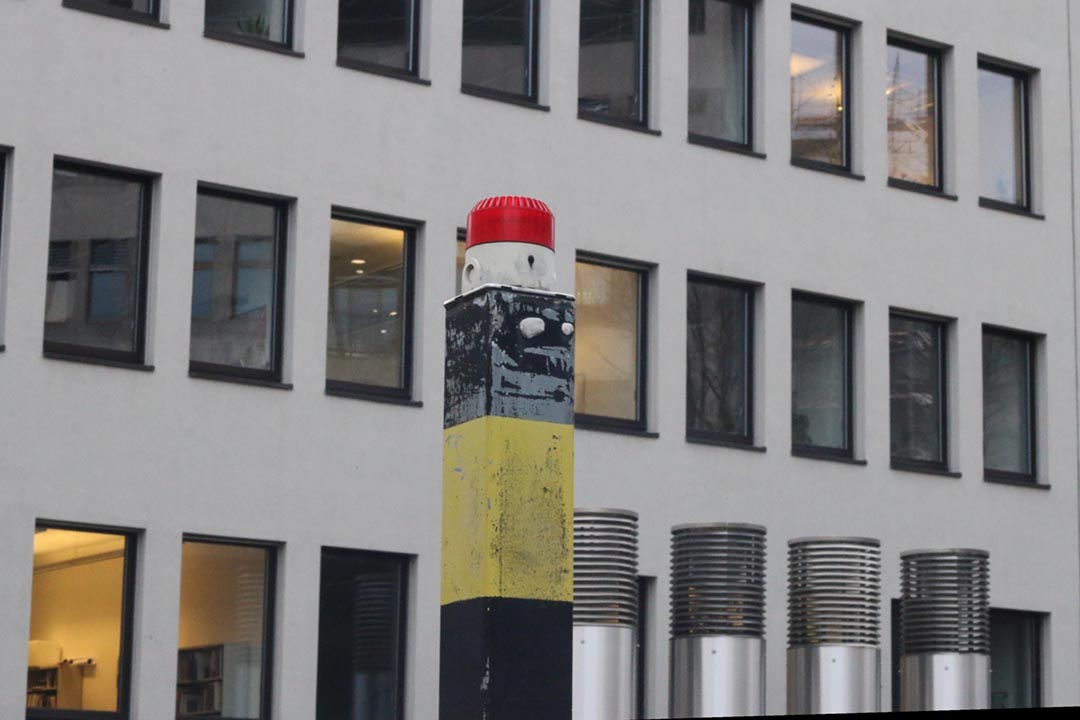}&
    \includegraphics[width=0.24\linewidth]{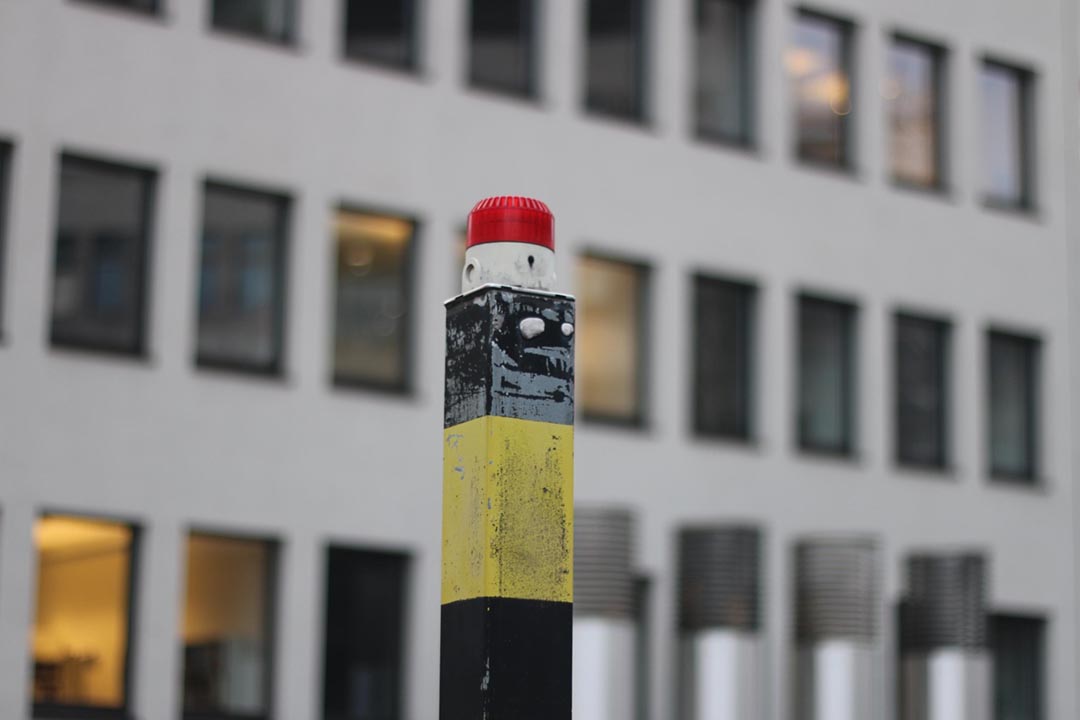}& \hspace{0.8mm}
    \includegraphics[width=0.24\linewidth]{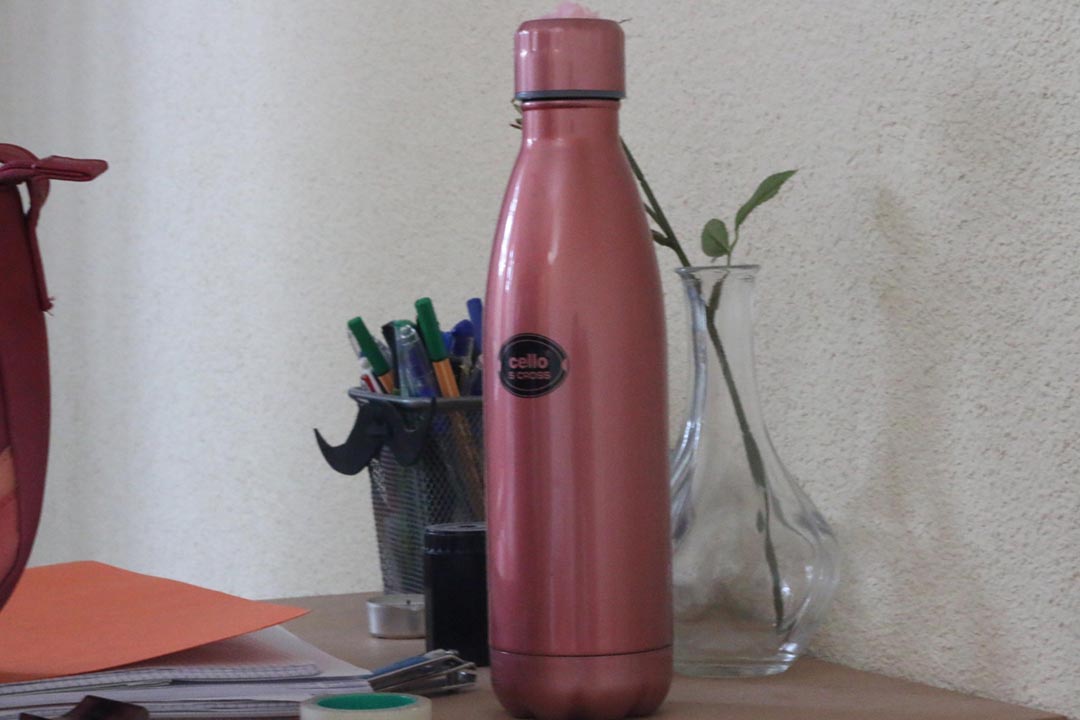}&
    \includegraphics[width=0.24\linewidth]{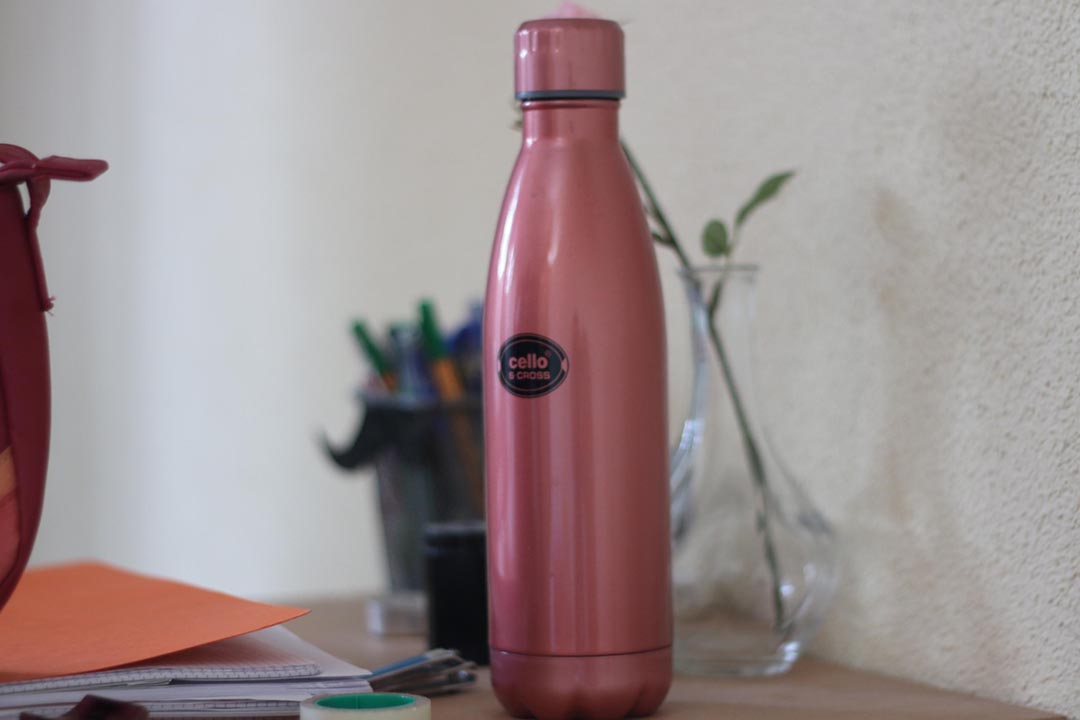} \\
    \includegraphics[width=0.24\linewidth]{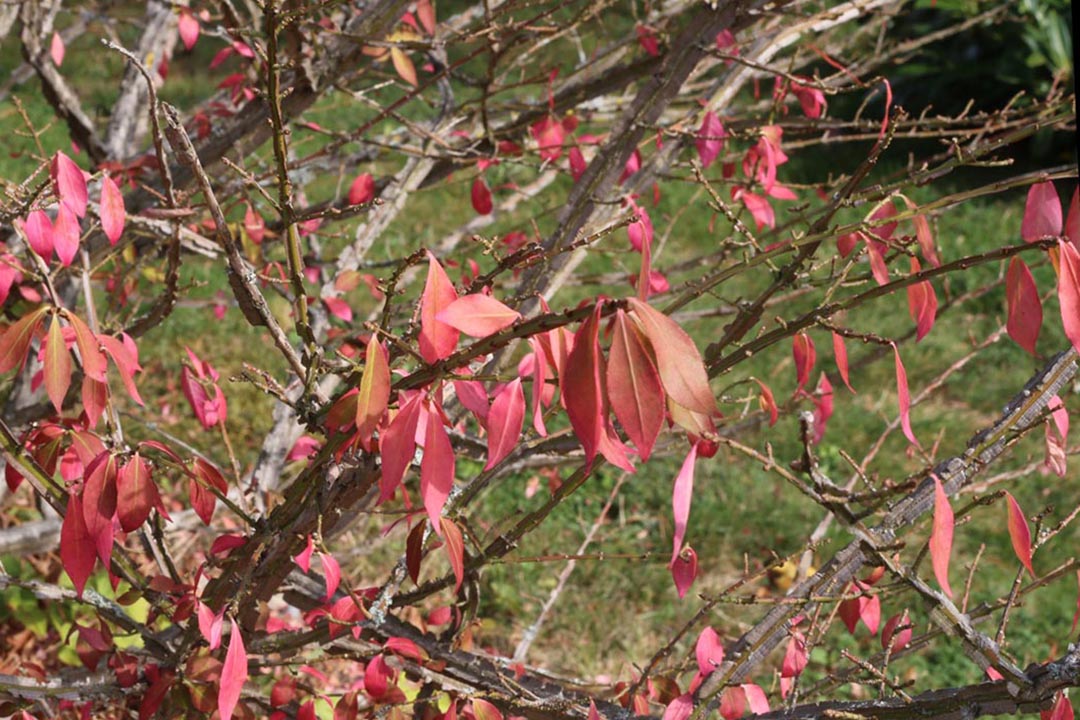}&
    \includegraphics[width=0.24\linewidth]{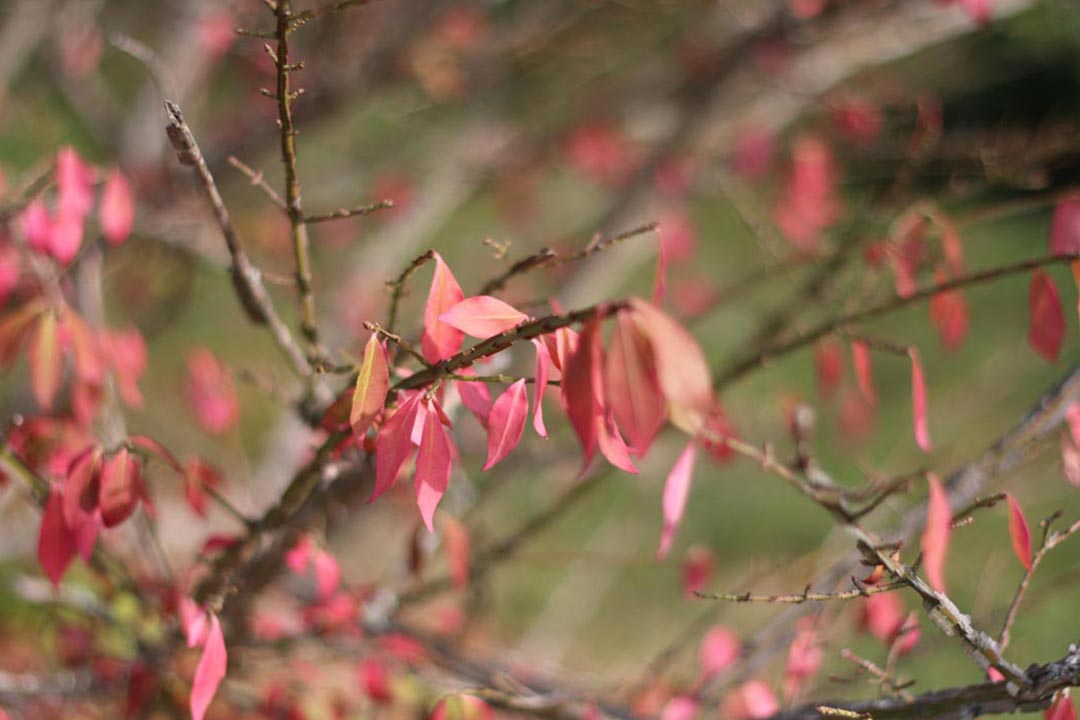}& \hspace{0.8mm}
    \includegraphics[width=0.24\linewidth]{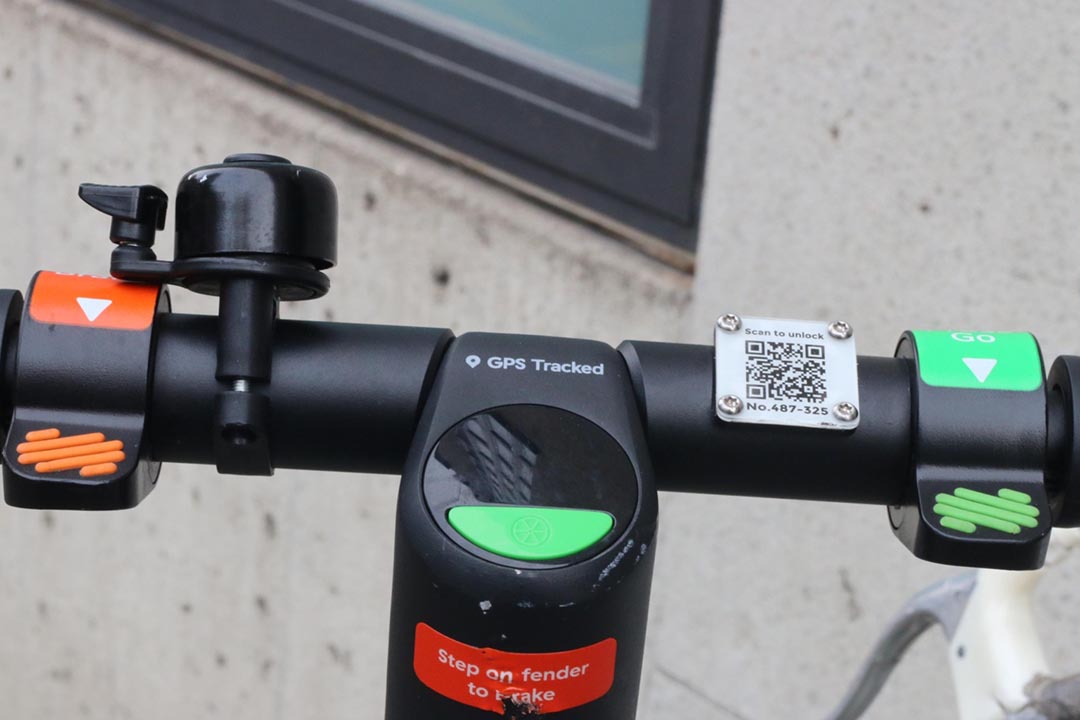}&
    \includegraphics[width=0.24\linewidth]{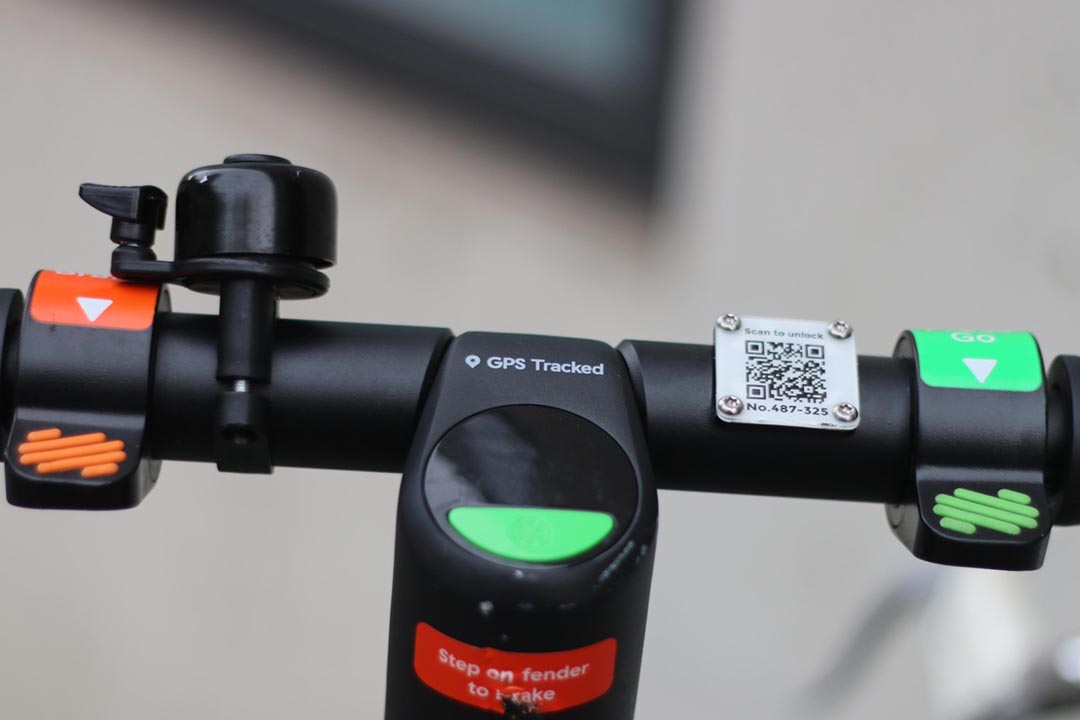}& \hspace{0.8mm}
    \includegraphics[width=0.24\linewidth]{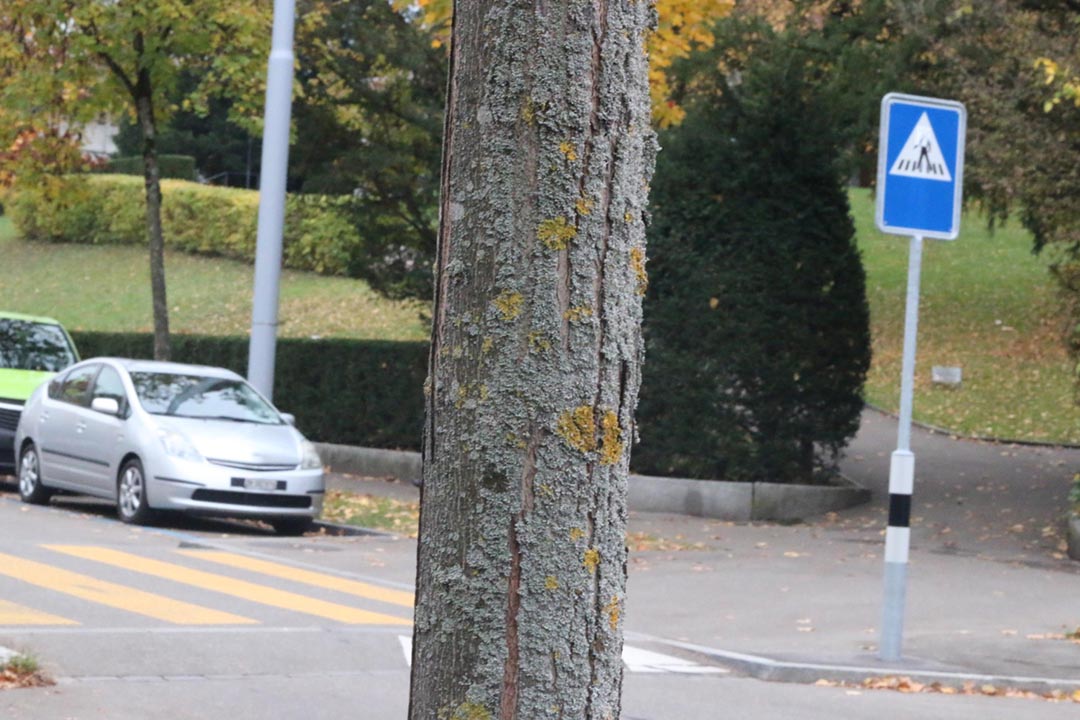}&
    \includegraphics[width=0.24\linewidth]{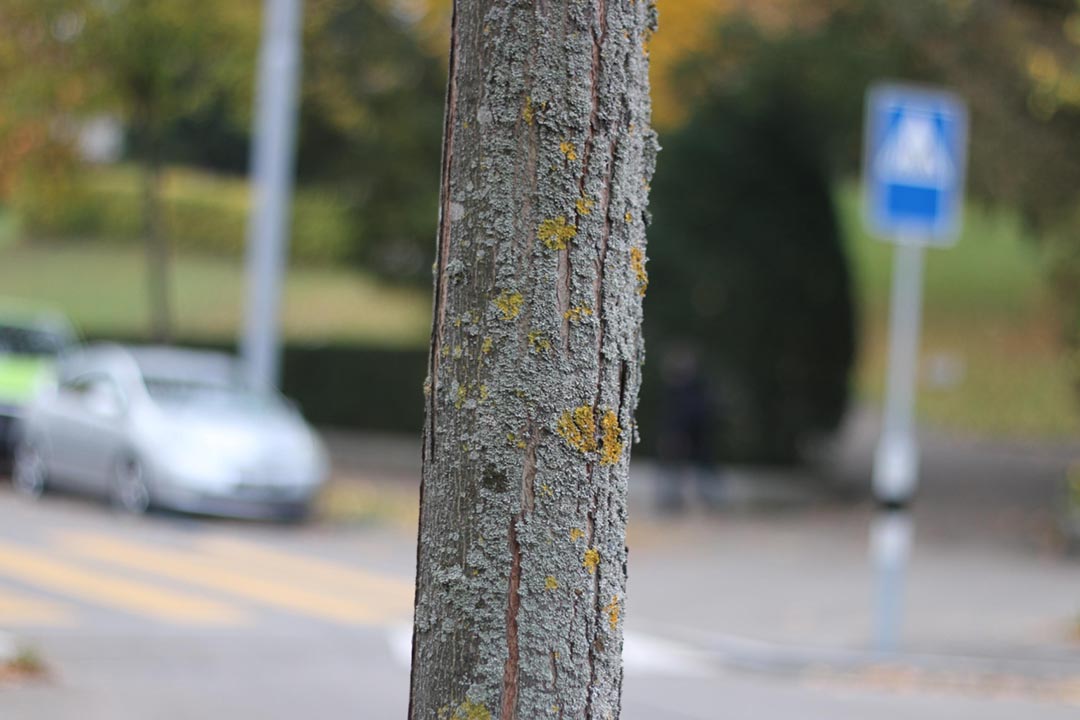} \\
    \includegraphics[width=0.24\linewidth]{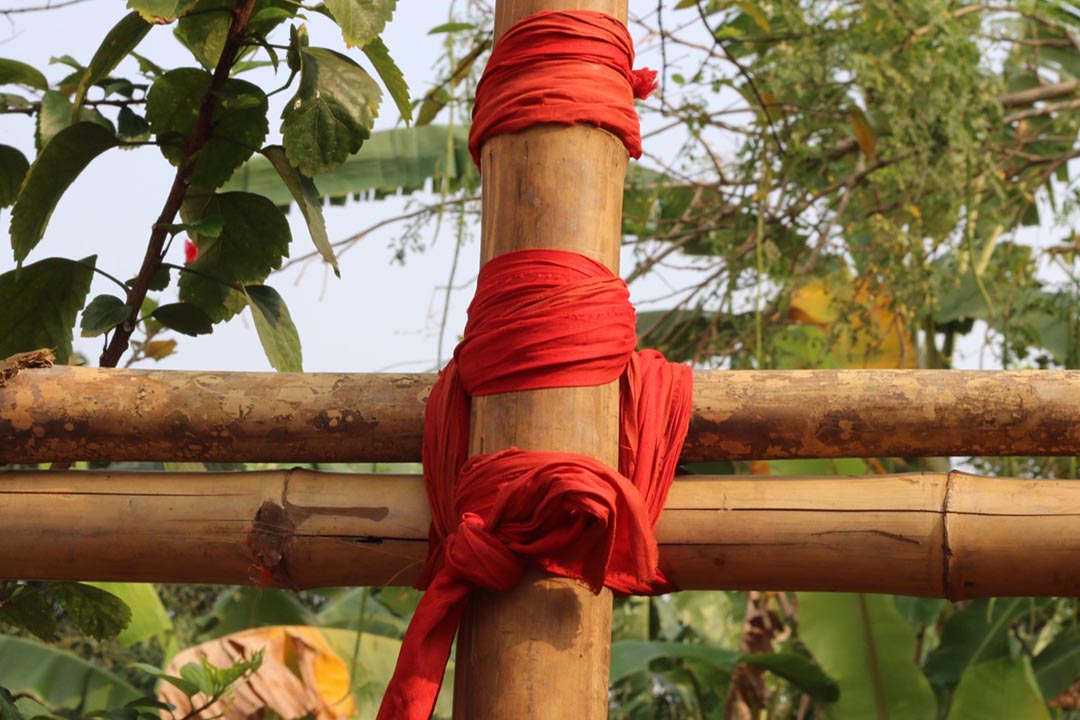}&
    \includegraphics[width=0.24\linewidth]{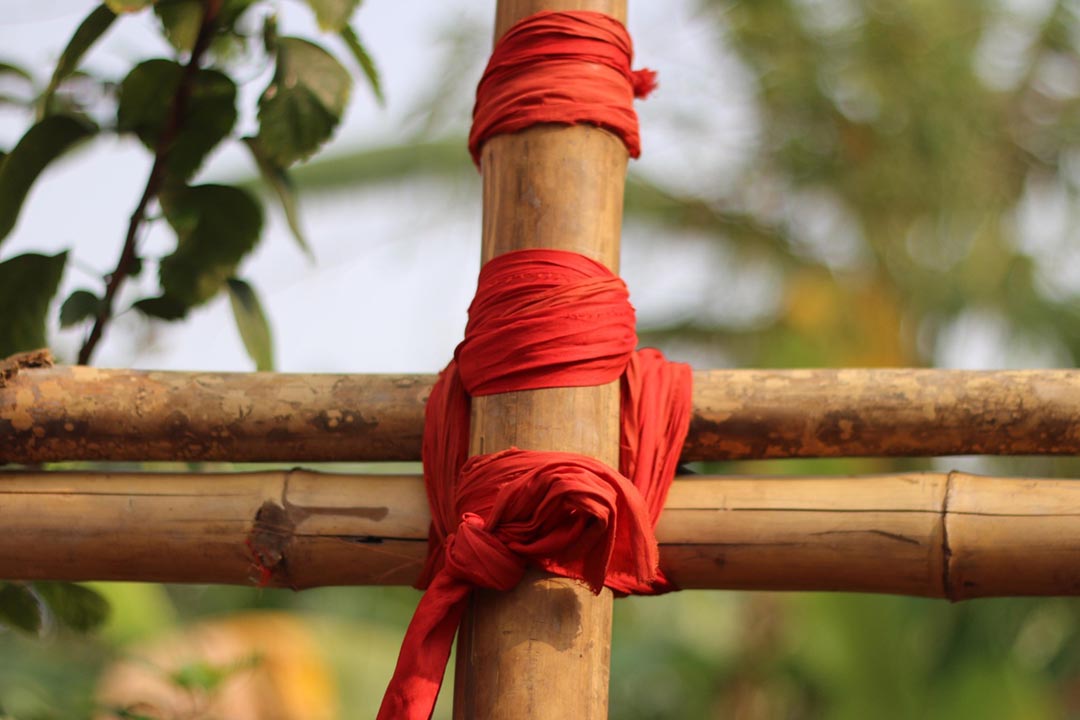}& \hspace{0.8mm}
    \includegraphics[width=0.24\linewidth]{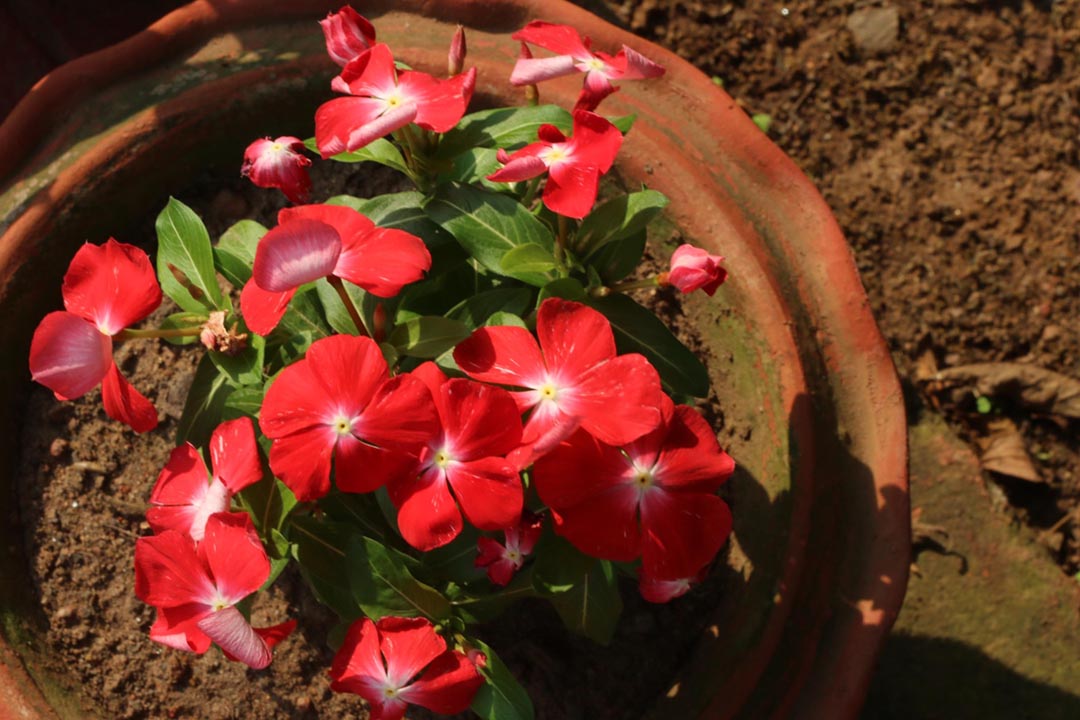}&
    \includegraphics[width=0.24\linewidth]{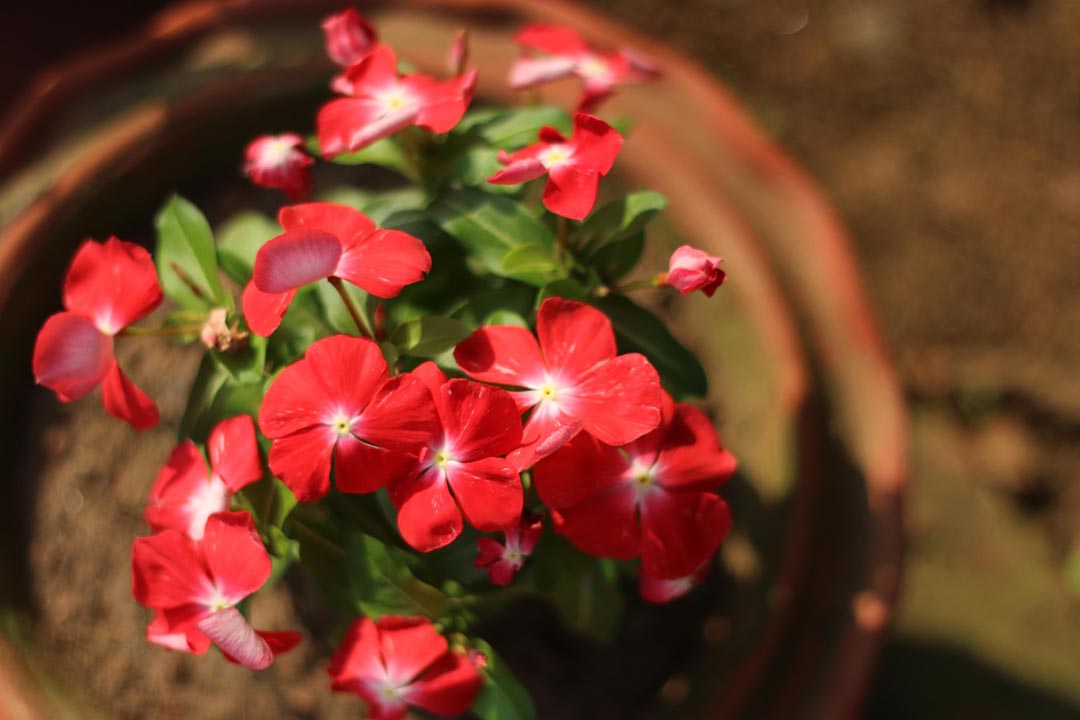}& \hspace{0.8mm}
    \includegraphics[width=0.24\linewidth]{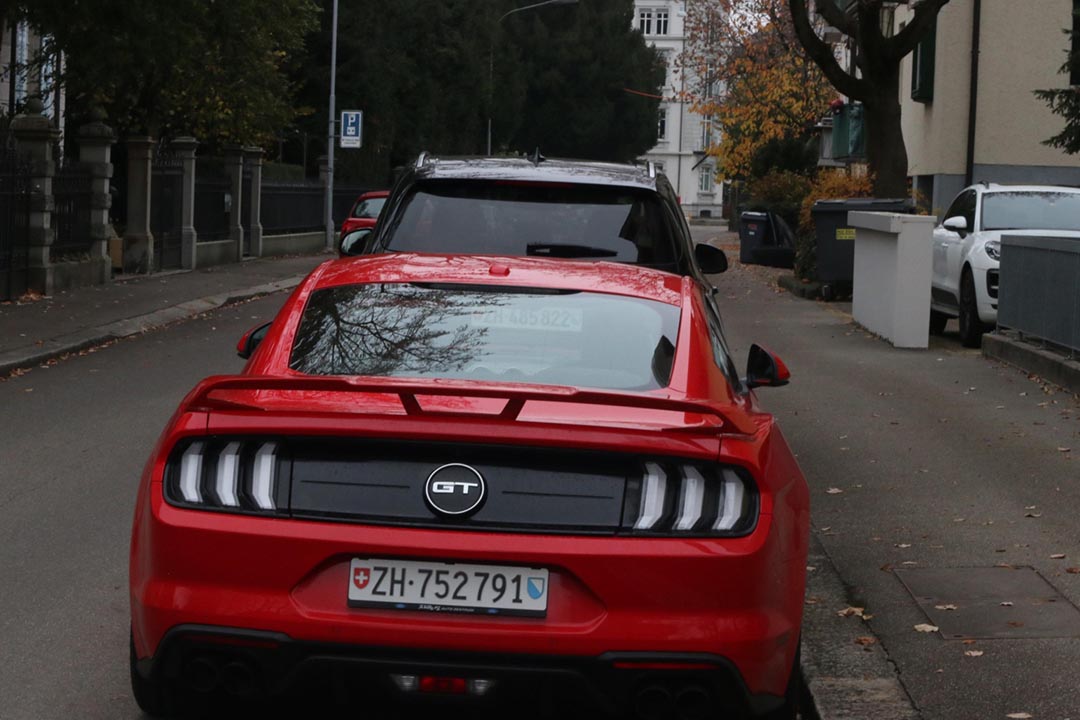}&
    \includegraphics[width=0.24\linewidth]{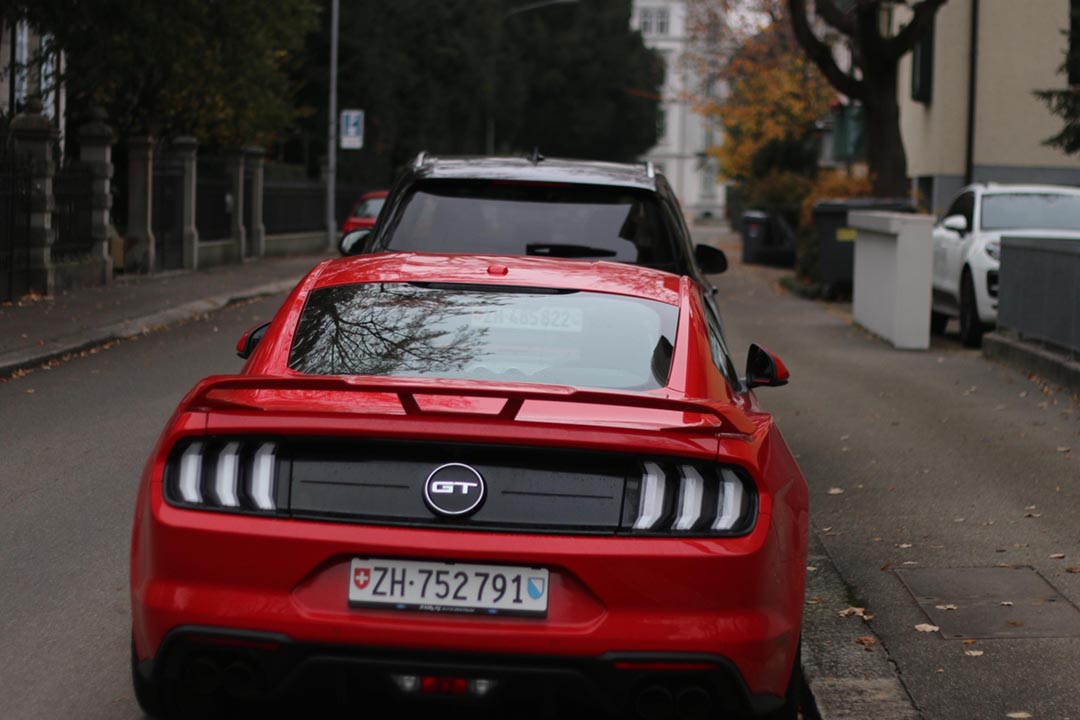} \\
\end{tabular}
}
\vspace{1.2mm}
\caption{Sample wide and shallow depth-of-field image pairs from the EBB! dataset.}
\label{fig:sample_images}
\vspace{-1.2mm}
\end{figure*}

Synthetic bokeh effect rendering is a relatively new machine learning topic that has emerged in the past years. \cite{shen2016automatic,shen2016deep} were one of the first papers exploring shallow depth-of-field simulation based on a single image. These works were focused on portrait photos only: they first segmented out people from the image with a convolutional neural network, and then uniformly blurred the remaining image background. The same approach was also considered in~\cite{zhu2017fast}, where the authors proposed an efficient solution for person segmentation. There also exists a separate group of works~\cite{Hernandez2014Lens,ha2016high,yu20143d} that are blurring the image based on the predicted depth map, however they require the camera to move in order to get the parallax effect, thus the practical applicability of these approaches is quite limited. An accurate depth map estimation can be also obtained with stereo vision~\cite{barron2015fast} using synchronously captured photos from multiple cameras.

Since 2015, the majority of flagship and mid-range mobile devices are getting special hardware for accelerating deep learning models~\cite{ignatov2018ai}. The performance of such hardware is already approaching the results of mid-range Nvidia GPUs presented in the past years~\cite{ignatov2019ai}, making the use of deep learning-based methods for bokeh effect simulation on mobile devices especially relevant. Paper~\cite{wadhwa2018synthetic} describes the synthetic depth-of-field rendering approach used on Google Pixel devices that is able to achieve very impressive results for the majority of photo scenes. In this work, the authors are predicting a depth map based on a single image and are using camera's dual-pixel auto-focus system when possible for refinement of the results. The weakest part of the paper is the actual bokeh rendering method~-- a simple approximated disk blur is used for this, making the visual results quite different from the natural bokeh photos produced by DSLR cameras.

In this paper, we present a different approach to the considered problem. Since the properties of the real bokeh depend on a large number of parameters such as focal length, distance to the subject, the acceptable circle of confusion, the aperture size, type and construction, various optical lens characteristics, \etc, it is almost impossible to render this effect precisely with the existing software tools and methods. Therefore, in this work we propose to learn the bokeh technique directly from the photos produced by a high-end DSLR camera using deep learning. The presented approach is camera independent, does not require any special hardware, and can also be applied to the existing images.

\bigskip

\textBF{Our main contributions are:}

\vspace{-0.8mm}
\begin{itemize}
\setlength\itemsep{-0.2mm}
\item A novel end-to-end deep learning solution for realistic bokeh effect rendering. The model is trained to map the standard  narrow-aperture images into shallow depth-of-field photos captured with a DSLR camera.
\item A large-scale dataset containing 5K shallow / wide depth-of-field image pairs collected in the wild with the Canon 7D DSLR camera and 50mm f/1.8 fast lens.
\item A comprehensive set of experiments evaluating the quantitative and perceptual quality of the rendered bokeh photos. We demonstrate that the proposed solution can process one 1024$\times$1536 px image under 5s on all high-end mobile chipsets when running it on GPU.
\end{itemize}
\vspace{-0.8mm}

\section{\textit{Everything is Better with Bokeh!} Dataset}

One of the biggest challenges in the bokeh rendering task is to get high-quality real data that can be used for training deep models. To tackle this problem, a large-scale \textit{Everything is Better with Bokeh!} (EBB!) dataset containing more than 10 thousand images was collected in the wild with the Canon 7D DSLR camera during several months. By controlling the aperture size of the lens, images with shallow and wide depth-of-field were taken. In each photo pair, the first image was captured with a narrow aperture (f/16) that results in a normal sharp photo, whereas the second one was shot using the highest aperture (f/1.8) leading to a strong bokeh effect. The photos were taken during the daytime in a wide variety of places and in various illumination and weather conditions. The photos were captured in automatic mode, the default settings were used throughout the entire collection procedure. An example set of collected images is presented in Figure~\ref{fig:sample_images}.

The captured image pairs are not aligned exactly, therefore they were first matched using SIFT keypoints and RANSAC method same as in~\cite{ignatov2017dslr}. The resulting images were then cropped to their intersection part and downscaled so that their final height is equal to 1024 pixels. Finally, we computed a coarse depth map for each wide depth-of-field image using the Megadepth model proposed in~\cite{li2018megadepth}. These maps can be stacked directly with the input images and used as an additional guidance for the trained model. From the resulting 10 thousand images, 200 image pairs were reserved for testing, while the other 4.8 thousand photo pairs can be used for training and validation.

\vspace{-2mm}
\section{Proposed Method}

\begin{figure}[t!]
\centering
\includegraphics[width=1.0\linewidth]{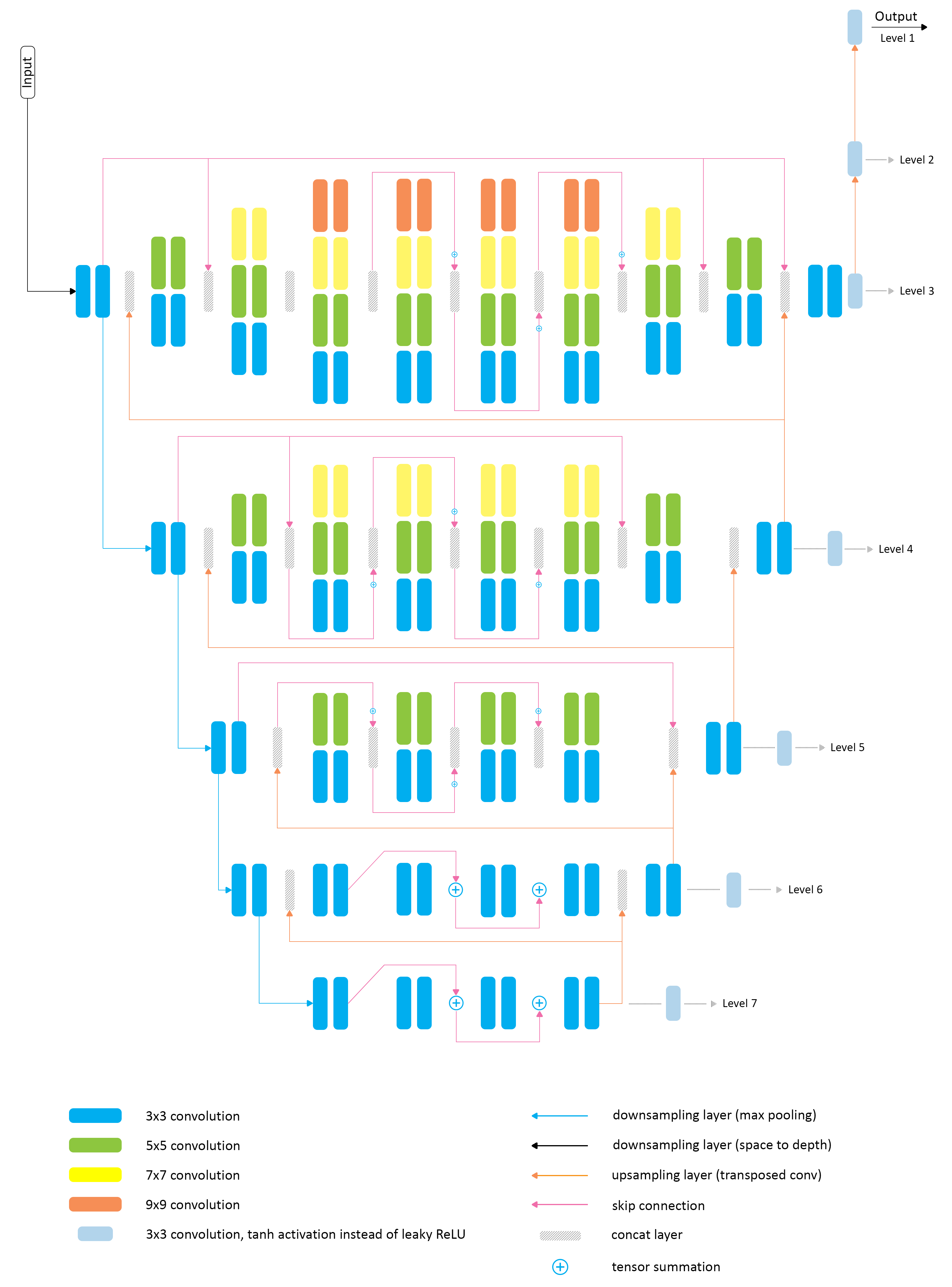}
\vspace{1.4mm}
\caption{\small{The architecture of the PyNET-based model. Concat and Sum ops are applied to the outputs of the adjacent layers.}}
\label{fig:Architecture}
\vspace{-2.4mm}
\end{figure}

Bokeh effect simulation problem belongs to a group of tasks dealing with both global and local image processing. High-level image analysis is needed here to detect the areas on the photo where the bokeh effect should be applied, whereas low-level processing is used for rendering the actual shallow depth-of-field images and refining the results. Therefore, in this work we base our solution on the PyNET architecture~\cite{ignatov2020replacing} designed specifically for this kind of tasks: it is processing the image at different scales and combining the learned global and local features together.

\begin{figure*}[t!]
\centering
\setlength{\tabcolsep}{1pt}
\resizebox{0.96\linewidth}{!}
{
\begin{tabular}{ccc}
\scriptsize{Original Image}\normalsize & \scriptsize{Bokeh image rendered with PyNET}\normalsize & \scriptsize{Canon 7D Photo}\normalsize\\
    \includegraphics[width=0.24\linewidth]{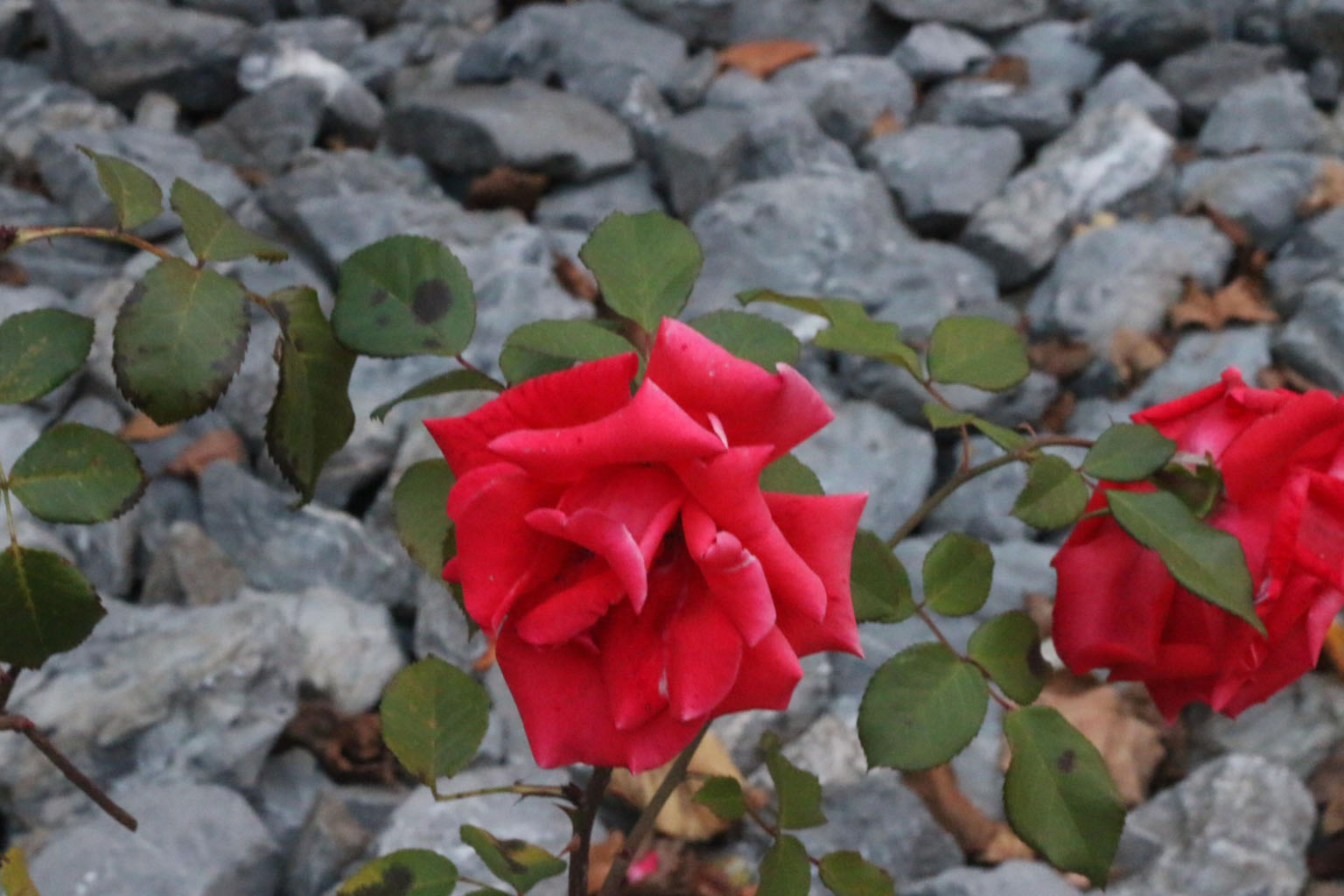}&
    \includegraphics[width=0.24\linewidth]{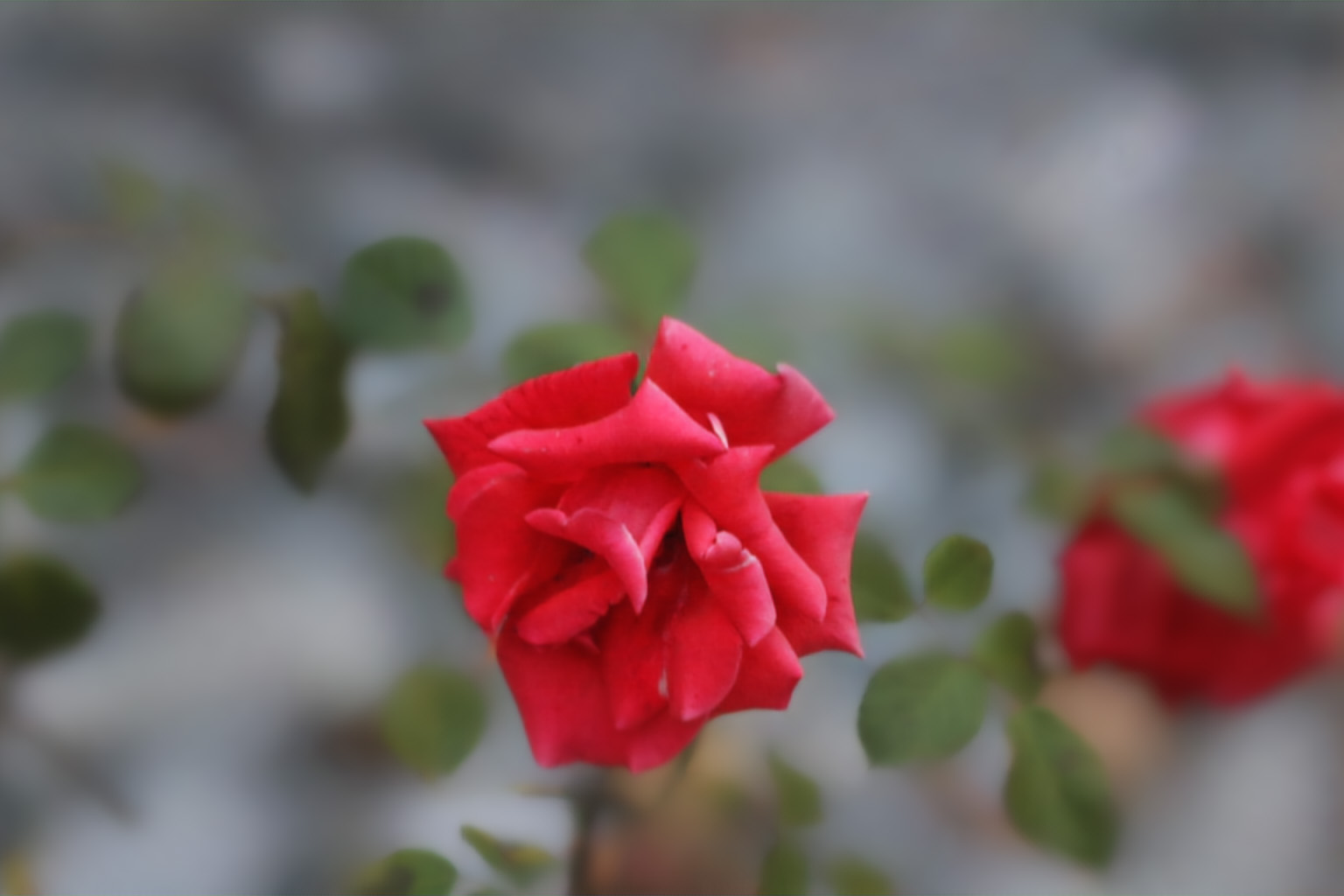}&
    \includegraphics[width=0.24\linewidth]{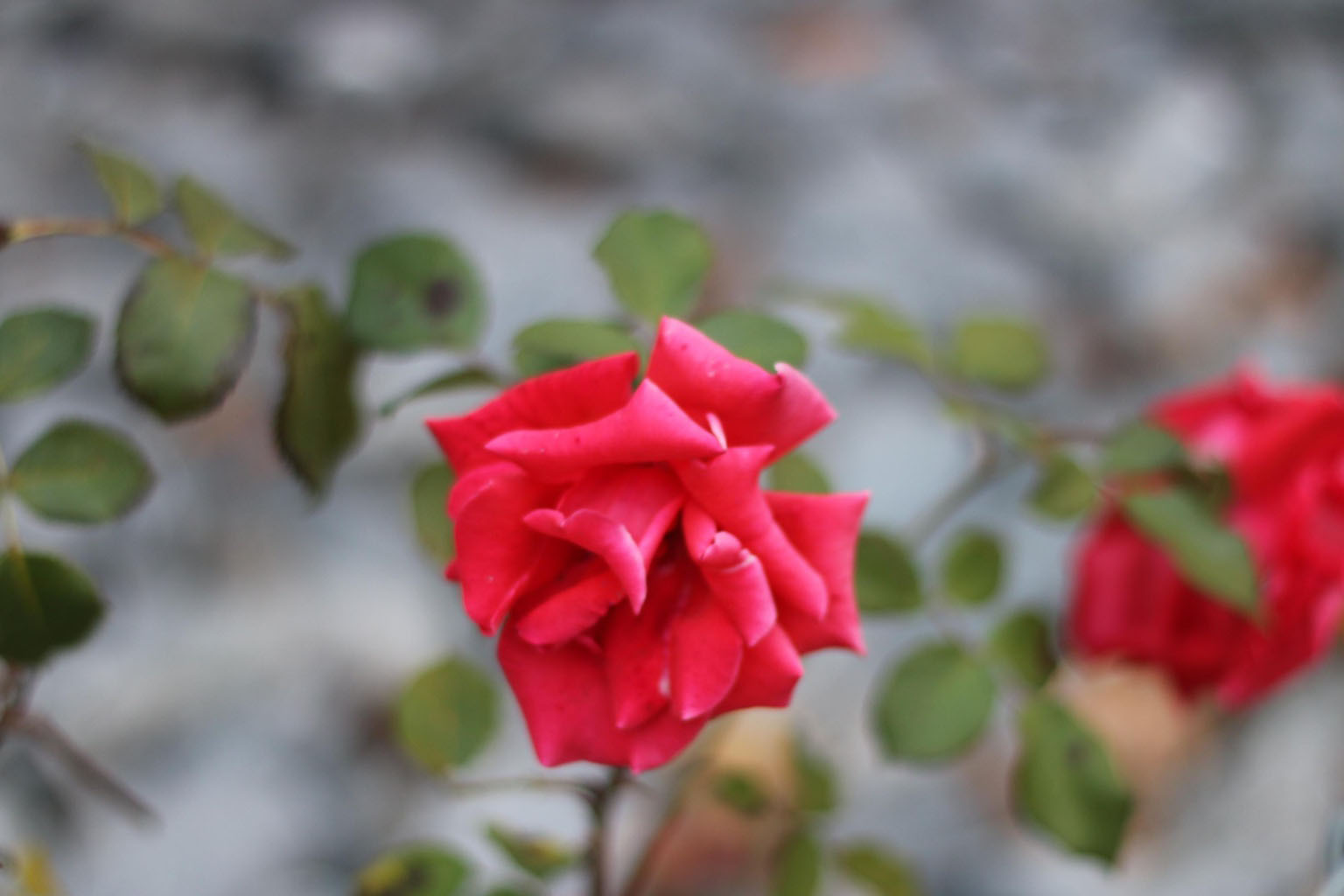}\\
    \includegraphics[width=0.24\linewidth]{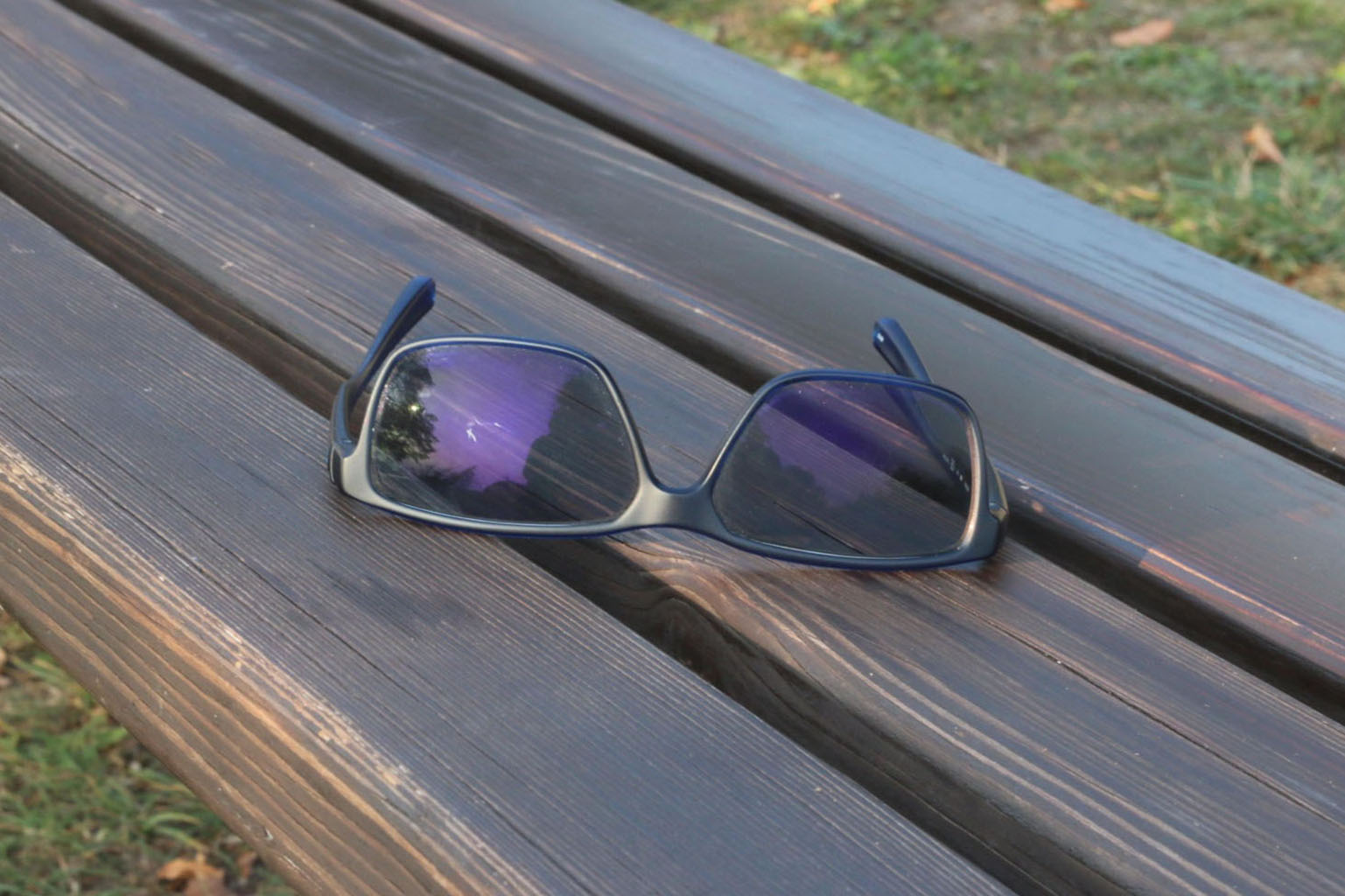}&
    \includegraphics[width=0.24\linewidth]{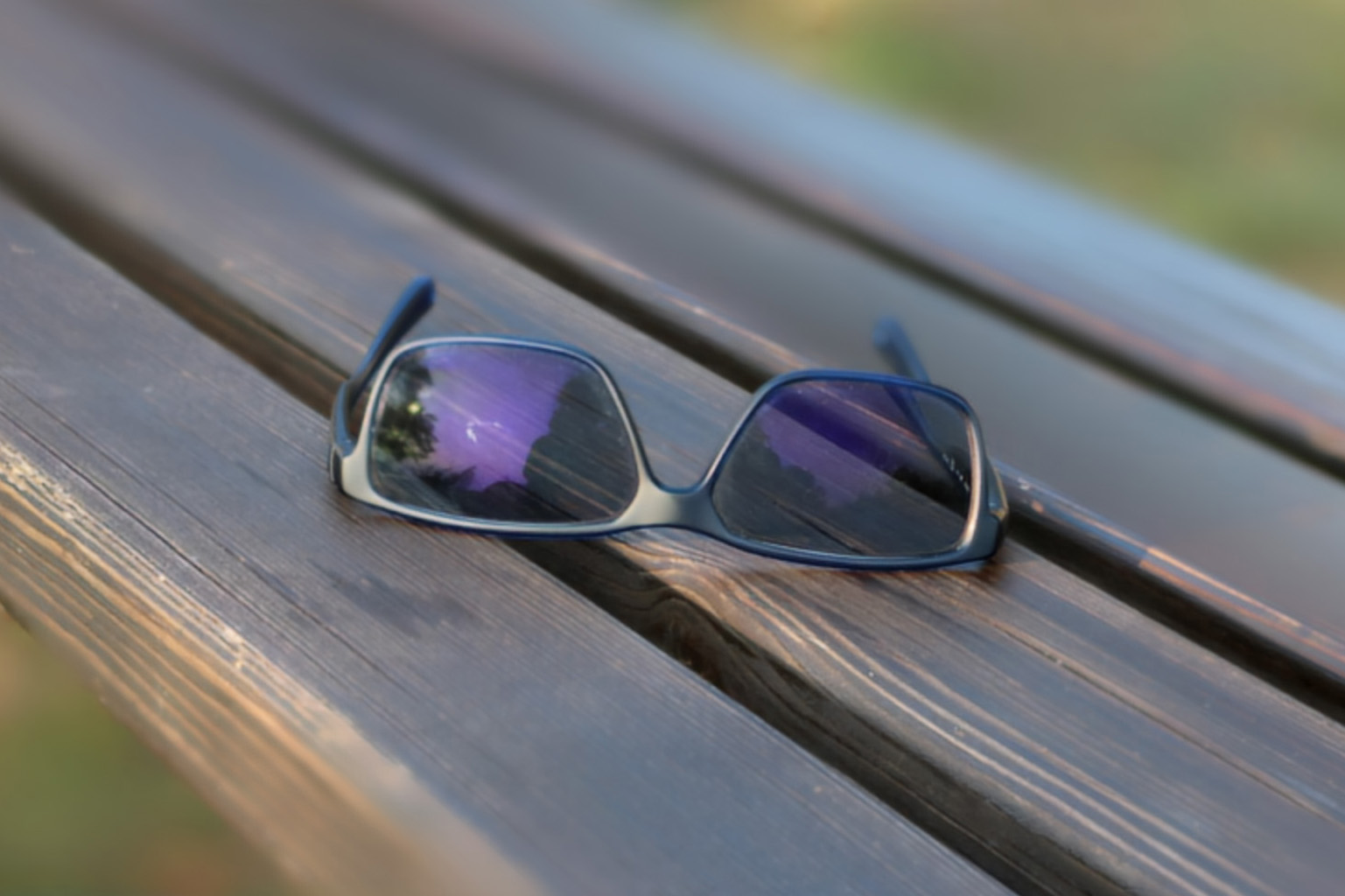}&
    \includegraphics[width=0.24\linewidth]{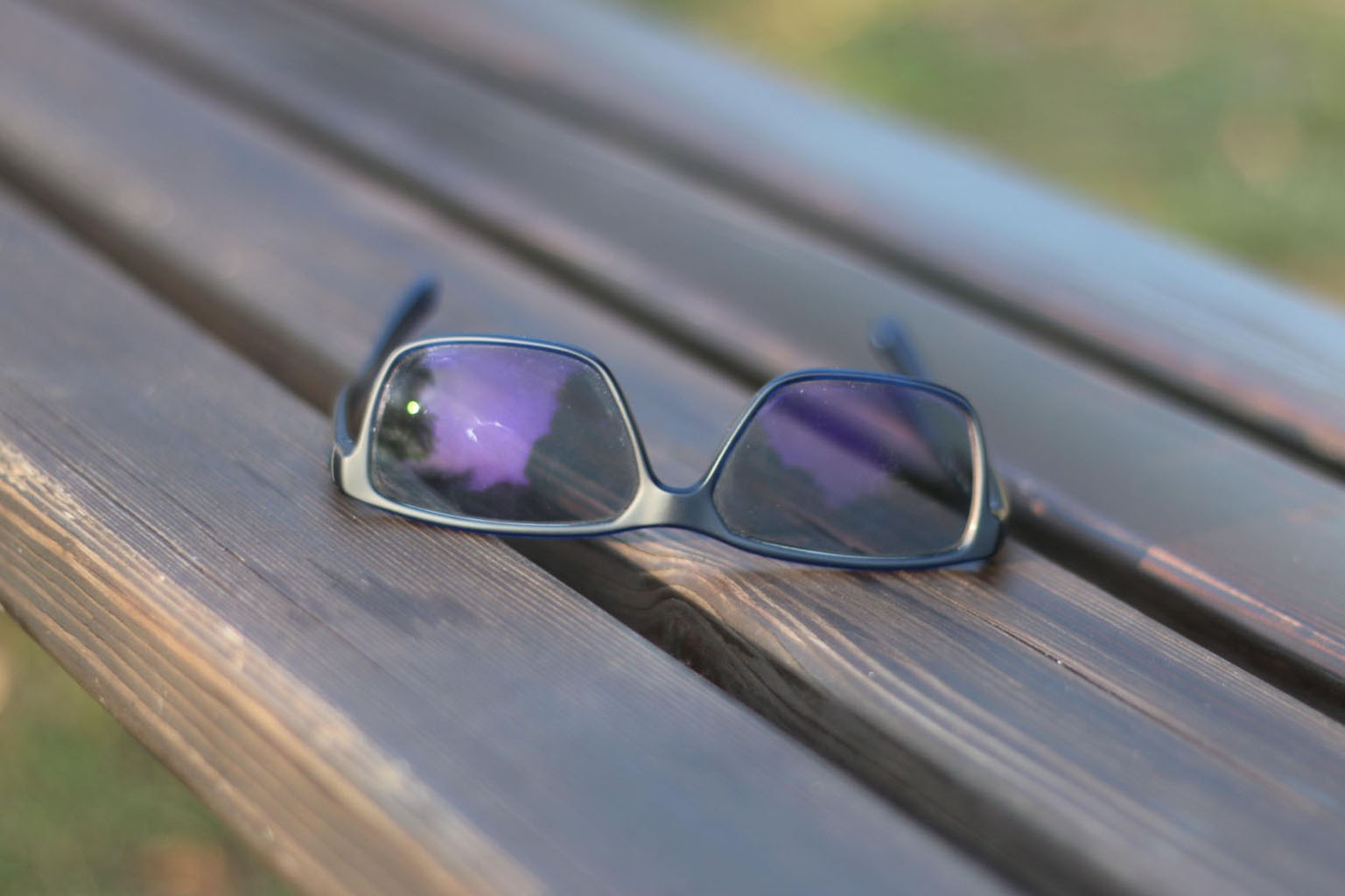}\\        \includegraphics[width=0.24\linewidth]{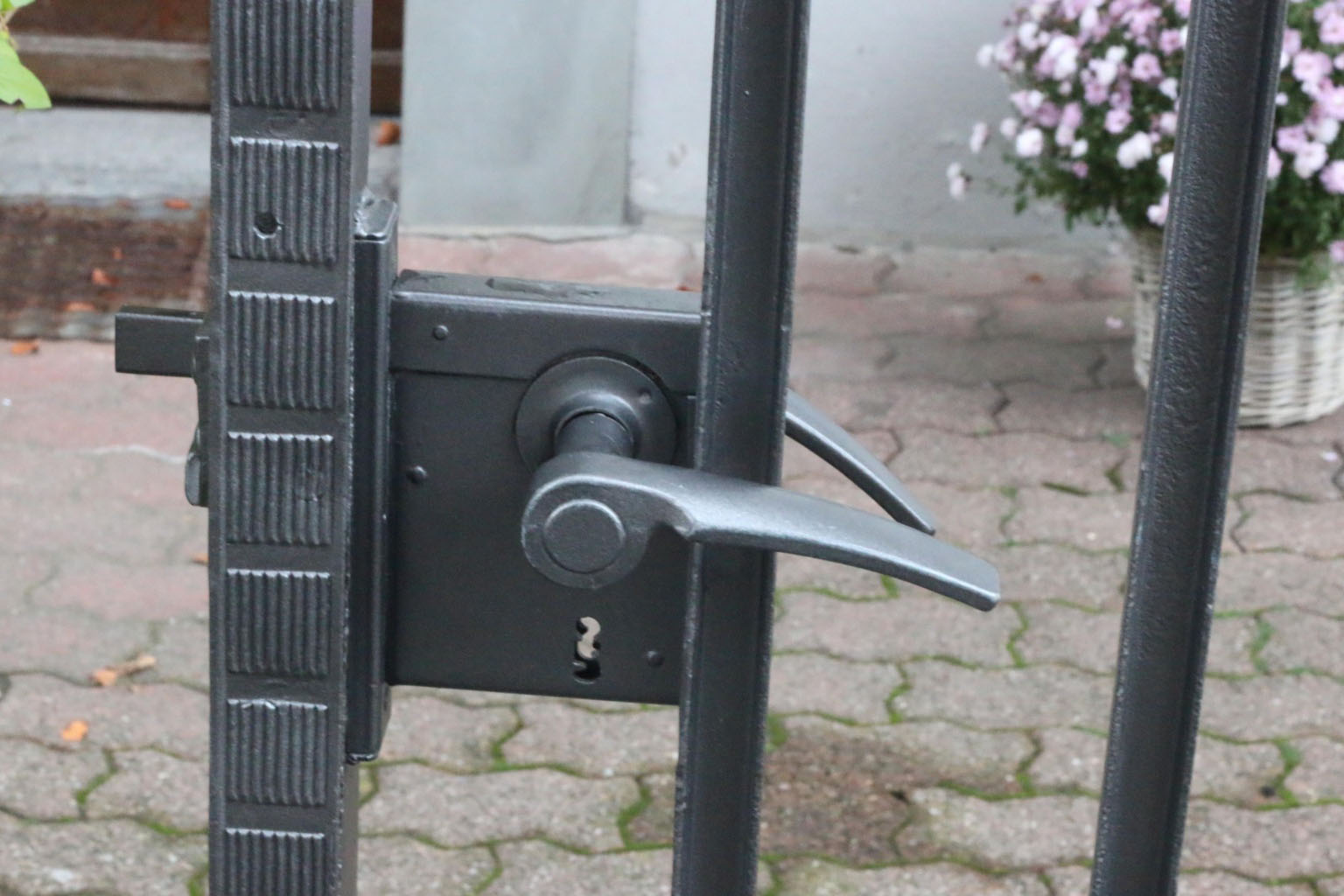}&
    \includegraphics[width=0.24\linewidth]{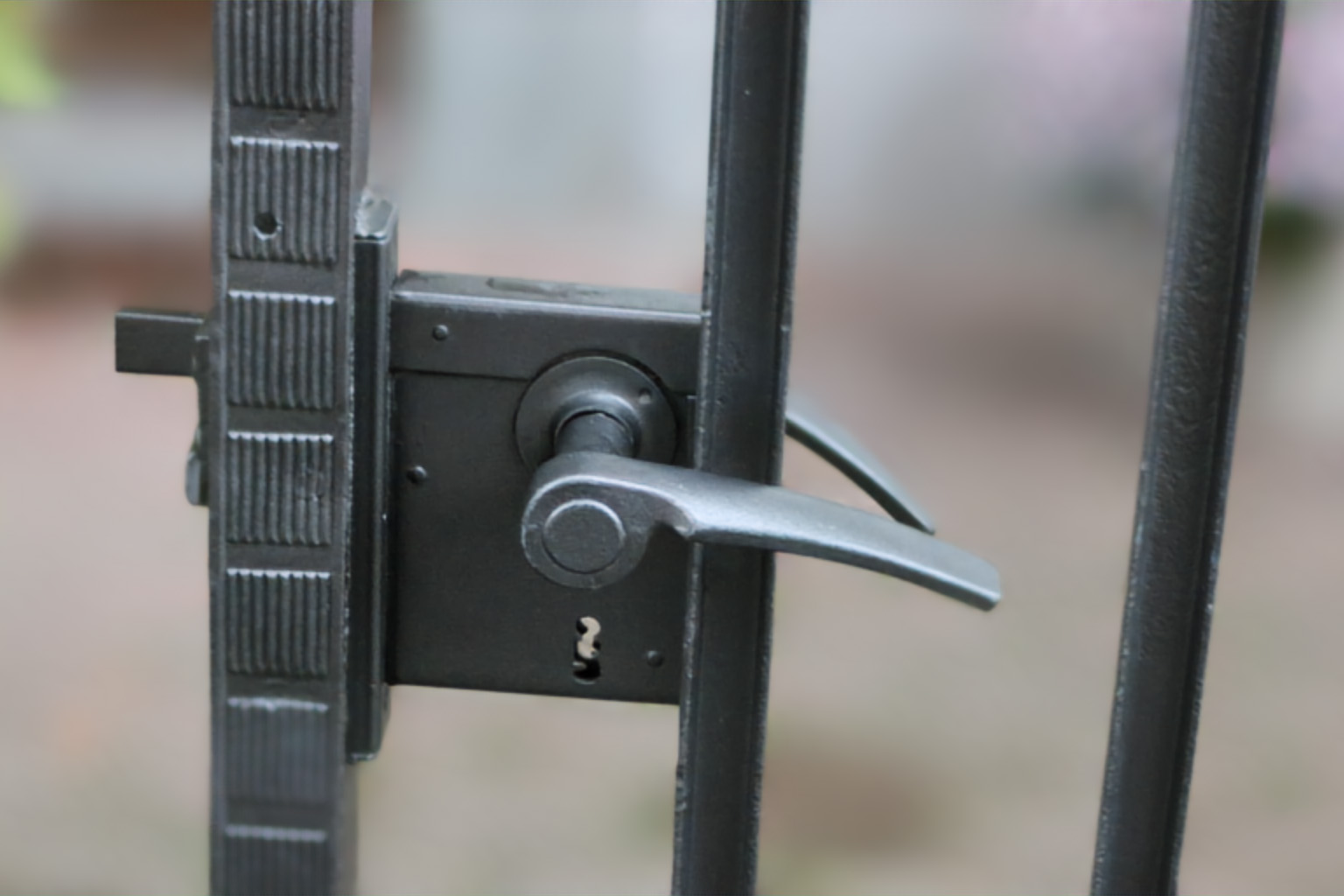}&
    \includegraphics[width=0.24\linewidth]{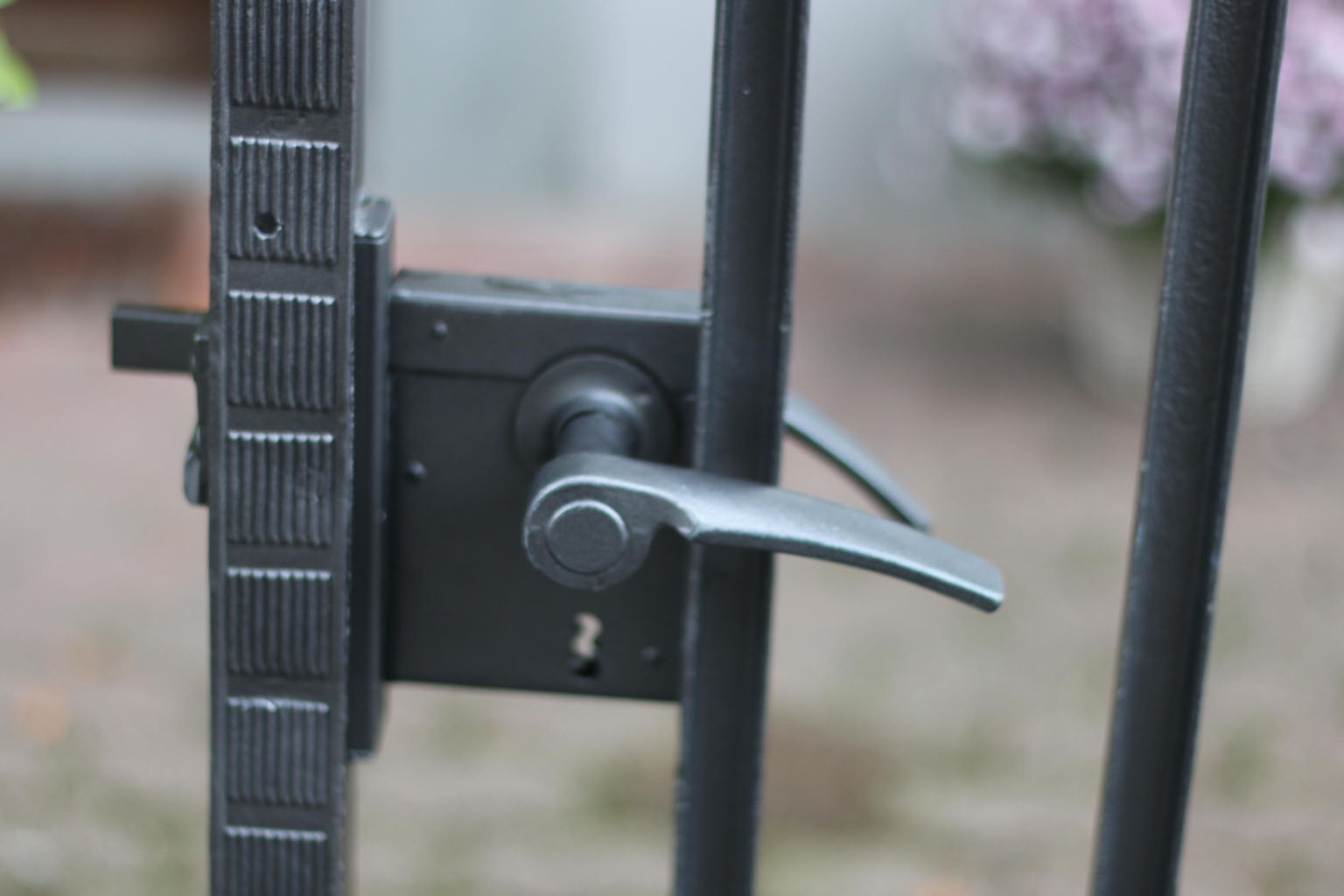}\\
    \includegraphics[width=0.24\linewidth]{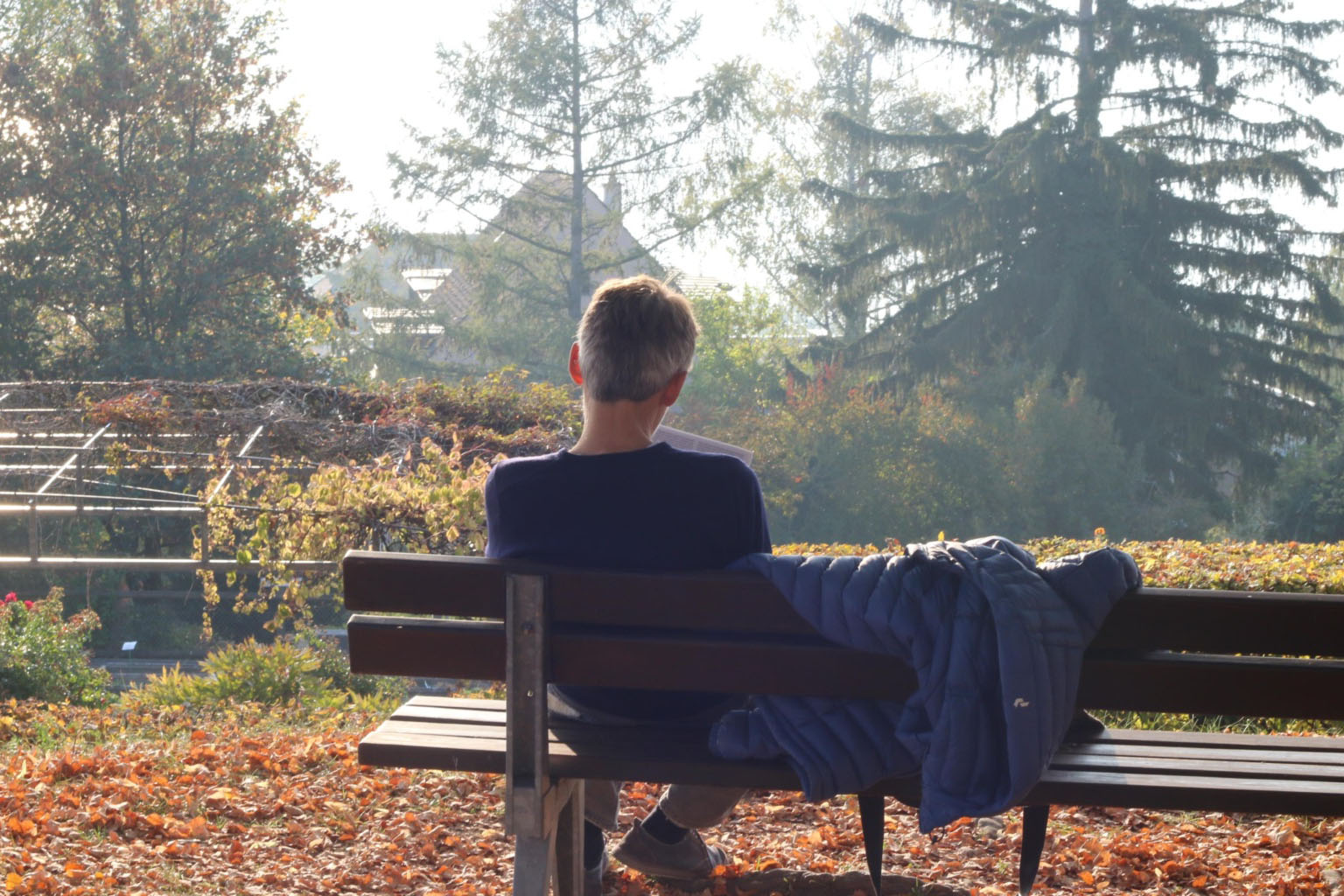}&
    \includegraphics[width=0.24\linewidth]{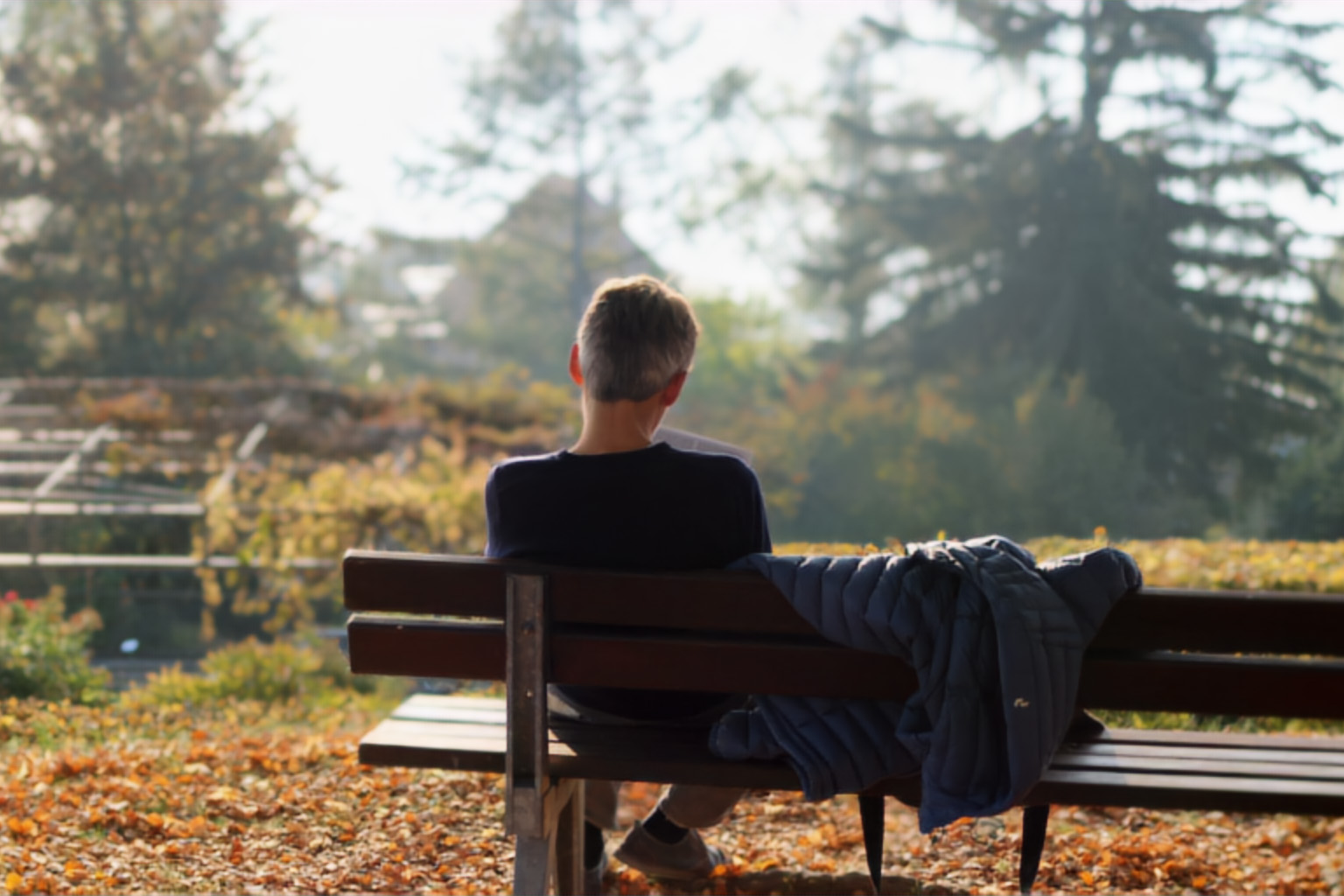}&
    \includegraphics[width=0.24\linewidth]{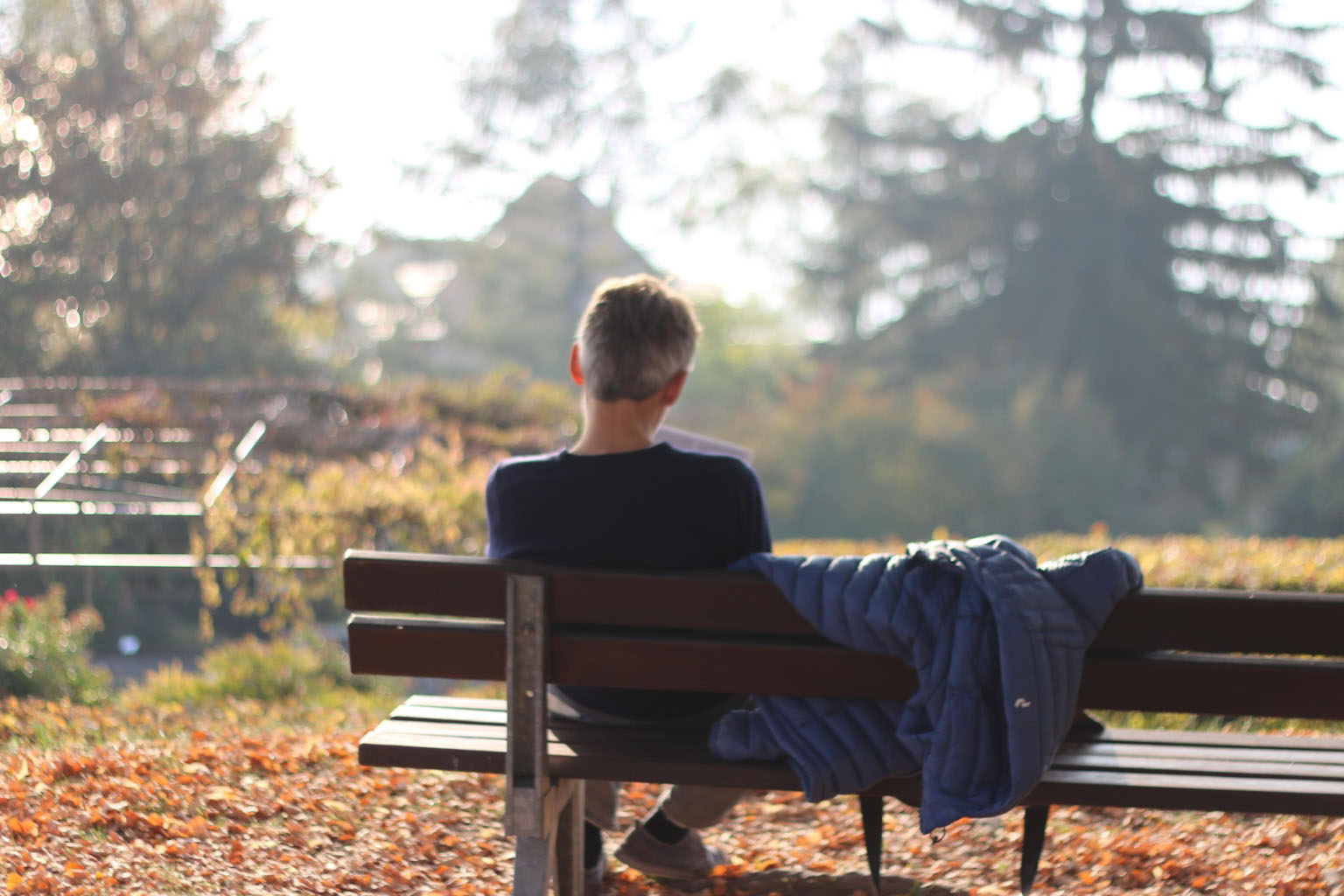}\\
\end{tabular}
}
\vspace{0.5mm}
\caption{Sample visual results obtained with the proposed deep learning method. Best zoomed on screen.}
\label{fig:results}
\vspace{-1.9mm}
\end{figure*}

\subsection{PyNET CNN Architecture}
\label{sec:pynet}

Figure~\ref{fig:Architecture} illustrates schematic representation of the PyNET-based architecture used in this work. The model has an inverted pyramidal shape and is processing the images at seven different scales. The proposed architecture has a number of blocks that are processing feature maps in parallel with convolutional filters of different size (from 3$\times$3 to 9$\times$9), and the outputs of the corresponding convolutional layers are then concatenated, which allows the network to learn a more diverse set of features at each level. The outputs obtained at lower scales are upsampled with transposed convolutional layers, stacked with feature maps from the upper level and then subsequently processed in the following convolutional layers. \textit{Leaky ReLU} activation function is applied after each convolutional operation, except for the output layers that are using \textit{tanh} function to map the results to (-1, 1) interval. Instance normalization is used in all convolutional layers that are processing images at lower scales (levels 2-5). We are additionally using two transposed convolutional layers on top of the main model that upsample the images to their target size.

The model is trained sequentially, starting from the lowest layer. This allows to achieve good semantically-driven reconstruction results at smaller scales that are working with images of very low resolution and thus performing mostly global image manipulations. After the bottom layer is pre-trained, the same procedure is applied to the next level till the training is done on the original resolution. Since each higher level is getting upscaled high-quality features from the lower part of the model, it mainly learns to reconstruct the missing low-level details and refines the results. Note that the input layer is always the same (and is getting images of size 512$\times$512 pixels during the training), though only a part of the model graph (all layers participating in producing the outputs at the corresponding scale) is trained. It should be also noted that the resolution of the produced images is twice higher than the size of the input data, which was done to increase the training and inference speed.

\begin{figure*}[t!]
\centering
\setlength{\tabcolsep}{1pt}
\resizebox{\linewidth}{!}
{
\begin{tabular}{cccccc}
    \includegraphics[width=0.24\linewidth]{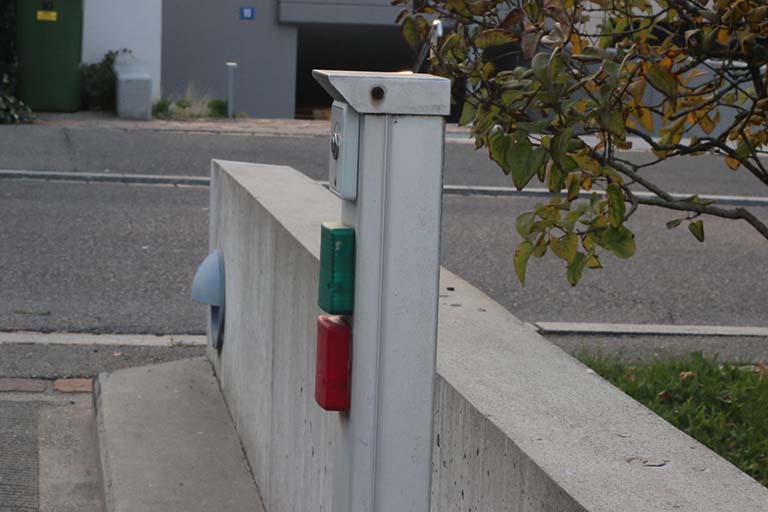}&
    \includegraphics[width=0.24\linewidth]{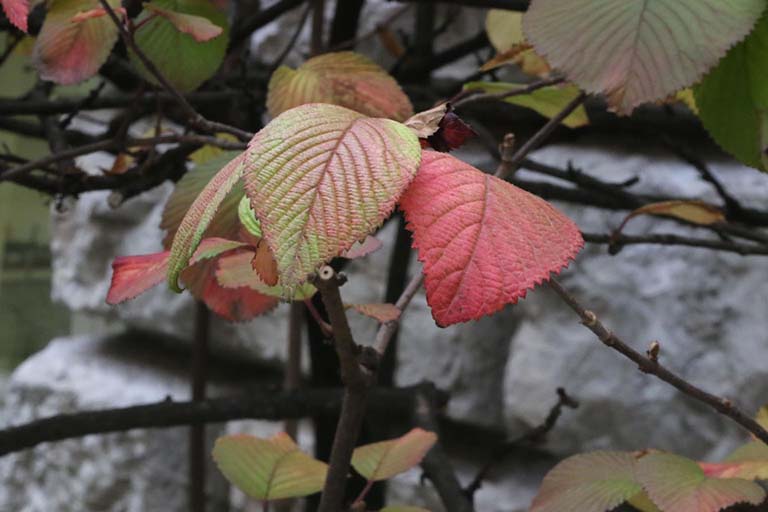}&
    \includegraphics[width=0.24\linewidth]{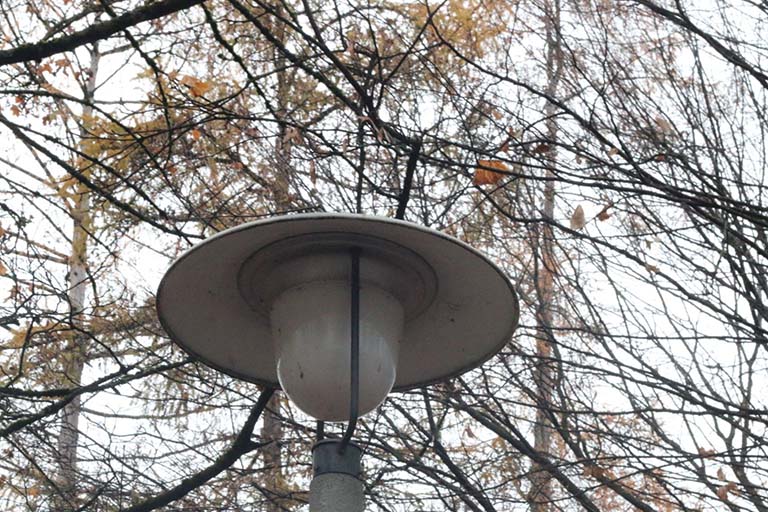}&
    \includegraphics[width=0.24\linewidth]{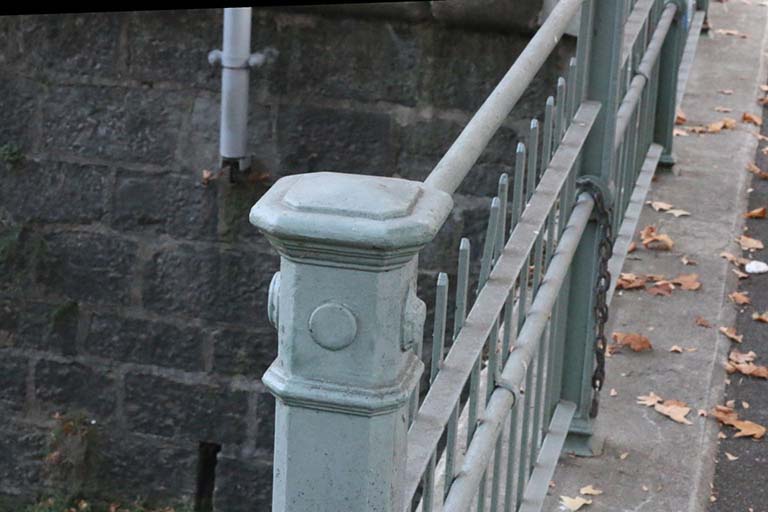}&
    \includegraphics[width=0.24\linewidth]{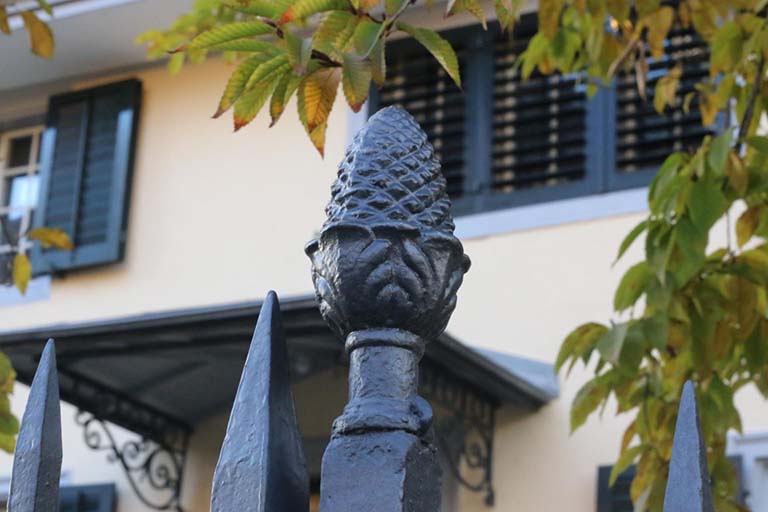}&
    \includegraphics[width=0.24\linewidth]{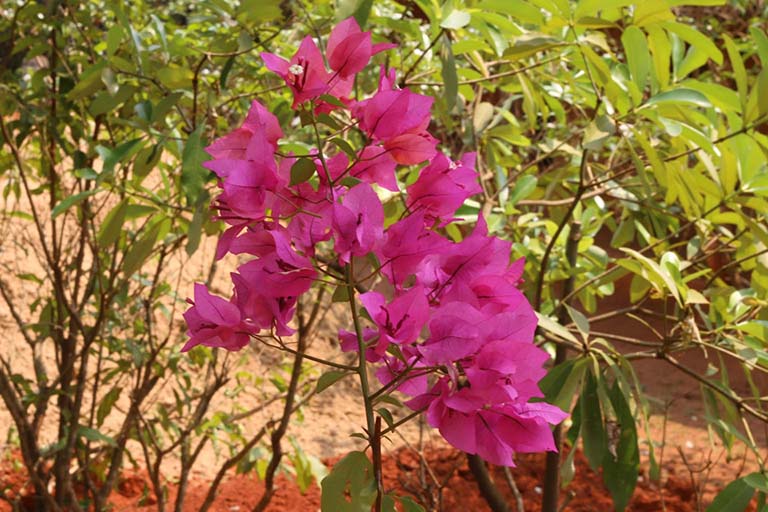}\\
    \includegraphics[width=0.24\linewidth]{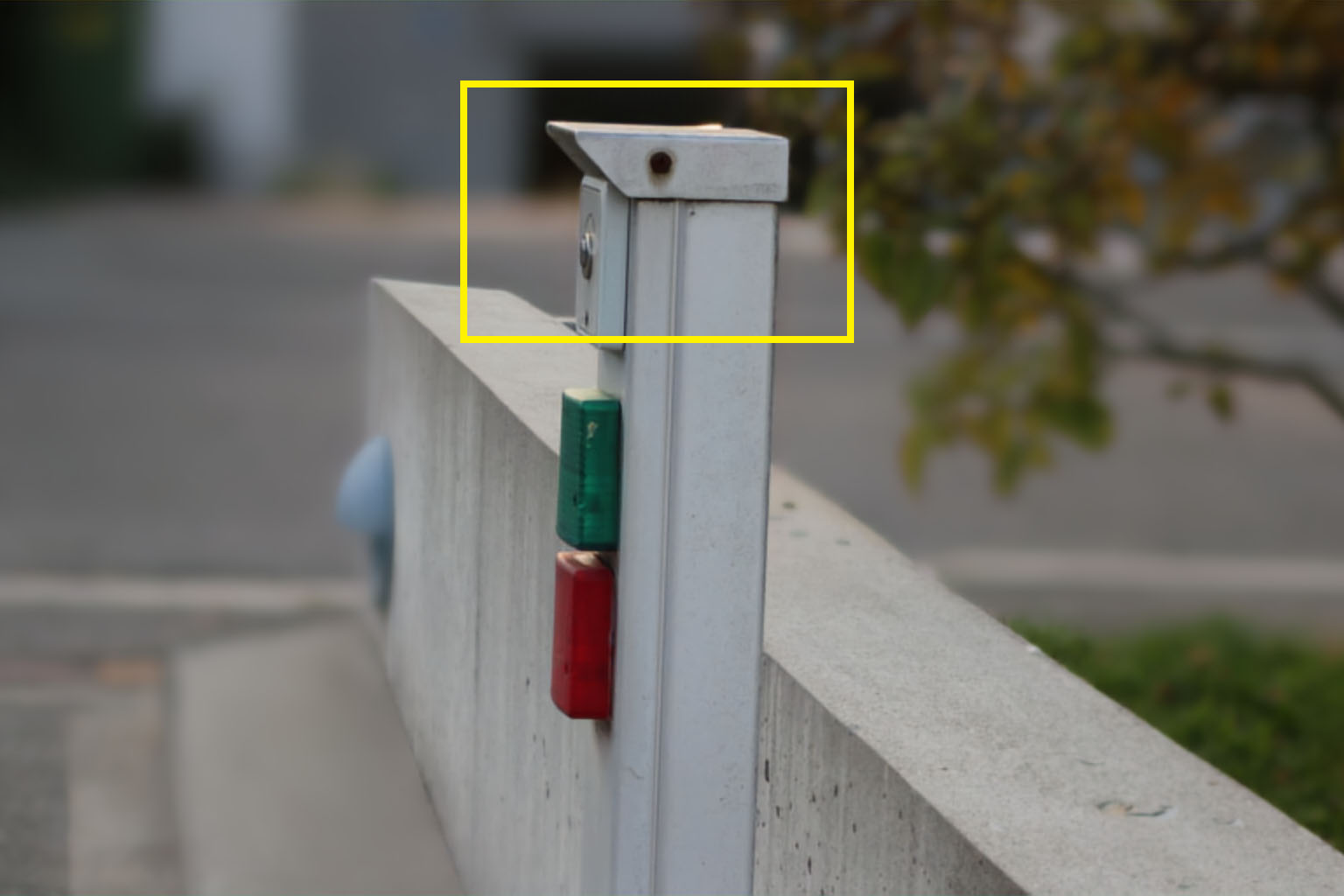}&
    \includegraphics[width=0.24\linewidth]{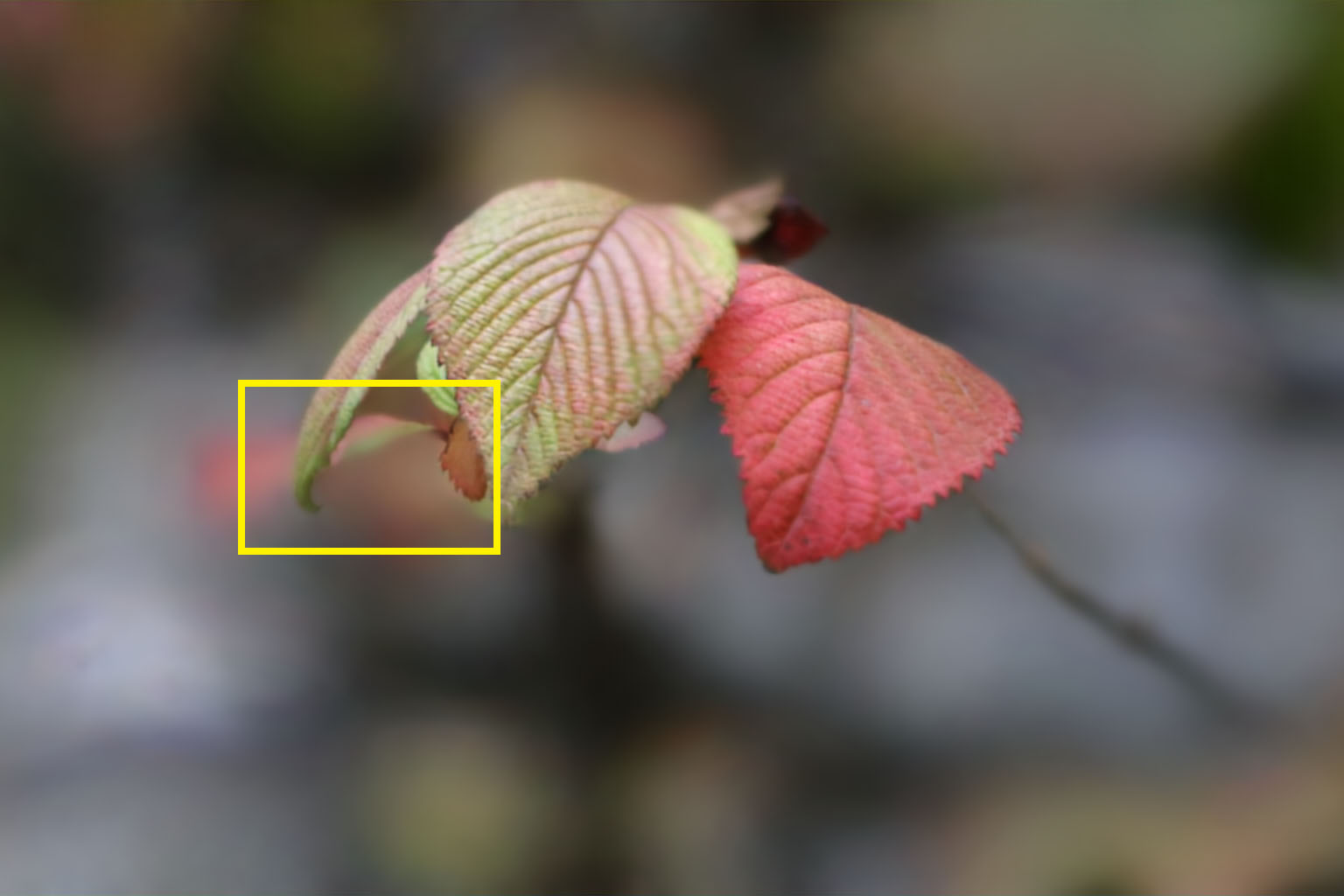}&
    \includegraphics[width=0.24\linewidth]{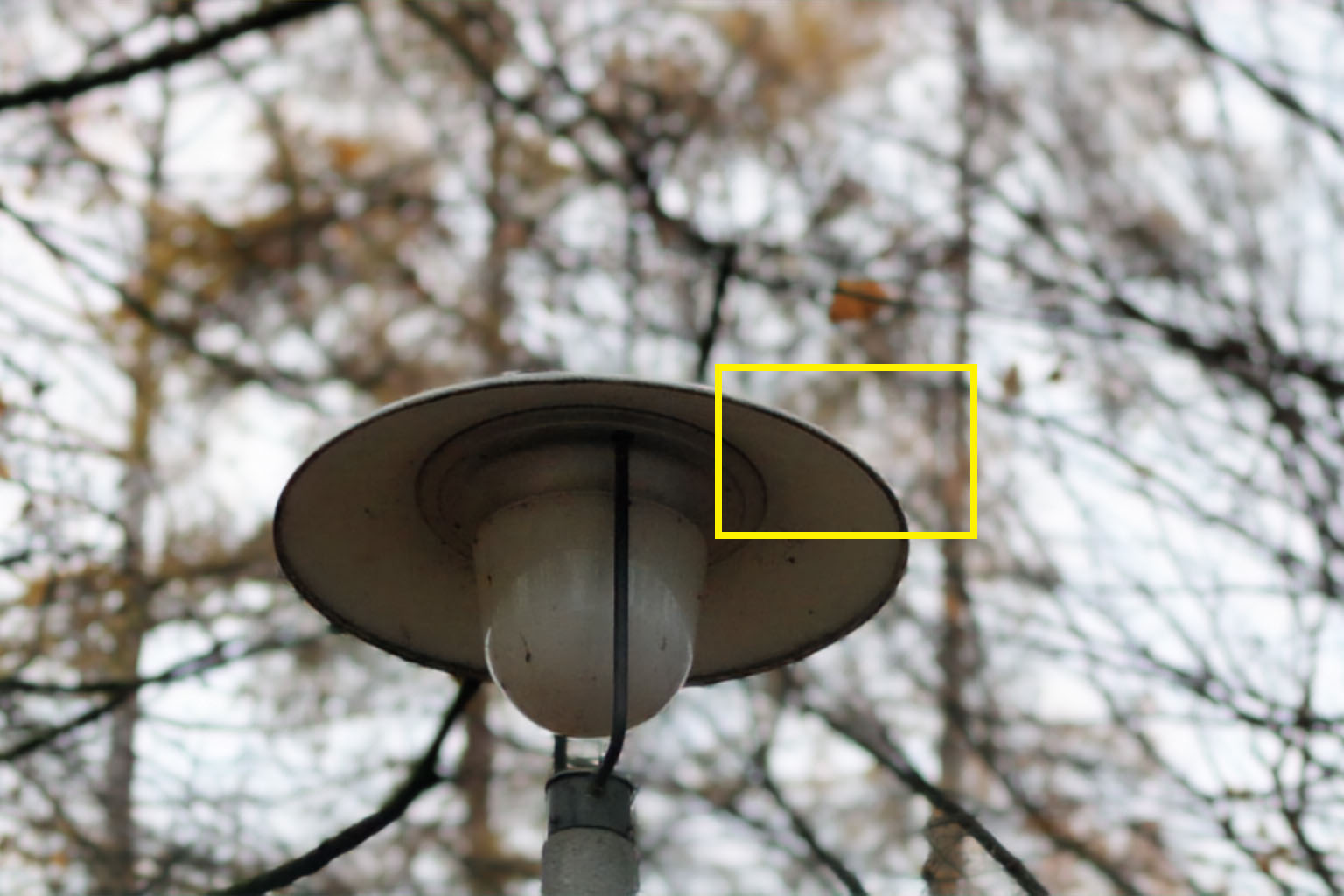}&
    \includegraphics[width=0.24\linewidth]{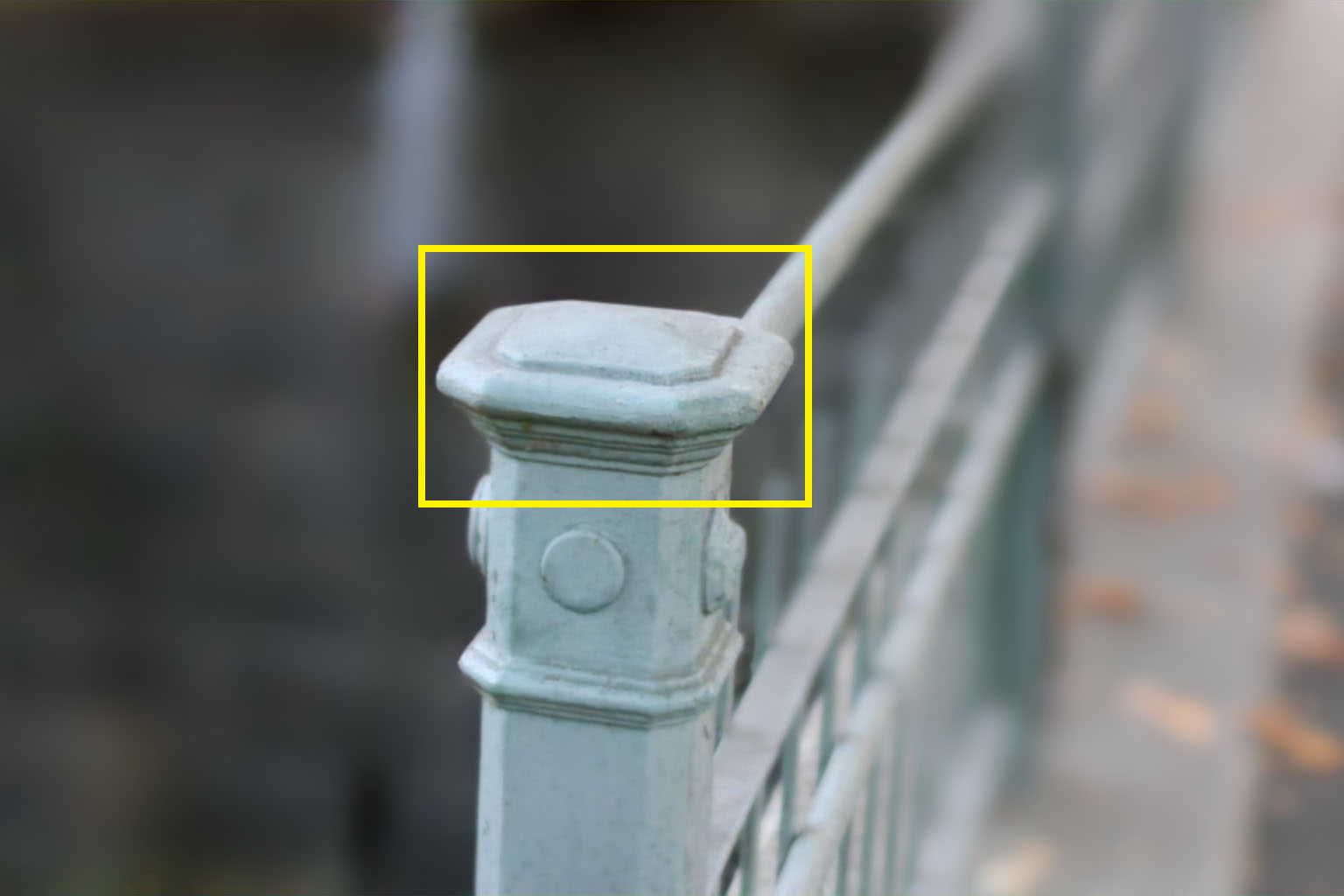}&
    \includegraphics[width=0.24\linewidth]{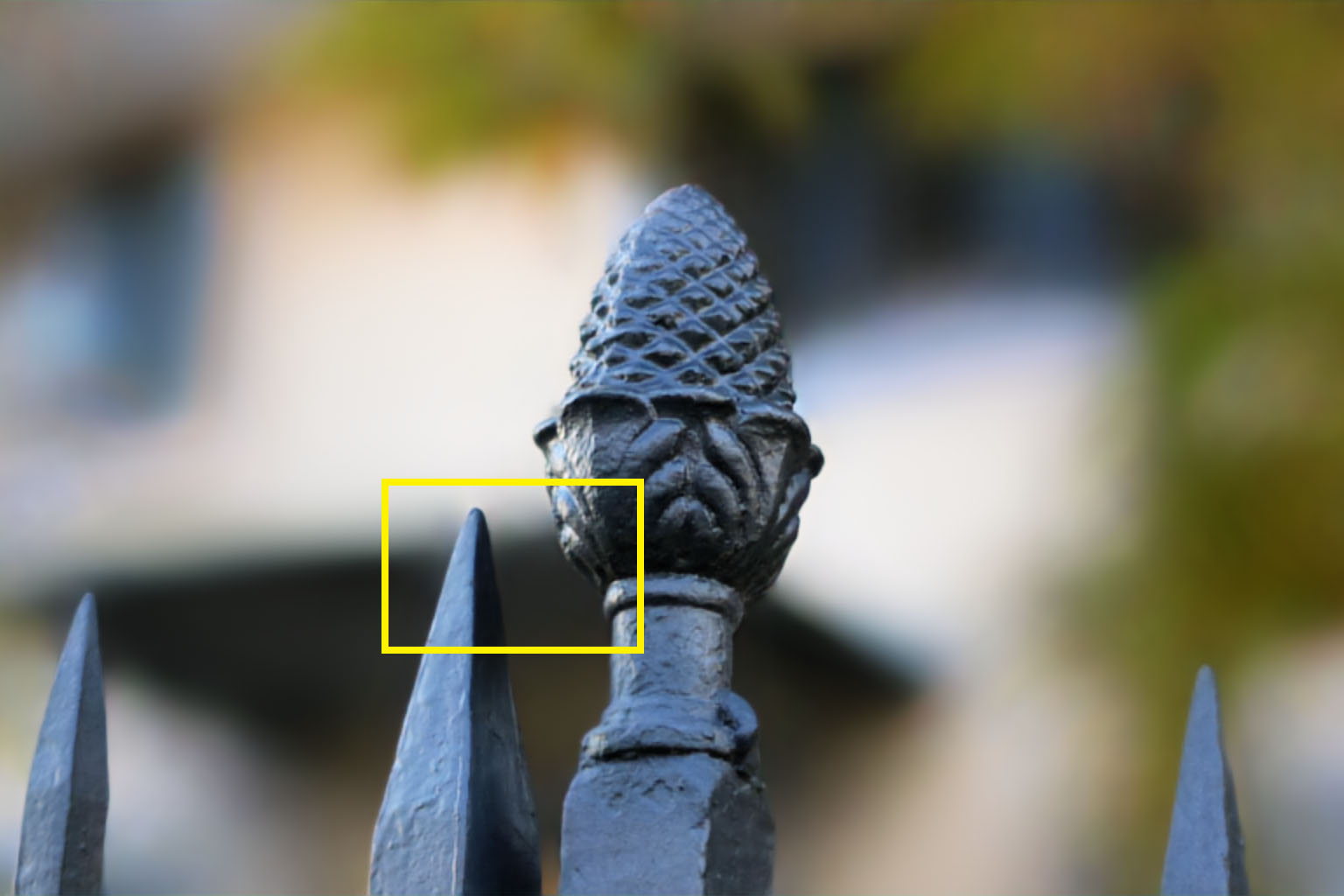}&
    \includegraphics[width=0.24\linewidth]{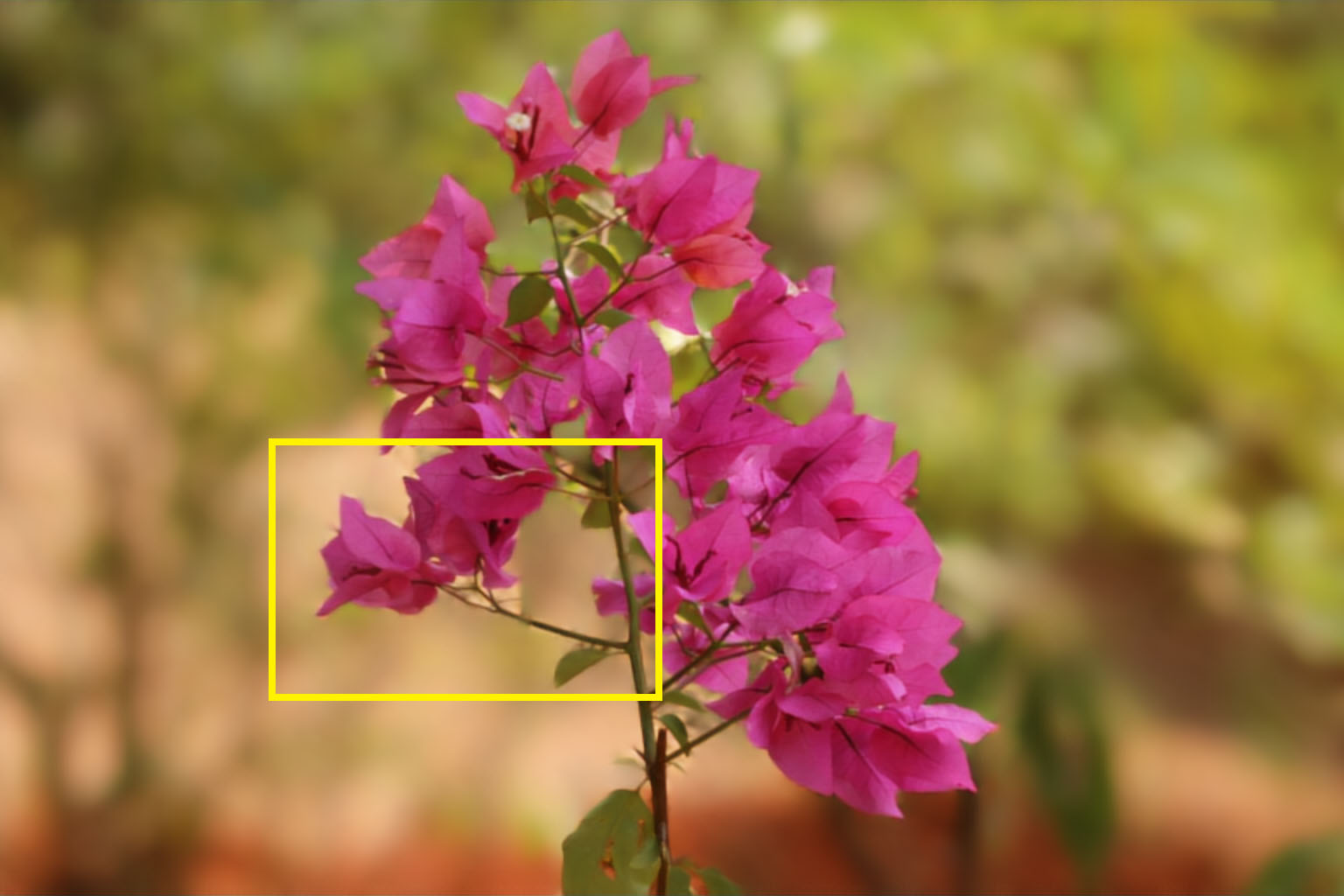}\\
    \includegraphics[width=0.24\linewidth]{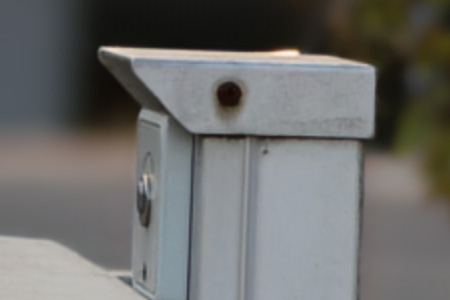}&
    \includegraphics[width=0.24\linewidth]{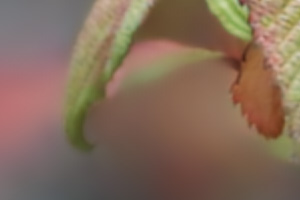}&
    \includegraphics[width=0.24\linewidth]{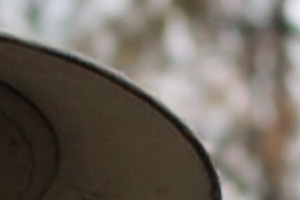}&
    \includegraphics[width=0.24\linewidth]{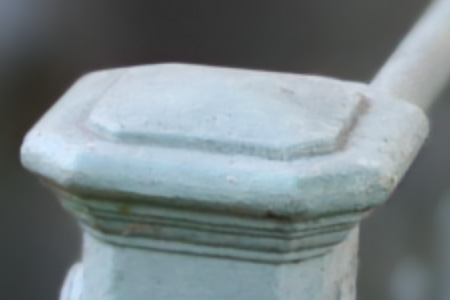}&
    \includegraphics[width=0.24\linewidth]{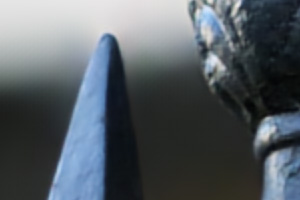}&
    \includegraphics[width=0.24\linewidth]{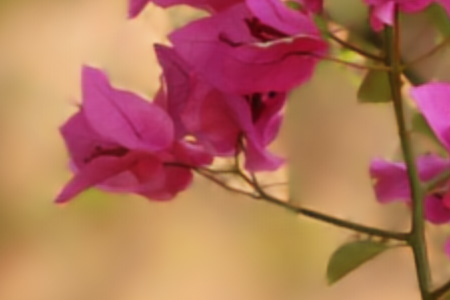}\\
\end{tabular}
}
\vspace{0.5mm}
\caption{The quality of the rendered boundary regions between in and out of focus areas for different image scenes.}
\label{fig:sample_crops}
\vspace{-0.9mm}
\end{figure*}

\vspace{-1.2mm}
\subsection{Training Details}

\begin{figure*}[t!]
\centering
\setlength{\tabcolsep}{1pt}
\resizebox{\linewidth}{!}
{
\begin{tabular}{ccccc}
    \includegraphics[width=0.24\linewidth]{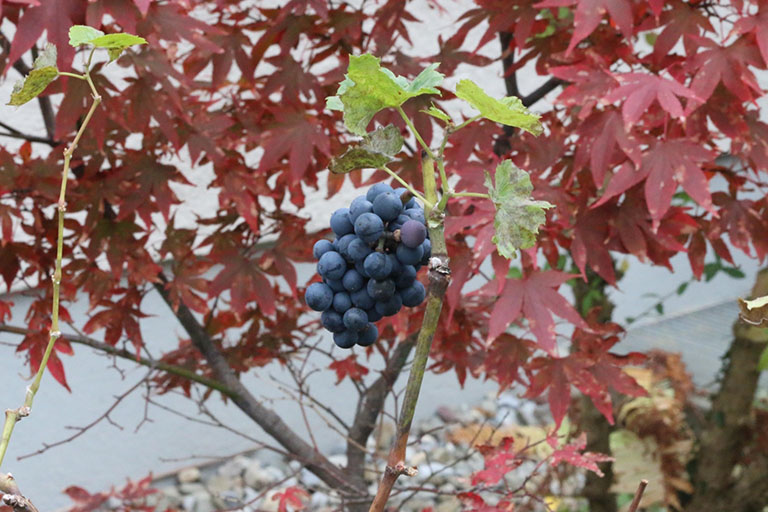}&
    \includegraphics[width=0.24\linewidth]{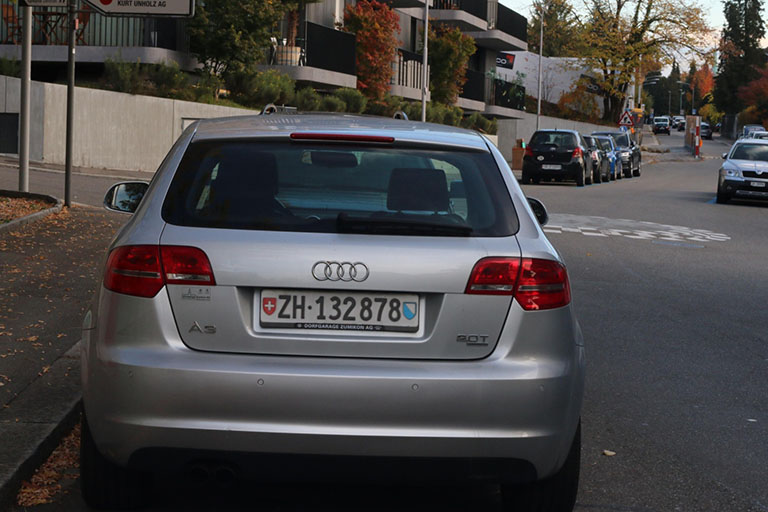}&
    \includegraphics[width=0.24\linewidth]{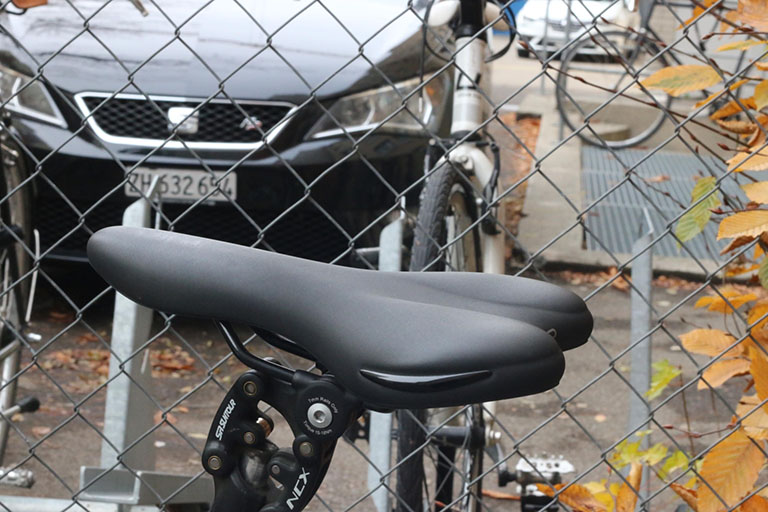}&
    \includegraphics[width=0.24\linewidth]{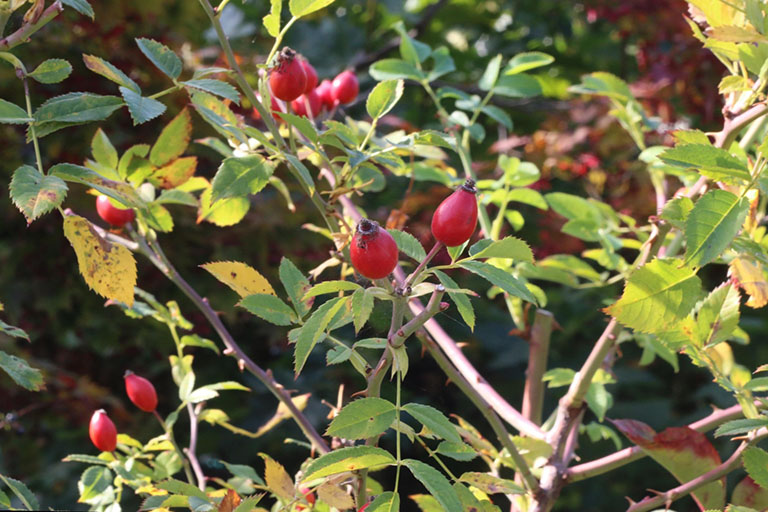}&
    \includegraphics[width=0.24\linewidth]{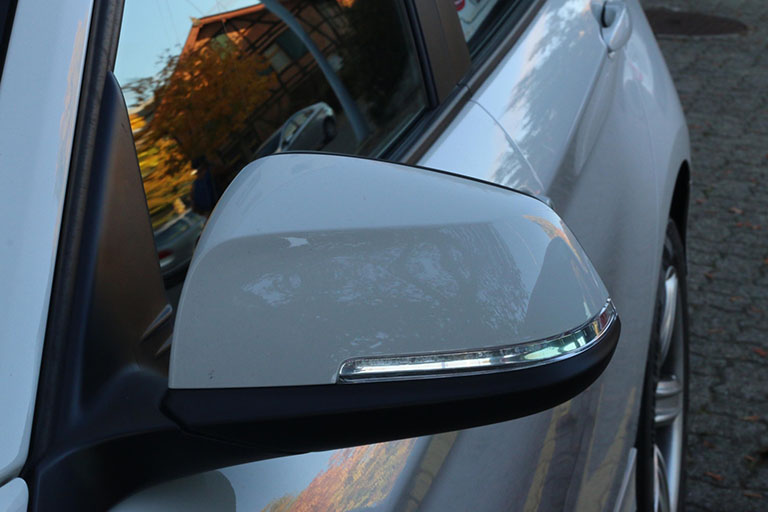}\\
    \includegraphics[width=0.24\linewidth]{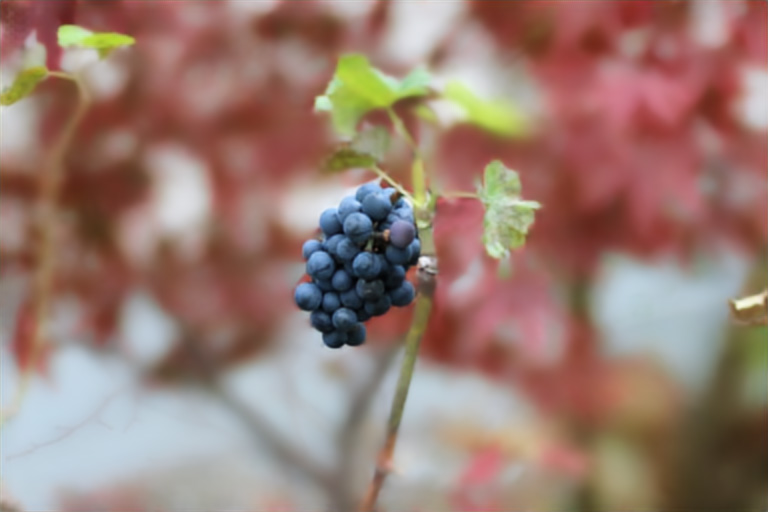}&
    \includegraphics[width=0.24\linewidth]{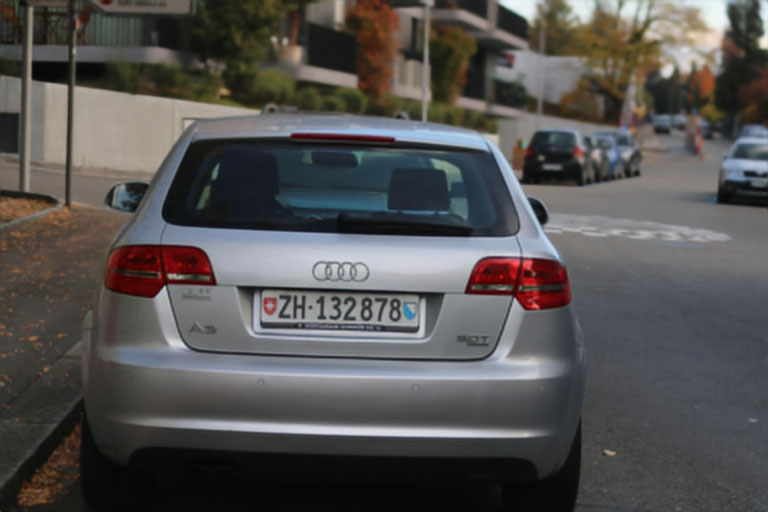}&
    \includegraphics[width=0.24\linewidth]{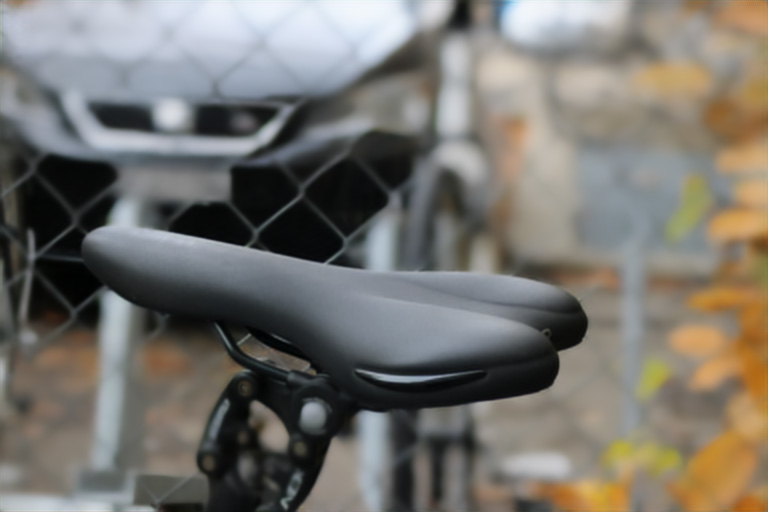}&
    \includegraphics[width=0.24\linewidth]{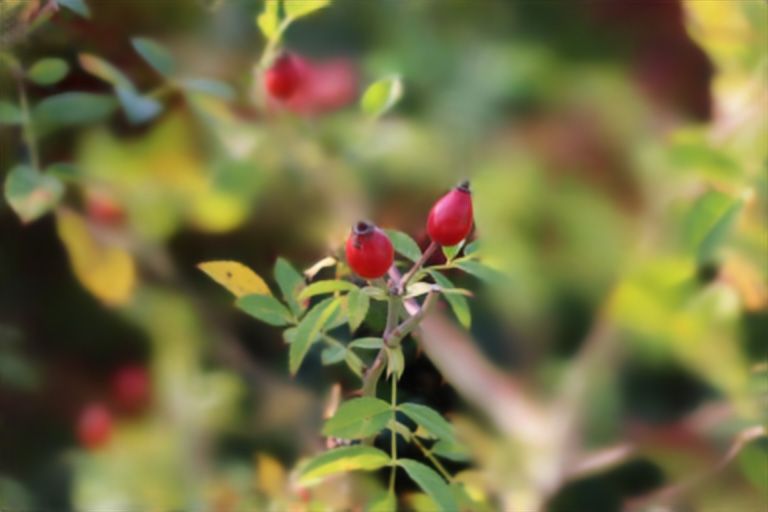}&
    \includegraphics[width=0.24\linewidth]{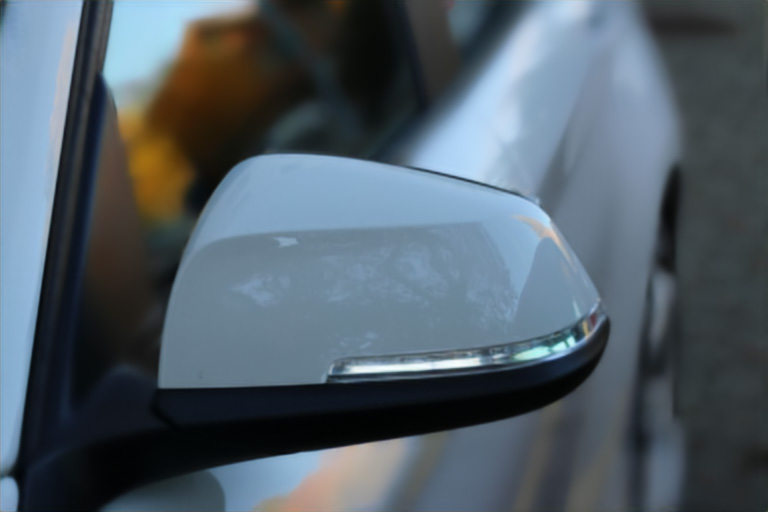}\\
    \includegraphics[width=0.24\linewidth]{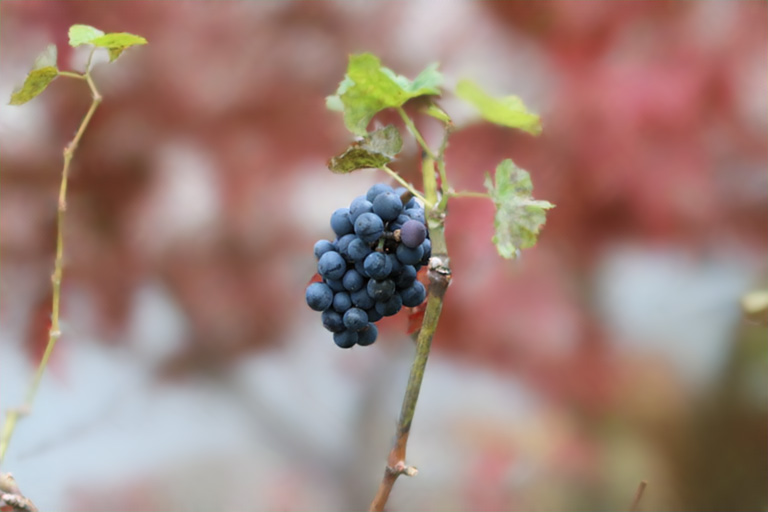}&
    \includegraphics[width=0.24\linewidth]{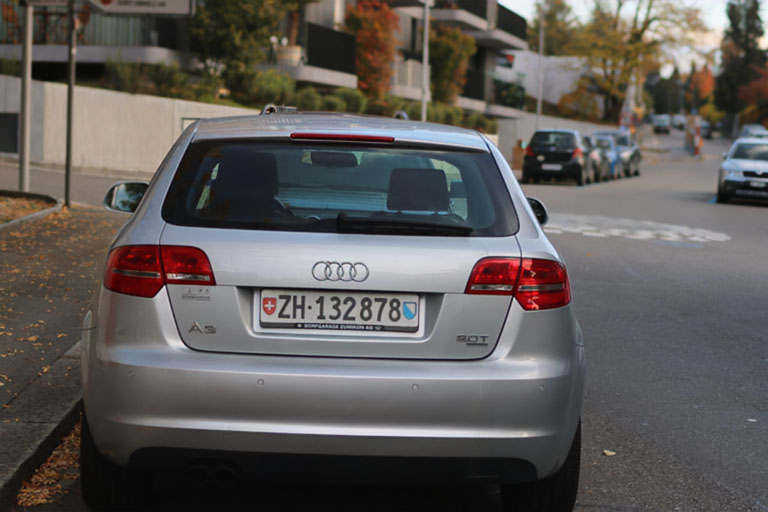}&
    \includegraphics[width=0.24\linewidth]{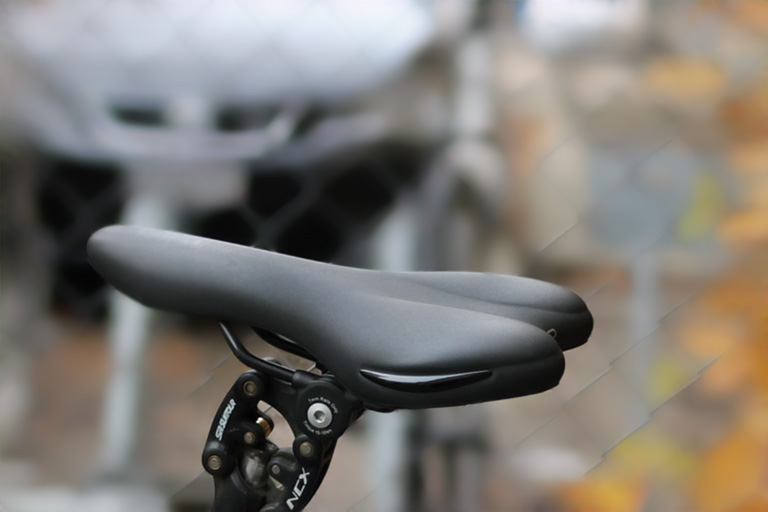}&
    \includegraphics[width=0.24\linewidth]{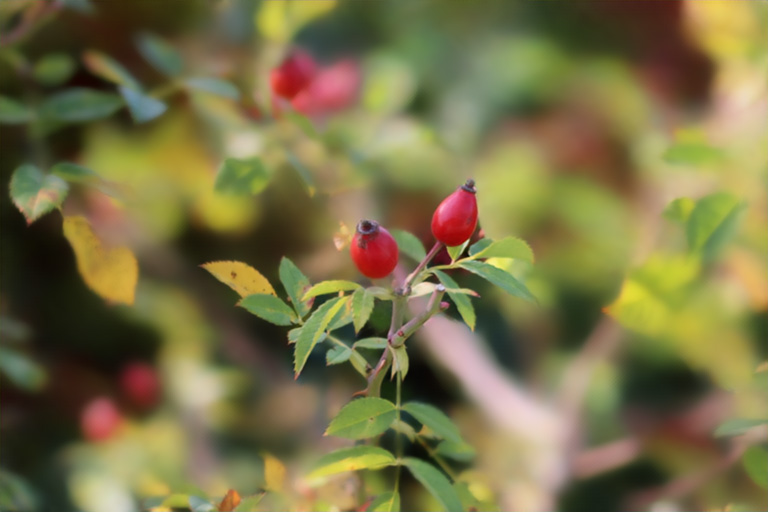}&
    \includegraphics[width=0.24\linewidth]{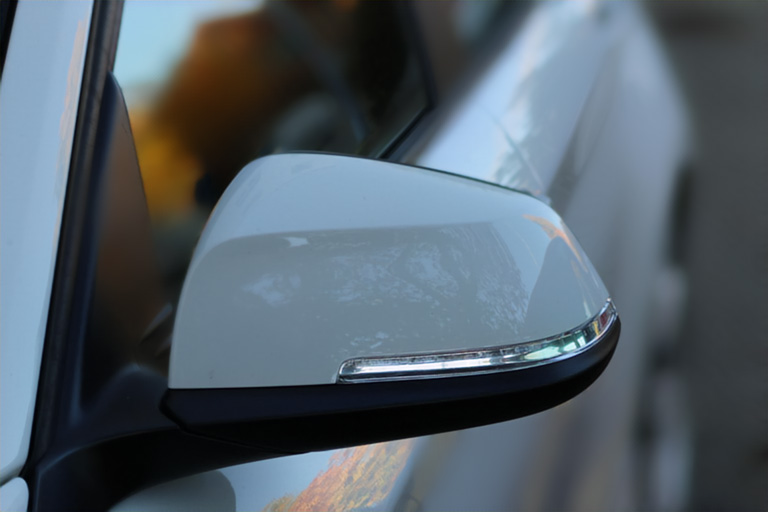}\\
\end{tabular}
}
\vspace{0.7mm}
\caption{Sample visual results obtained without (2nd row) and with (3rd row) the estimated depth map.}
\label{fig:depth_impact}
\vspace{-0.7mm}
\end{figure*}

Our initial results demonstrated that though adding the pre-computed depth map to the input data does not change the results radically (Fig.~\ref{fig:depth_impact}), this helps to refine the shape of the blurred image area and to deal with the most complex photo scenes. Therefore, in all subsequent experiments the estimated depth map was used by default.

All model levels were trained with the $L_1$ loss except for the output level 1 where the following combination of the loss functions was used:
\begin{equation*}
\label{eq:loss}
\mathcal{L}_{\text{Level 1}} = \mathcal{L}_{L_1} + (1 - \mathcal{L}_{\text{SSIM}}) + 0.01 \cdot \mathcal{L}_{\text{VGG}},
\end{equation*}
where $\mathcal{L}_{\text{SSIM}}$ is the structural similarity (SSIM) loss~\cite{wang2003multiscale} and $\mathcal{L}_{\text{VGG}}$ is the perceptual VGG-based~\cite{johnson2016perceptual} loss function. The above coefficients were chosen based on the results of the preliminary experiments on the considered EBB! dataset. We should emphasize that each level is trained together with all (already pre-trained) lower levels to ensure a deeper connection between the layers.

The model was implemented in TensorFlow~\footnote{\,\,\url{https://github.com/aiff22/PyNET-Bokeh}} and was trained on a single \textit{Nvidia Tesla V100} GPU with a batch size ranging from 8 to 50 depending on the training scale. The parameters of the model were optimized for 20 $\sim$ 100 epochs using ADAM~\cite{kingma2014adam} algorithm with a learning rate of $5e-5$. The entire PyNET model consists of 47.5M parameters, and it takes on average 143 milliseconds to produce one bokeh image of size 1024$\times$1536 pixels on the above mentioned GPU.

\begin{figure*}[b!]
\centering
\vspace{-0.7mm}
\caption{Examples of strong visual artifacts present on the EBB! test photos.}
\vspace{0.2mm}
\setlength{\tabcolsep}{1pt}
\resizebox{\linewidth}{!}
{
\begin{tabular}{ccccccc}
    \includegraphics[width=0.24\linewidth]{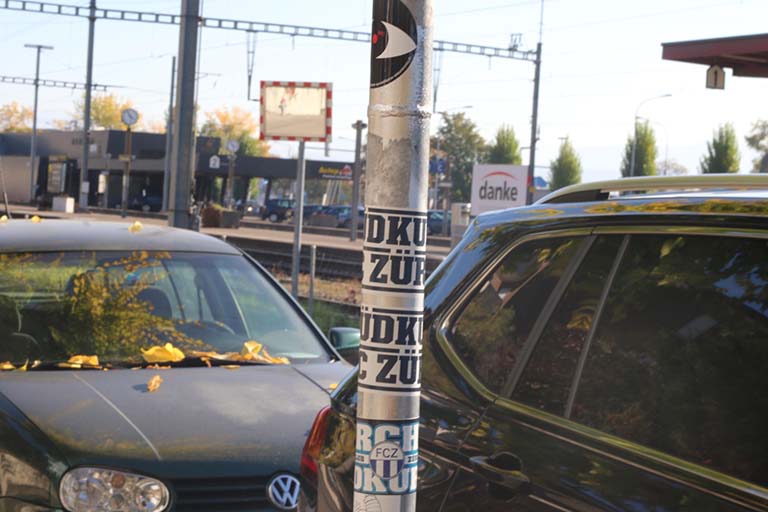}&
    \includegraphics[width=0.24\linewidth]{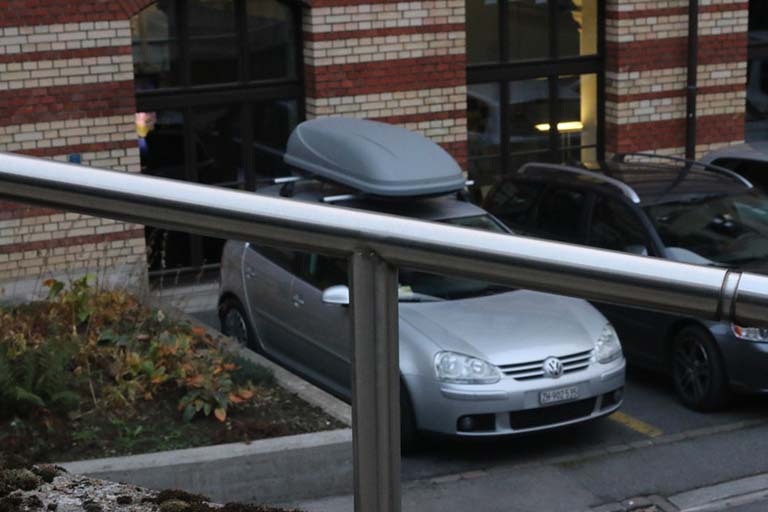}&
    \includegraphics[width=0.24\linewidth]{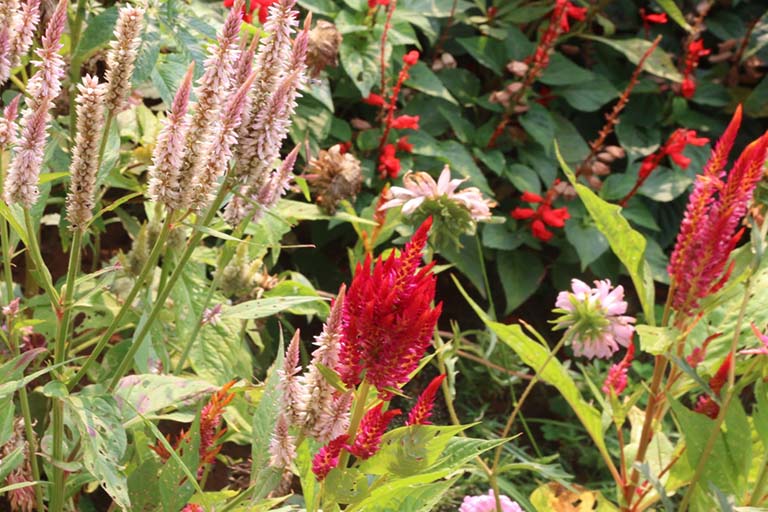}&
    \includegraphics[width=0.24\linewidth]{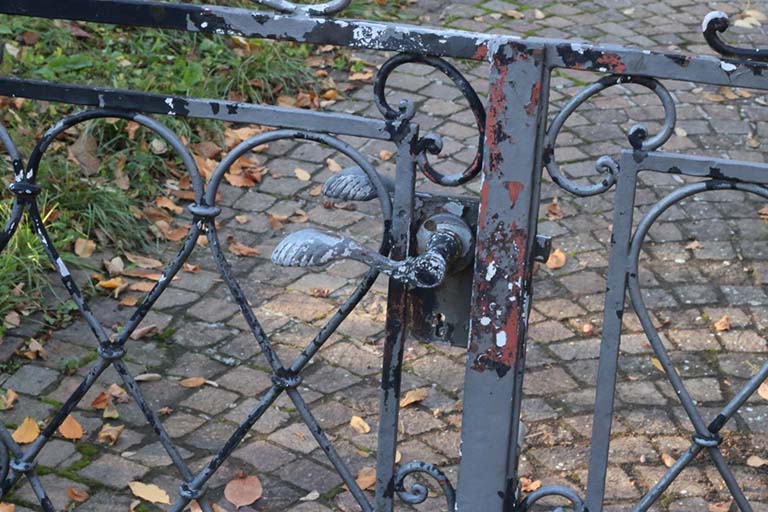}
    \includegraphics[width=0.24\linewidth]{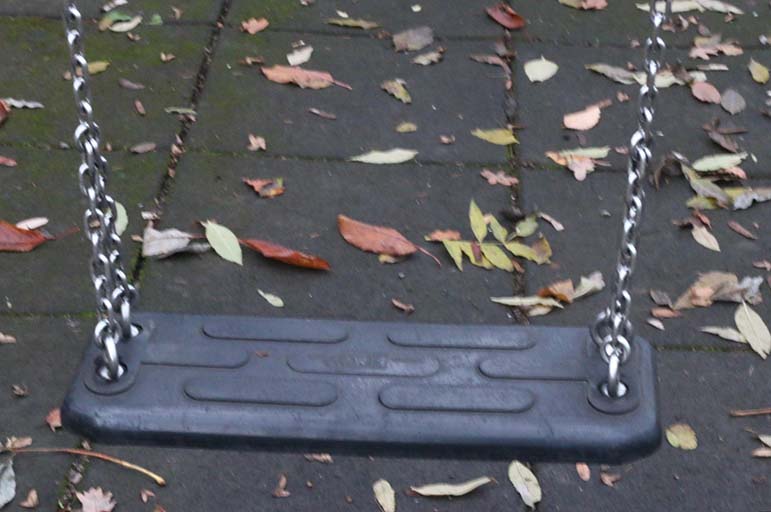}&
    \includegraphics[width=0.24\linewidth]{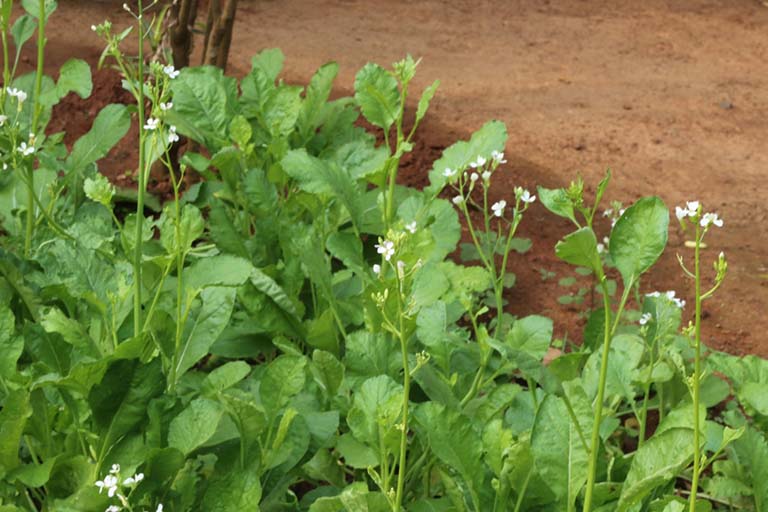}&
    \includegraphics[width=0.24\linewidth]{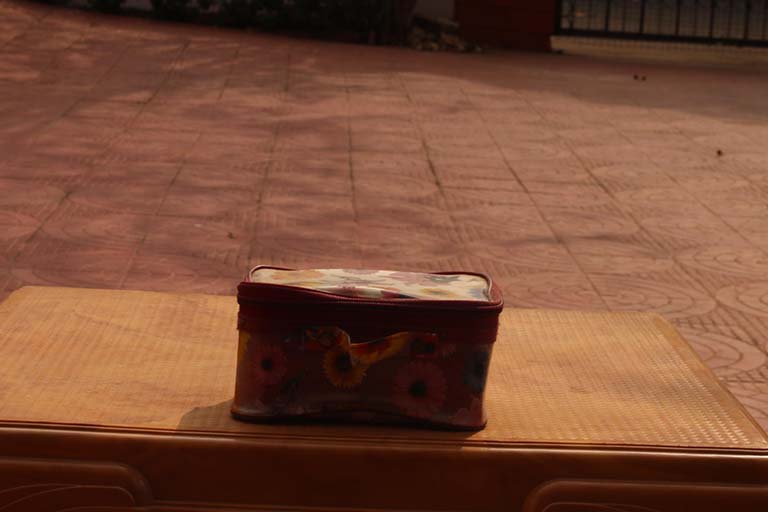}\\
    \includegraphics[width=0.24\linewidth]{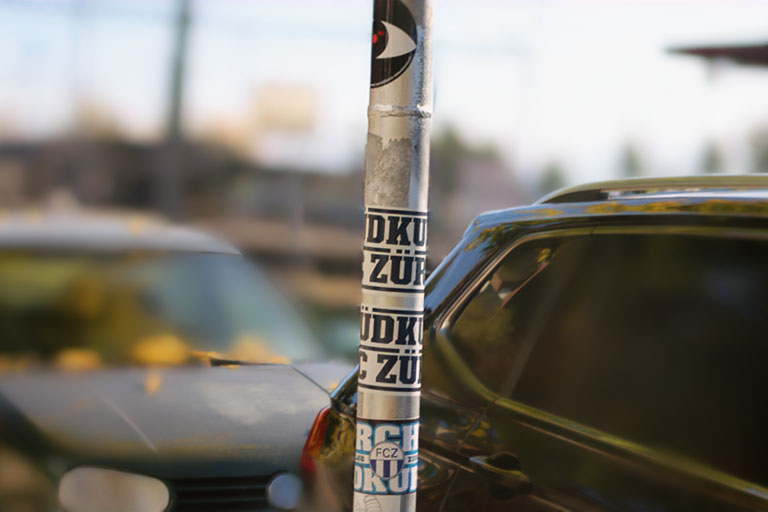}&
    \includegraphics[width=0.24\linewidth]{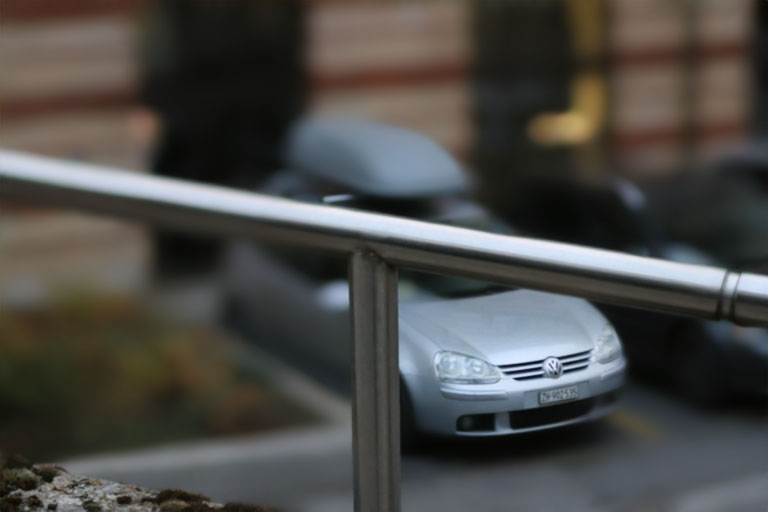}&
    \includegraphics[width=0.24\linewidth]{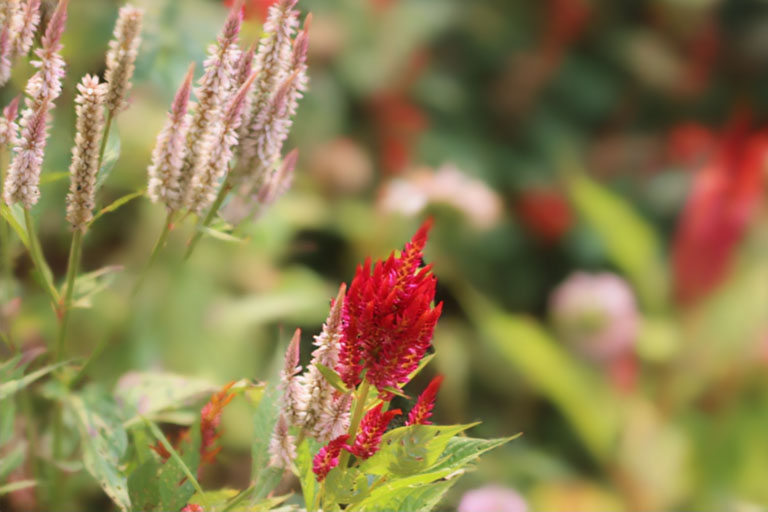}&
    \includegraphics[width=0.24\linewidth]{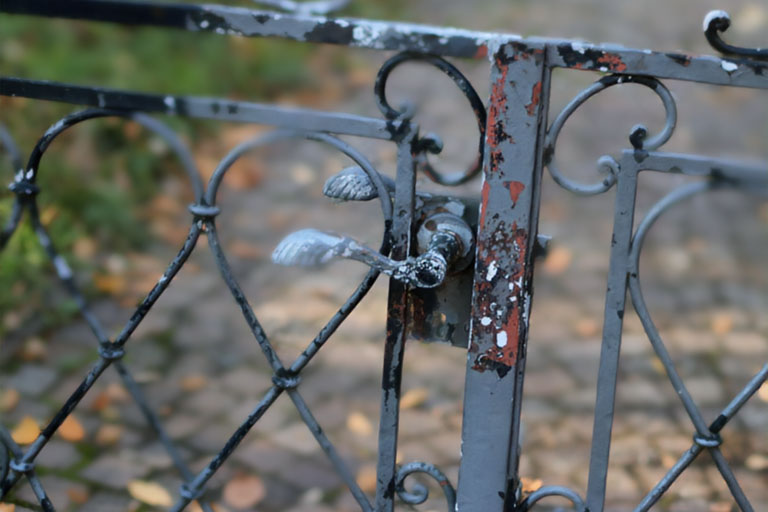}
    \includegraphics[width=0.24\linewidth]{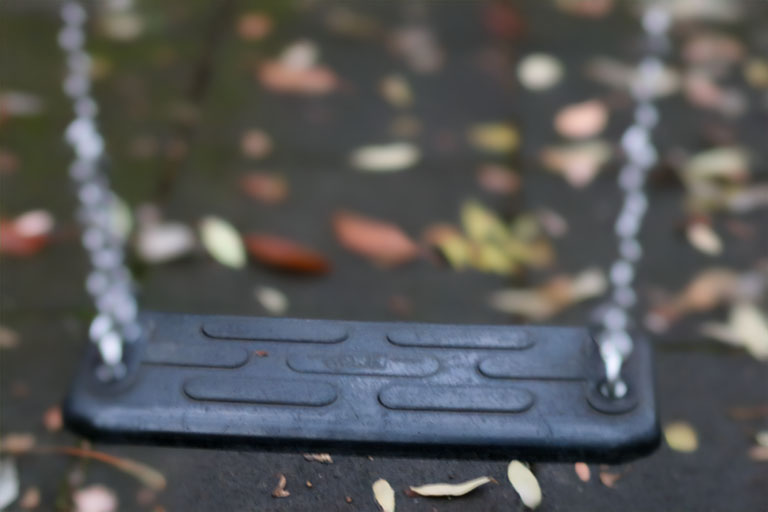}&
    \includegraphics[width=0.24\linewidth]{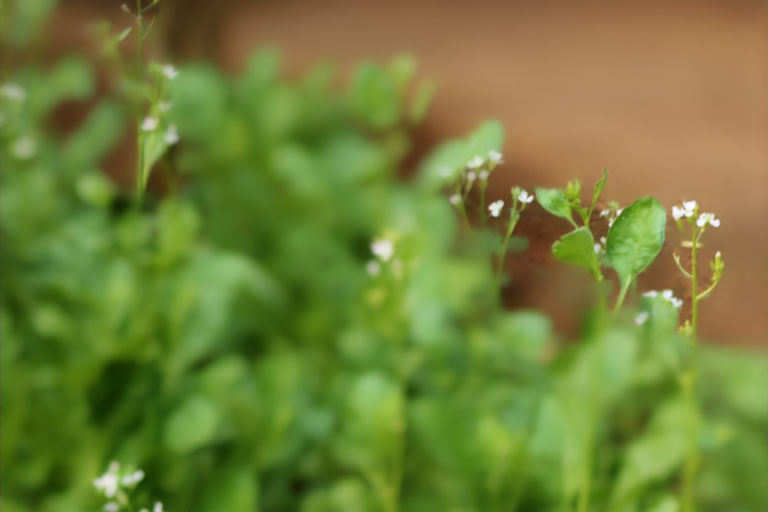}&
    \includegraphics[width=0.24\linewidth]{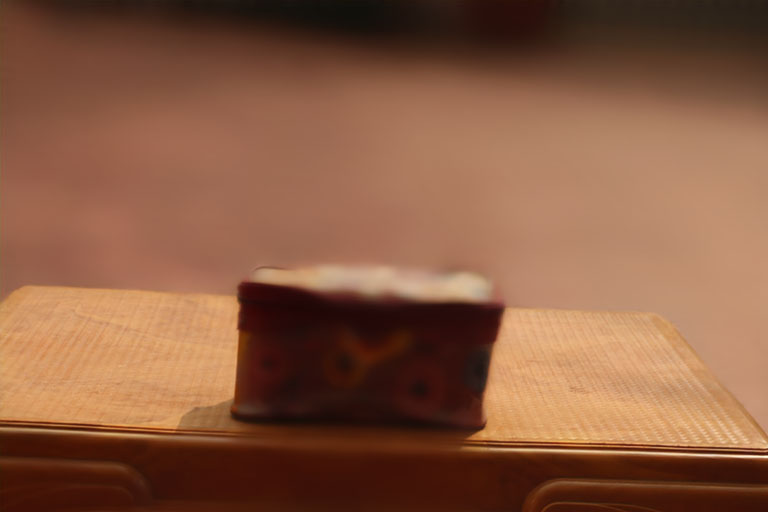}\\
\end{tabular}
}
\label{fig:artifacts}
\end{figure*}

\section{Experiments}

In this section, we evaluate the quantitative and qualitative performance of the proposed solution on the real bokeh effect rendering problem. In particular, our goal is to answer the following questions:

\begin{itemize}
\item How good is the visual quality of the rendered bokeh photos and how delicate are the transitions to the blurred areas on the images.
\item Is the network able to deal with complex image scenes, and what is the percentage of failures in such cases.
\item How well our solution performs numerically and perceptually compared to the commonly used deep learning models tuned for this problem.
\item How fast can the considered PyNET-based model run and render bokeh photos on mobile devices.
\item How well the proposed solution performs compared to the commonly used ``Portrait Mode'' in the Google Pixel Camera app that is using dual-pixel auto-focus hardware for estimating the depth map of the image.
\end{itemize}

In order to answer these questions, we performed a wide range of experiments, which results are described in detail in the following five sections.

\vspace{-1.7mm}
\subsection{Analyzing the Results}

Since the perceptual quality of the rendered bokeh images is our primary target, we start the experiments with the analysis of the obtained visual results. Figure~\ref{fig:results} shows sample bokeh images rendered with the proposed PyNET model, the original wide depth-of-field and the target Canon photos. The produced images exhibit a pronounced bokeh effect that is globally very close to the one on DSLR photos, though some local low-level details are rendered differently. The main focus subject and the strength of the blur are selected correctly based on the distance to the in-focus image plane. The bokeh on the reconstructed images looks quite natural, without any strong artifacts or corruptions at both the local and global levels. There is also no notable sharpness degradation in the in-focus image regions.

Figure~\ref{fig:sample_crops} displays the boundary regions between in and out of focus image parts. As one can see, the transitions to the blurred areas are rendered quite accurately and gently, without mixing the two regions and the halo artifacts. The problems occur only in relatively complex borderline areas where it is hard to distinguish between the fore- and background objects, which results in distinctive under- or over-blurred image segments.

According to the obtained visual results, around 5-8\% of the rendered bokeh images might contain strong visible artifacts. The problems usually happen in complex scenes with multiple objects that have either similar colors or are occluded by each other (Fig.~\ref{fig:artifacts}). In such cases, it is difficult to separate the corresponding objects without knowing the detailed semantic information or an accurate depth map estimation that can be potentially inferred from multiple cameras with the conventional stereo vision algorithms.

\subsection{Analyzing the PyNET Architecture}

\begin{figure*}[t!]
\centering
\setlength{\tabcolsep}{1pt}
\resizebox{\linewidth}{!}
{
\begin{tabular}{ccccc}
    \includegraphics[width=0.24\linewidth]{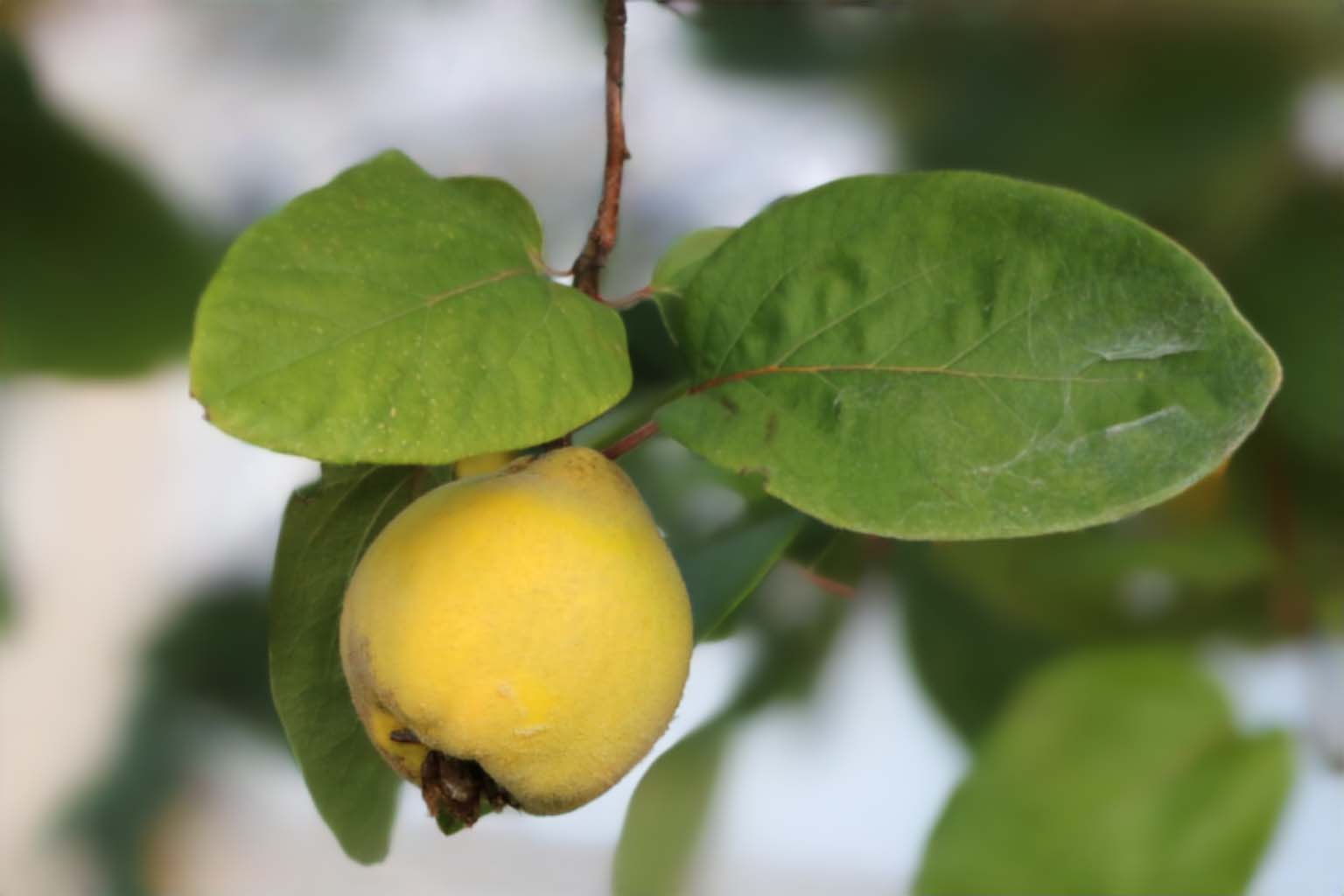}&
    \includegraphics[width=0.24\linewidth]{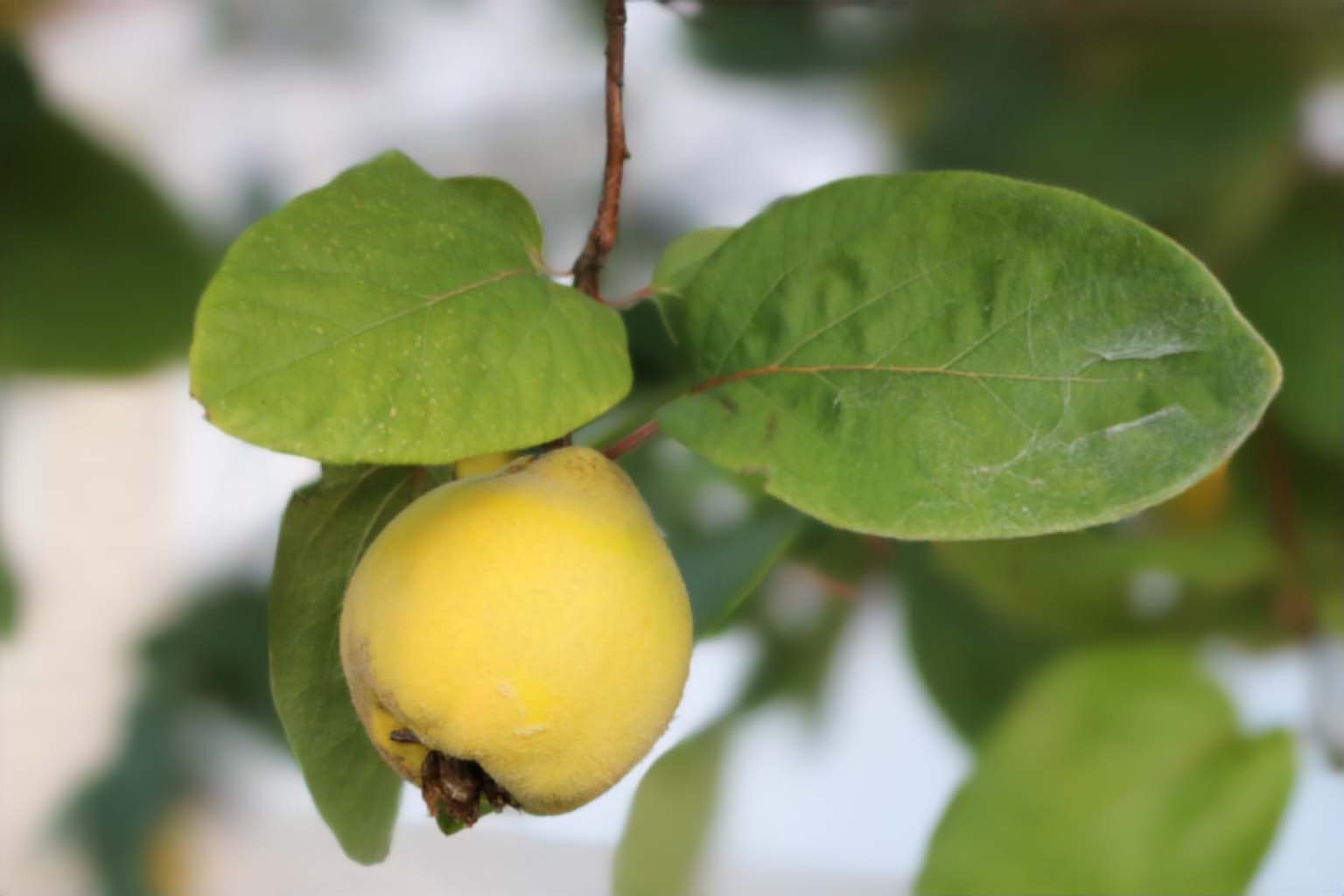}&
    \includegraphics[width=0.24\linewidth]{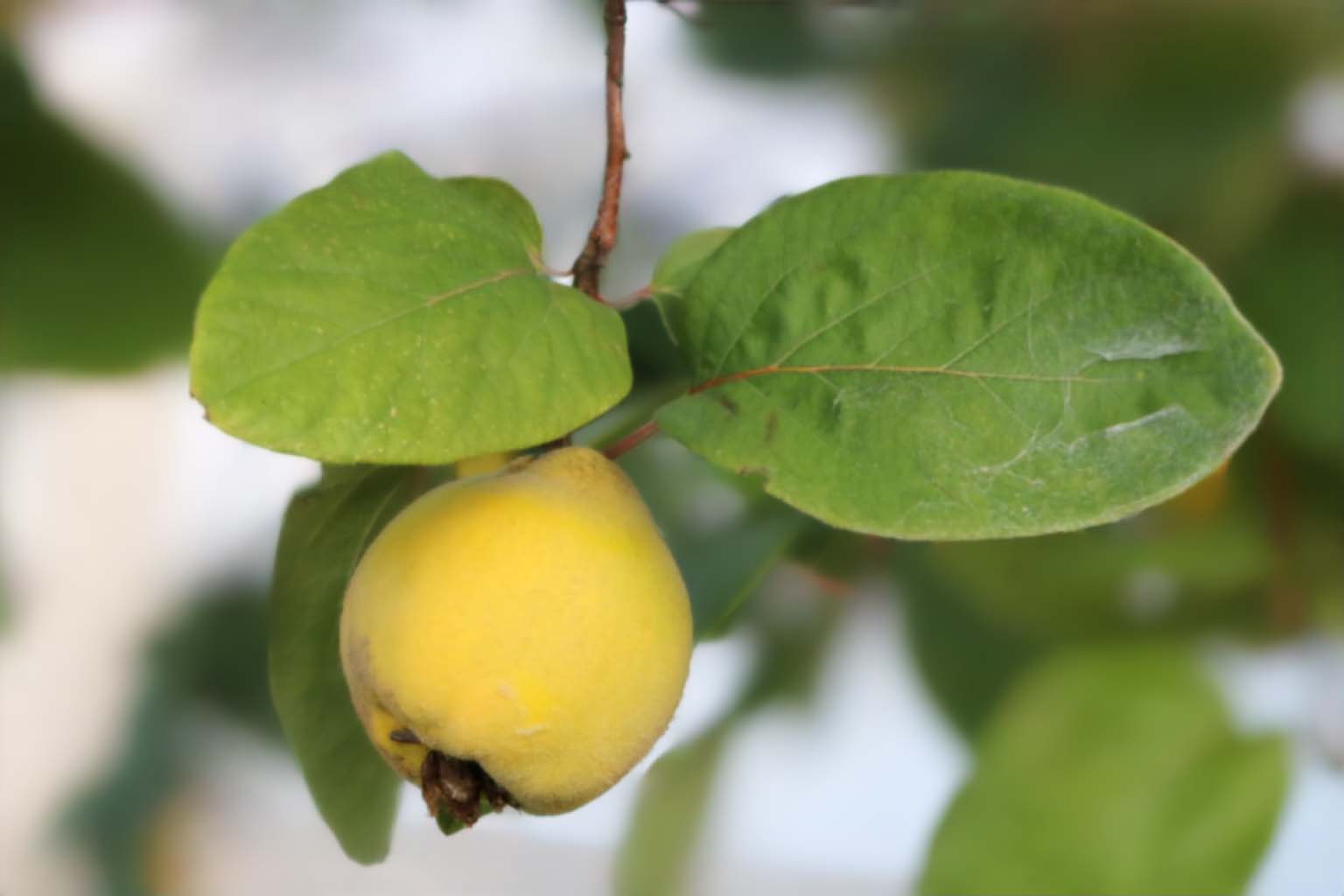}&
    \includegraphics[width=0.24\linewidth]{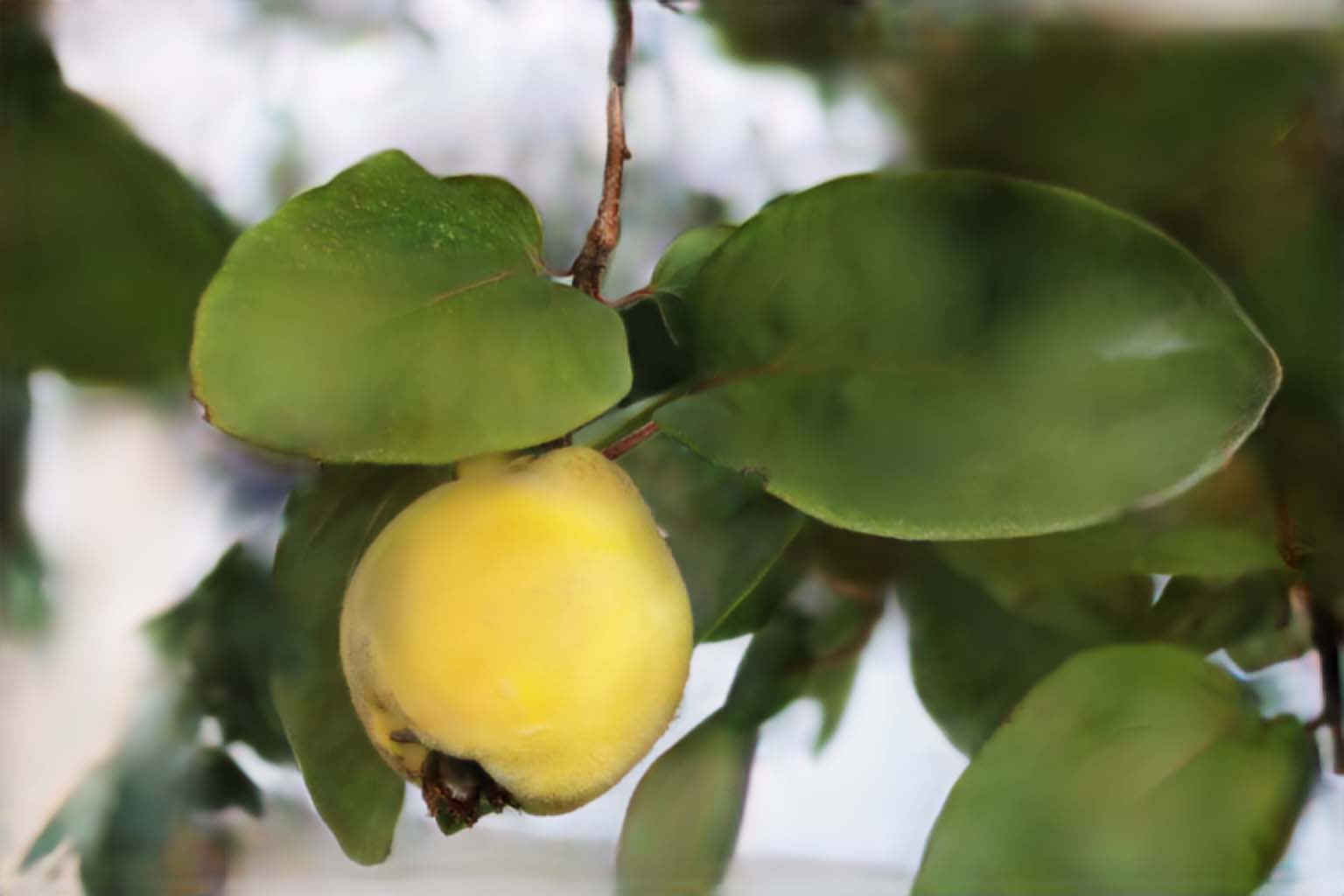}&
    \includegraphics[width=0.24\linewidth]{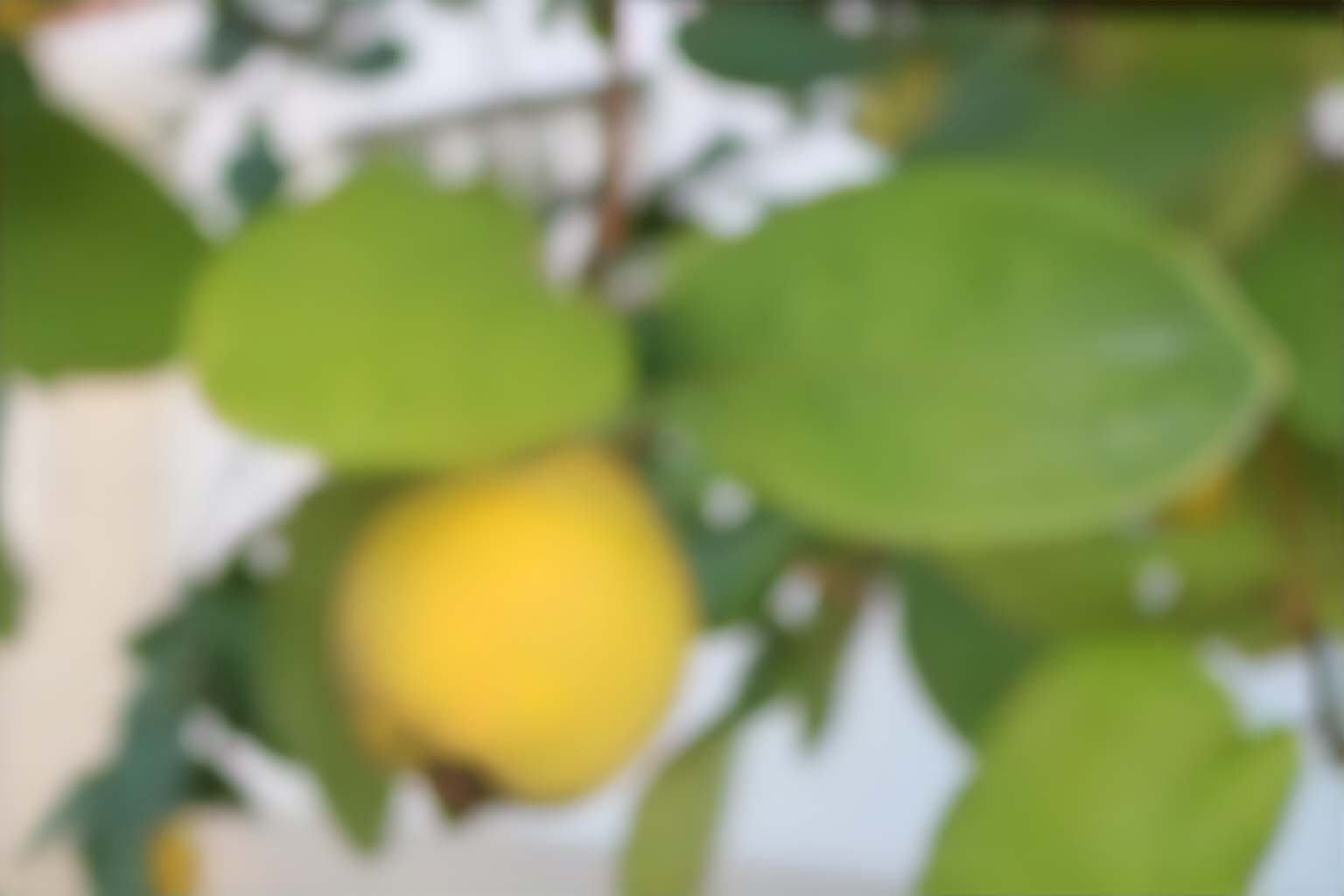}\\
    \includegraphics[width=0.24\linewidth]{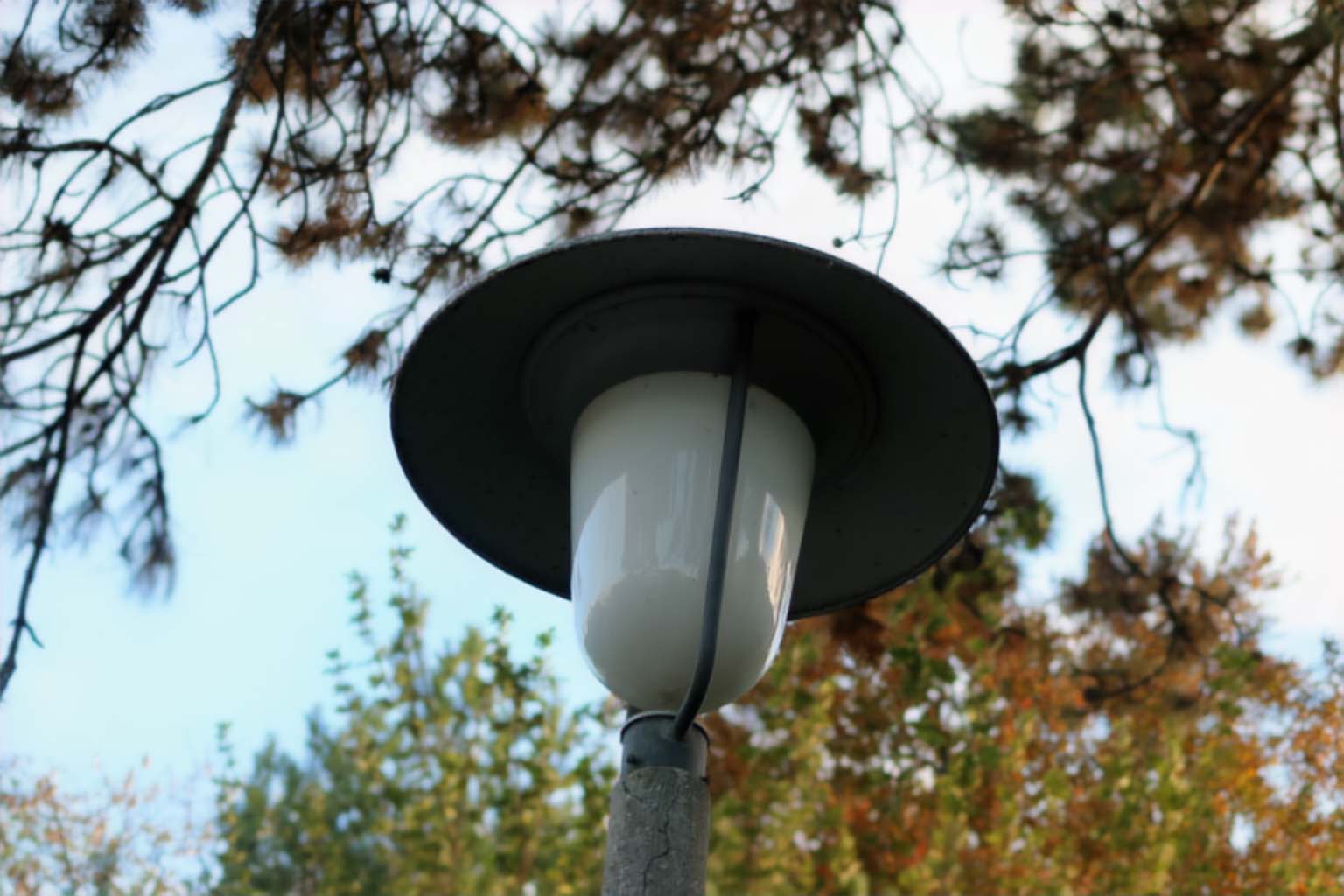}&
    \includegraphics[width=0.24\linewidth]{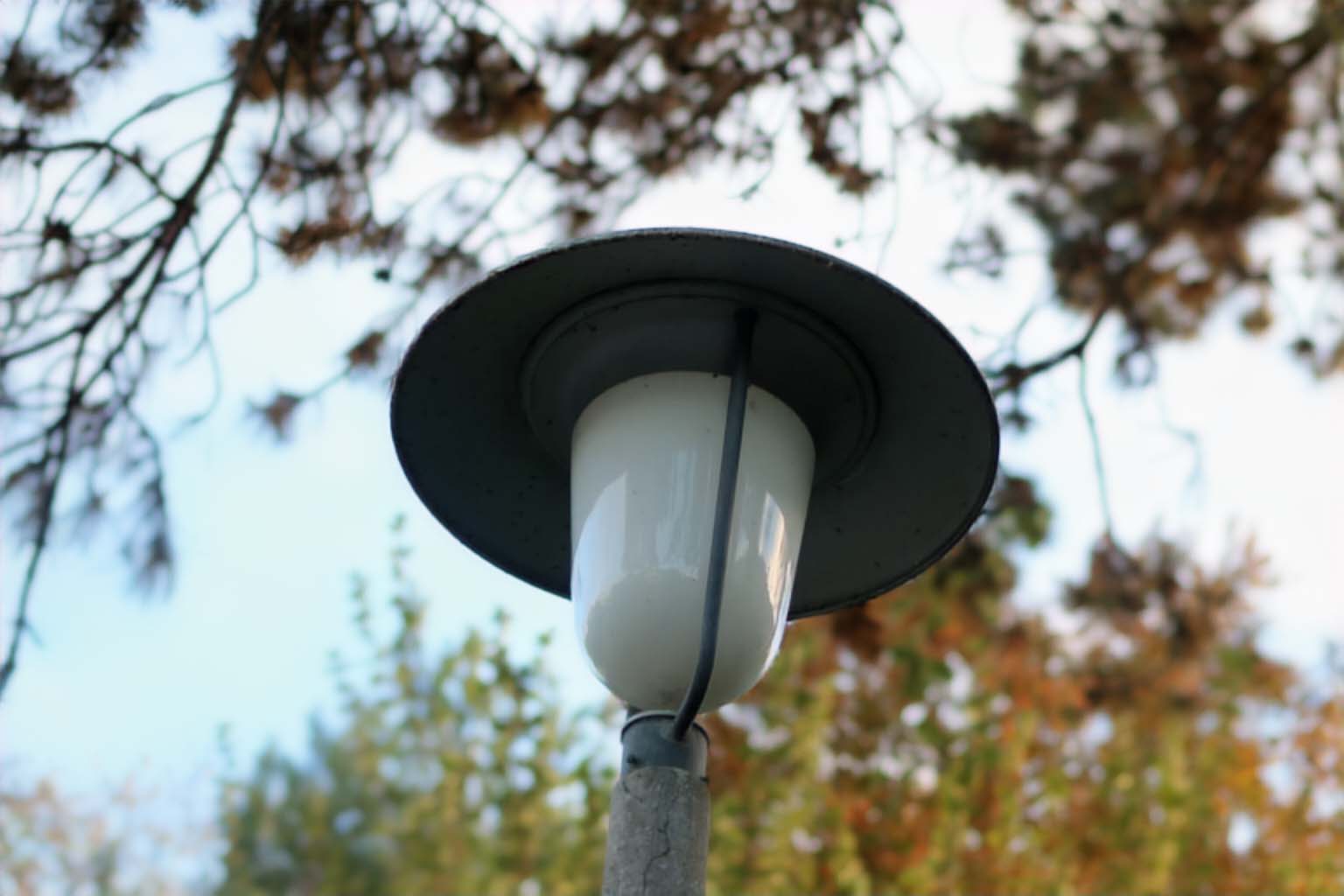}&
    \includegraphics[width=0.24\linewidth]{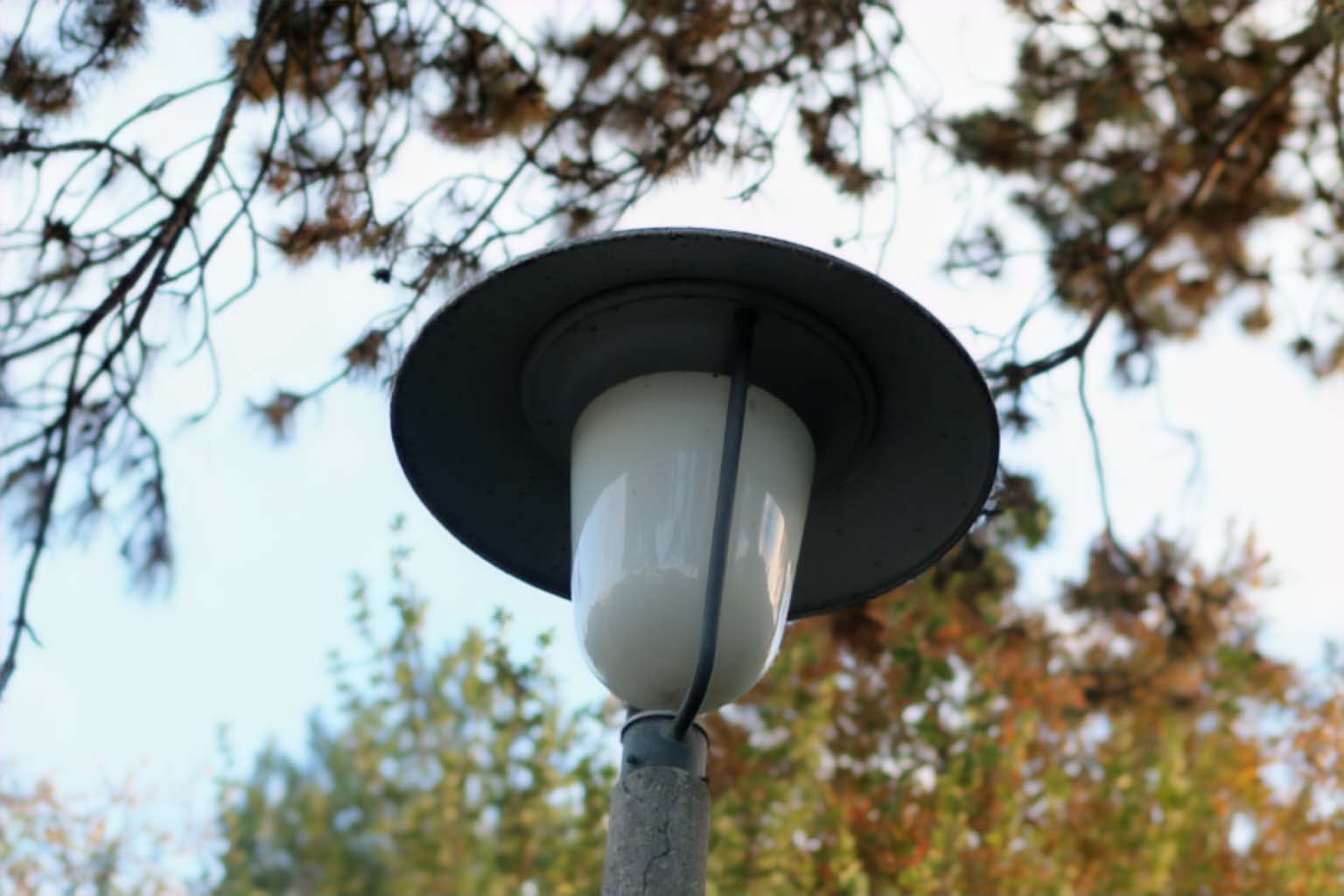}&
    \includegraphics[width=0.24\linewidth]{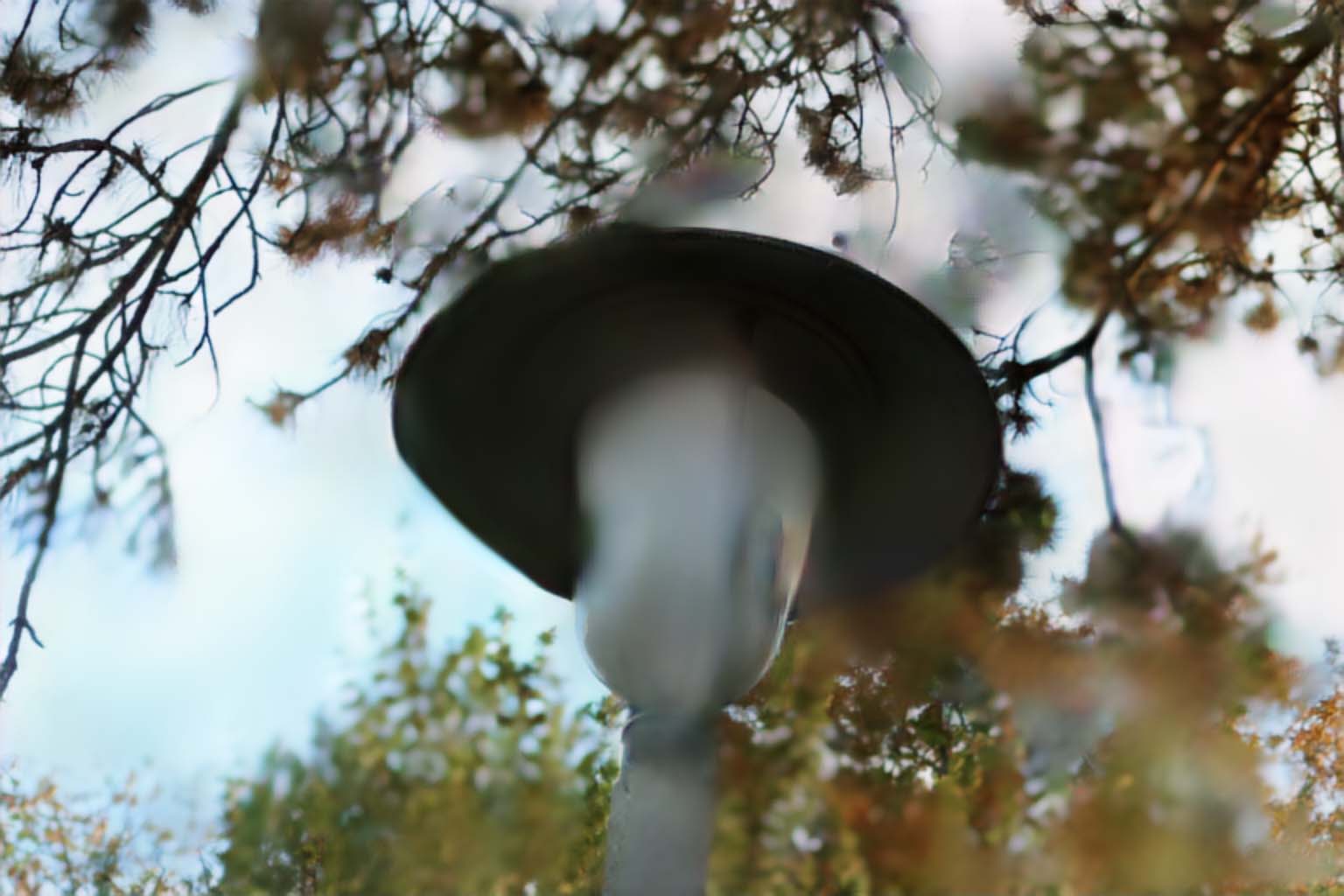}&
    \includegraphics[width=0.24\linewidth]{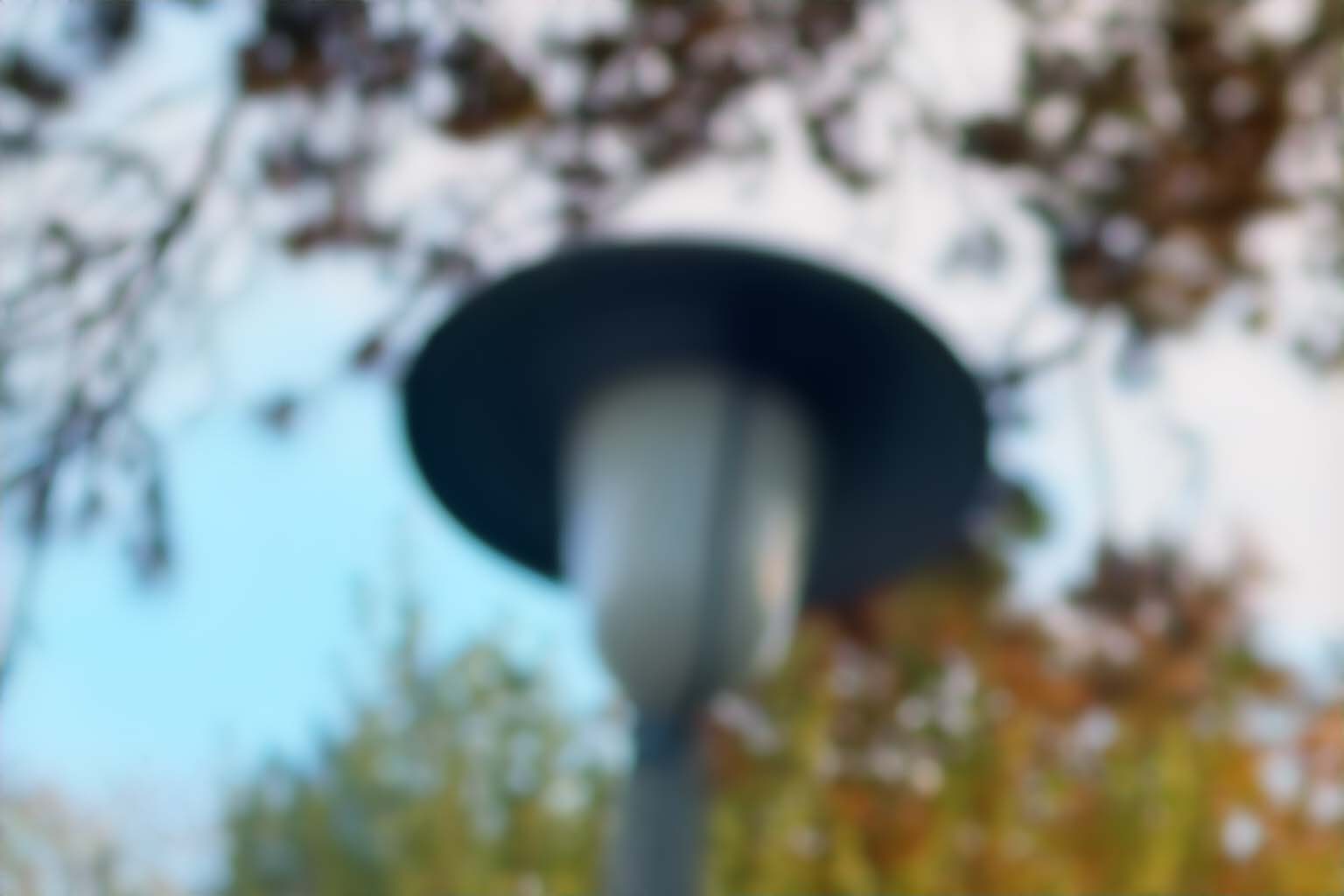}\\
a) all levels are enabled & b) level 7 is disabled & c) levels 6+ are disabled & d) levels 5+ are disabled & e) levels 4+ are disabled
\end{tabular}
}
\vspace{0.5mm}
\caption{Visual results obtained by disabling PyNET's levels. From left to right: a) all levels are enabled, b) level 7 is disabled, c) levels 6+ are disabled, d) levels 5+ are disabled, e) levels 4+ are disabled. Best zoomed on screen.}
\label{fig:levels_impact}
\vspace{-0.7mm}
\end{figure*}

\begin{table*}[b!]
\centering
\resizebox{\linewidth}{!}
{
\begin{tabular}{l|ccc|ccc|c}
\hline
Team \, & \, Hardware, GPU \, & \, Runtime, s \, & \, Normalized Runtime, s \, & \, PSNR$\uparrow$ \, & \, SSIM$\uparrow$ \, & \, LPIPS$\downarrow$ \, & \, MOS$\downarrow$ \\
\hline
\hline
PyNET & Nvidia Tesla V100 & 0.14 & 0.14 & 23.28 & 0.8780 & 0.2438 & \textBF{0.85$_{(1)}$} \\
Zheng~\etal~\cite{ignatov2019aim} & GeForce GTX 2080 Ti & 0.27 & 0.26 & 23.44 & 0.8874 & 0.2452 & 0.87$_{(2)}$ \\
Dutta~\etal~\cite{ignatov2019aim} & GeForce GTX 1080 Ti & 0.8 & 0.57 & 22.14 & 0.8633 & 0.2811 & 0.88$_{(3)}$ \\
Purohit~\etal~\cite{purohit2019depth} & Nvidia TITAN X & 2.5 & 1.15 & 23.62 & 0.8836 & \textBF{0.2248} & 0.93$_{(4)}$ \\
Xiong~\etal~\cite{ignatov2019aim} & GeForce GTX 1080 Ti & 0.89 & 0.63 & \textBF{23.93} & \textBF{0.8917} & 0.2300 & 0.98$_{(5)}$ \\
Yang~\etal~\cite{ignatov2019aim} & Nvidia TITAN & 0.6 & 0.15 & 23.18 & 0.8851 & 0.2467 & 1.02$_{(6)}$ \\
\end{tabular}
}
\vspace{0.4mm}
\caption{\small{The results on the EBB! test subset obtained with different solutions. The runtime of all methods was normalized based on the aggregated GPU performance index reported at: \url{http://ai-benchmark.com/ranking_deeplearning.html}}}
\label{tab:results}
\end{table*}

When analyzing the obtained image results, one of the key questions that we had is the role of different PyNET levels in rendering the final bokeh image. To understand this, an experiment was conducted where we were disabling PyNET levels from the 7th (lowest) to the 4th one by setting their outputs / activations to zeros, and observing the changes in the produced images. The corresponding results are presented in Fig.~\ref{fig:levels_impact}. It turned out that this is one of the rare cases when it is possible to directly identify the role of different layers in data processing and generation of the reconstructed image. We provide the description of our findings below:
\smallskip

\noindent \textit{Level 7} is responsible for the overall image brightness, blur strength in the background areas and blur shift. When disabling this level, the final image results do not change significantly except for a slightly shifted and less / more blurred background, and a bit different global image brightness (Fig.~\ref{fig:levels_impact}, b).
\smallskip

\noindent \textit{Level 6} is generally refining the outputs obtained from level 7, the same changes can be also observed here (Fig.~\ref{fig:levels_impact}, c).
\smallskip

\noindent \textit{Level 5} produces a coarse blur mask that is defining what image areas should be blurred and what regions should stay in focus (Fig.~\ref{fig:levels_impact}, d).
\smallskip

\noindent \textit{Level 4} provides a detailed refined blur mask that is used by the upper level to generate the actual results. When disabling this level, the network produces images that are completely blurred (Fig.~\ref{fig:levels_impact}, d).
\smallskip

\noindent \textit{Level 3} is responsible for generating the actual bokeh effect. When it does not receive any information from the lower layer, it is just blurring the entire image with the default blur (Fig.~\ref{fig:levels_impact}, d). We should emphasize that the out-of-focus areas on the final images (Fig.~\ref{fig:levels_impact}, a) are looking quite different from the ones on the totally blurred images, which shows that the type and strength of the bokeh effect is also learned and provided by levels 4 and 5.
\smallskip

The obtained results are generally following our expectations described in Section~\ref{sec:pynet}. A more advanced study might also reveal how one can alter the features from the lower layers to change the final reconstructed image in a predictable way, which can be used, \eg, for explicitly adjusting the strength of the bokeh effect.

\subsection{Quantitative and Qualitative Evaluation}

\begin{figure*}[t!]
\vspace{-7mm}
\centering
\setlength{\tabcolsep}{1pt}
\resizebox{\linewidth}{!}
{
\begin{tabular}{cccccccc}
    \includegraphics[width=0.24\linewidth]{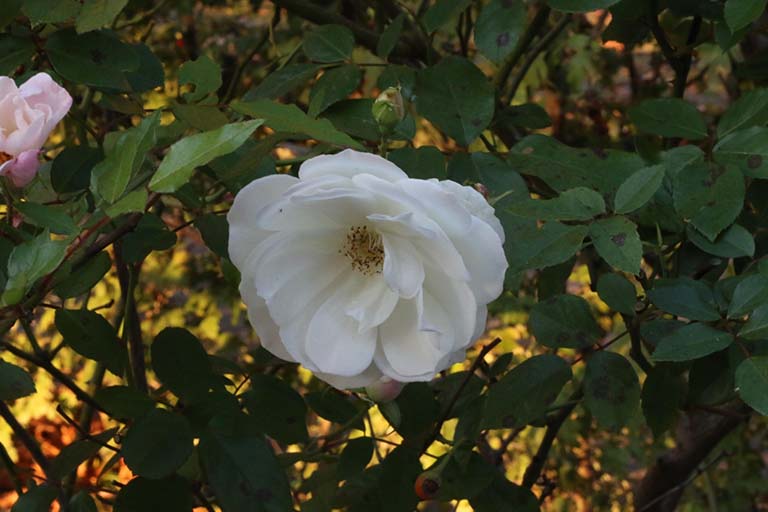}&
    \includegraphics[width=0.24\linewidth]{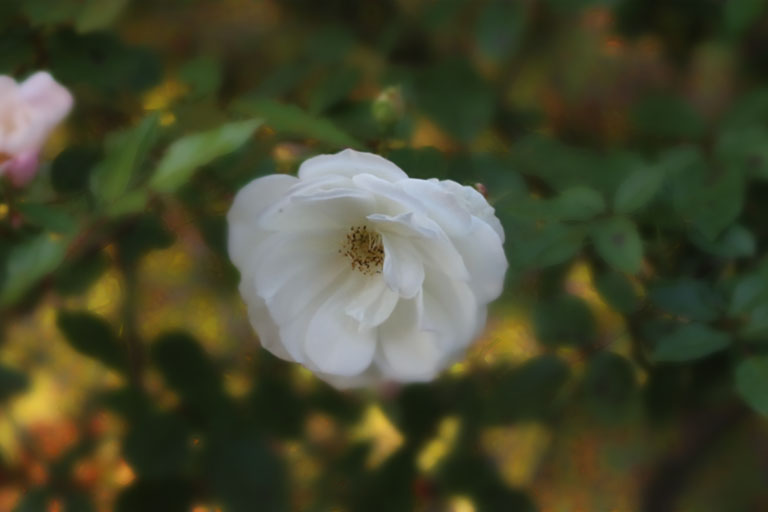}&
    \includegraphics[width=0.24\linewidth]{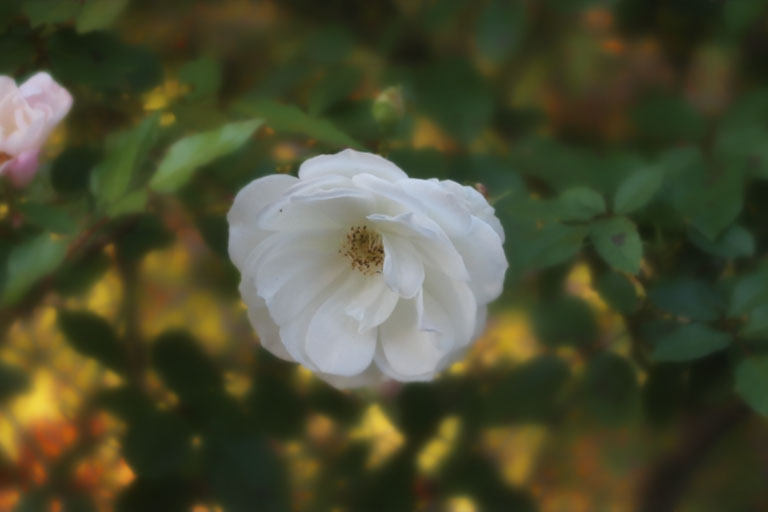}&
    \includegraphics[width=0.24\linewidth]{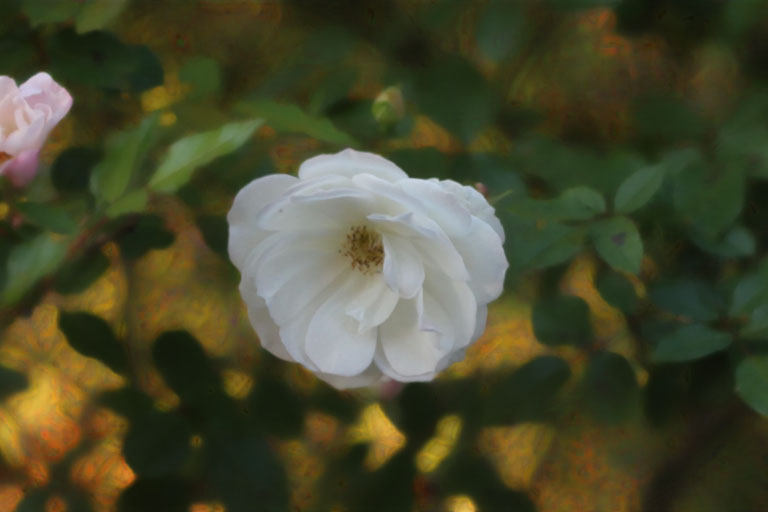}&
    \includegraphics[width=0.24\linewidth]{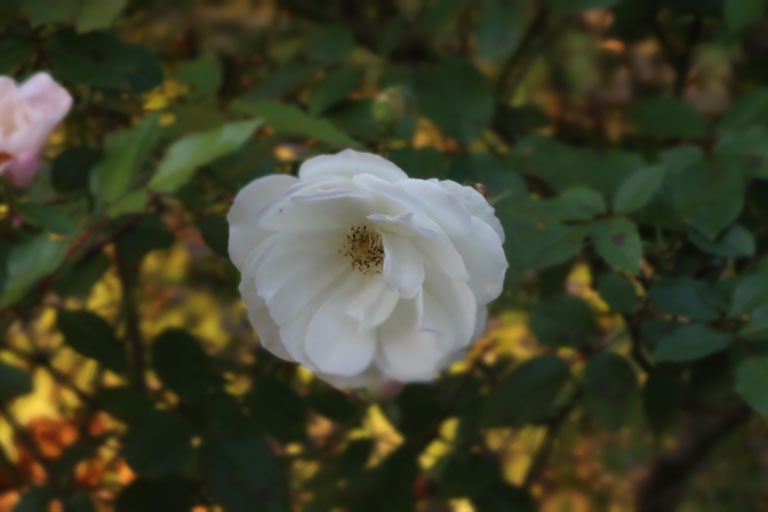}&
    \includegraphics[width=0.24\linewidth]{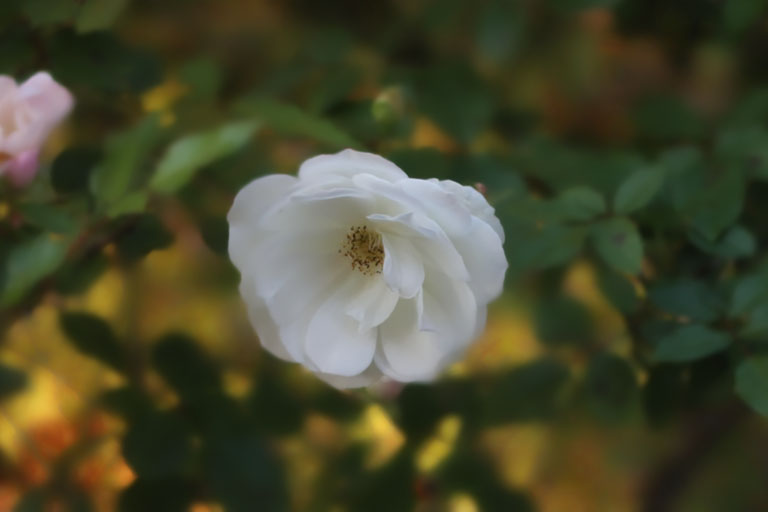}&
    \includegraphics[width=0.24\linewidth]{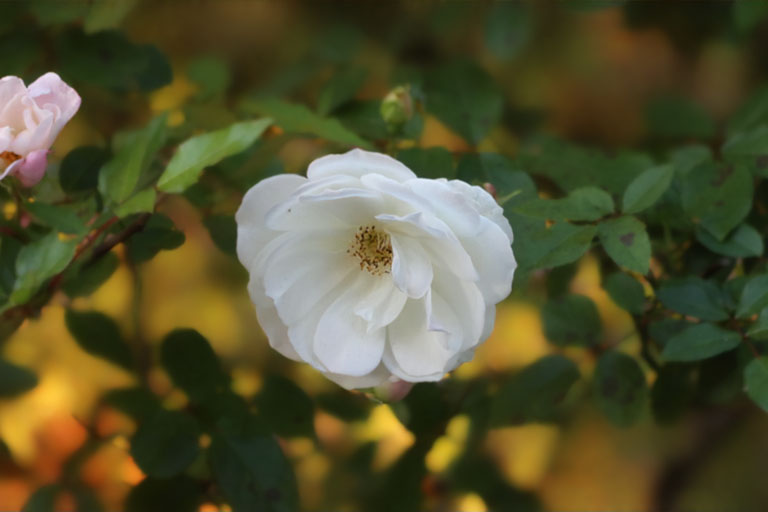}&
    \includegraphics[width=0.24\linewidth]{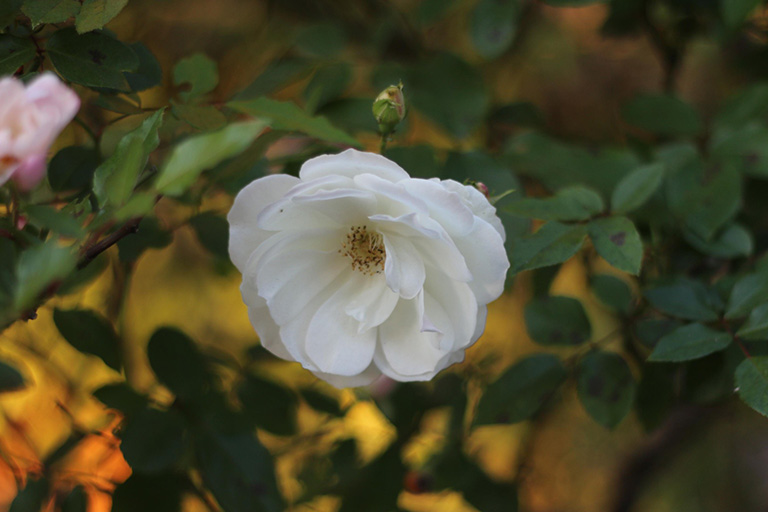}\\
    \includegraphics[width=0.24\linewidth]{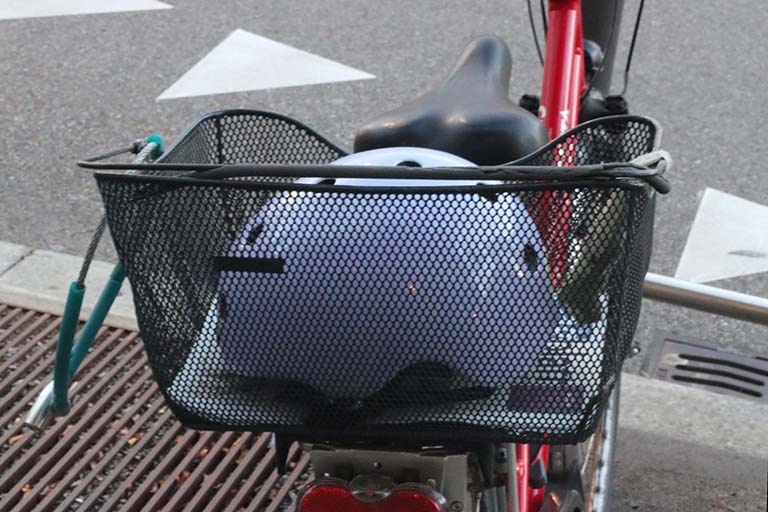}&
    \includegraphics[width=0.24\linewidth]{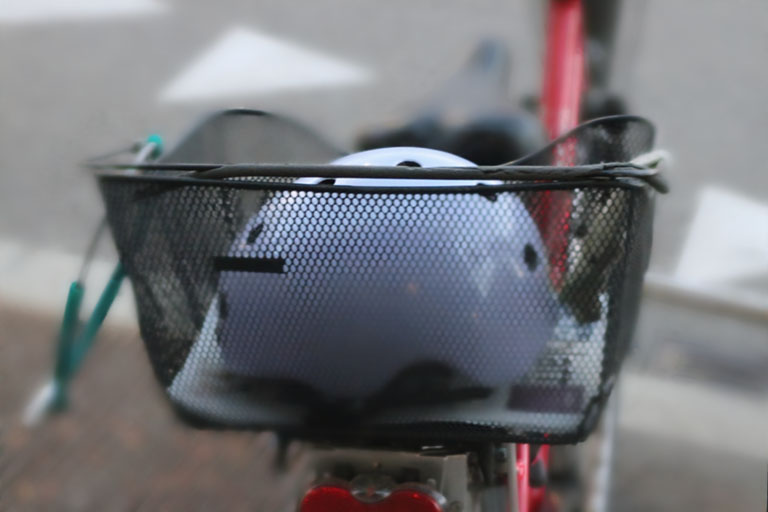}&
    \includegraphics[width=0.24\linewidth]{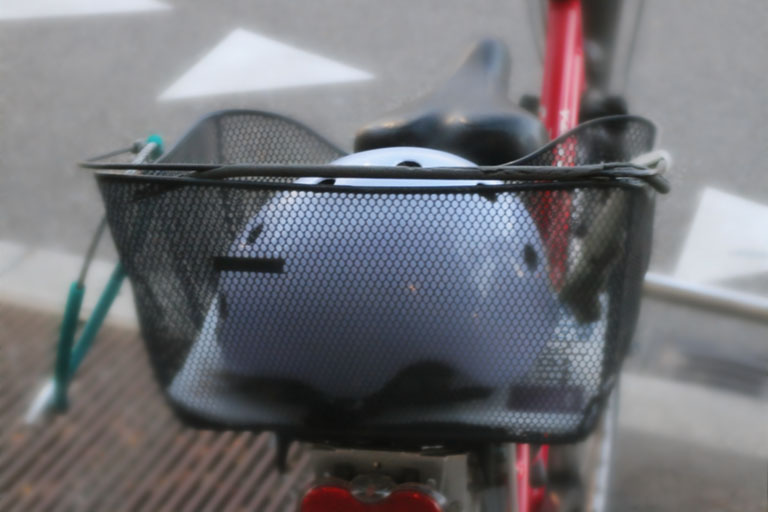}&
    \includegraphics[width=0.24\linewidth]{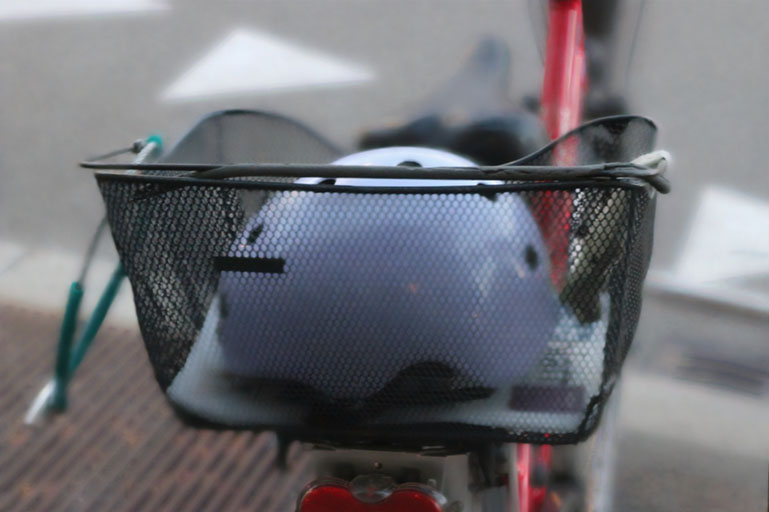}&
    \includegraphics[width=0.24\linewidth]{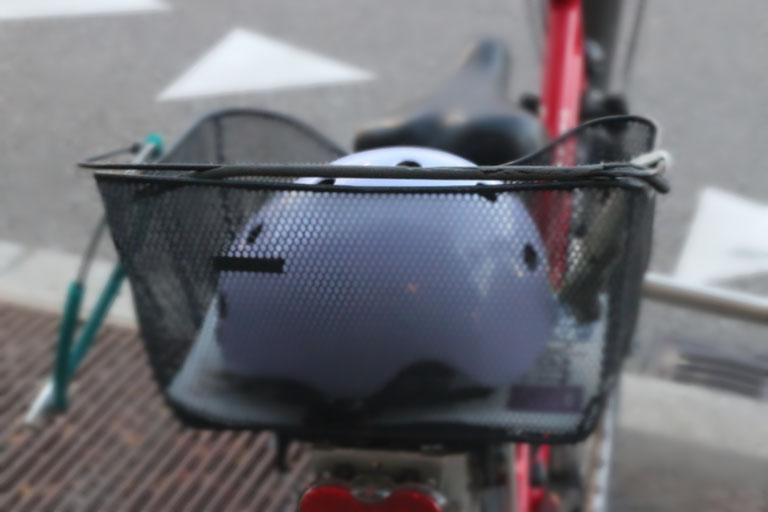}&
    \includegraphics[width=0.24\linewidth]{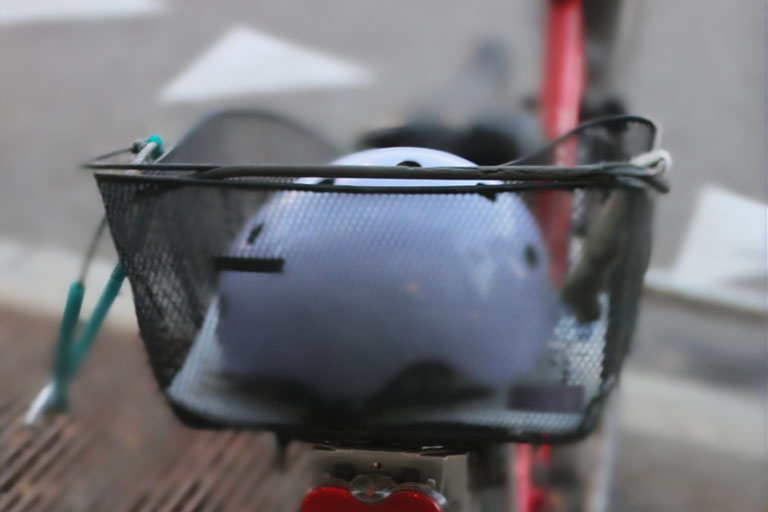}&
    \includegraphics[width=0.24\linewidth]{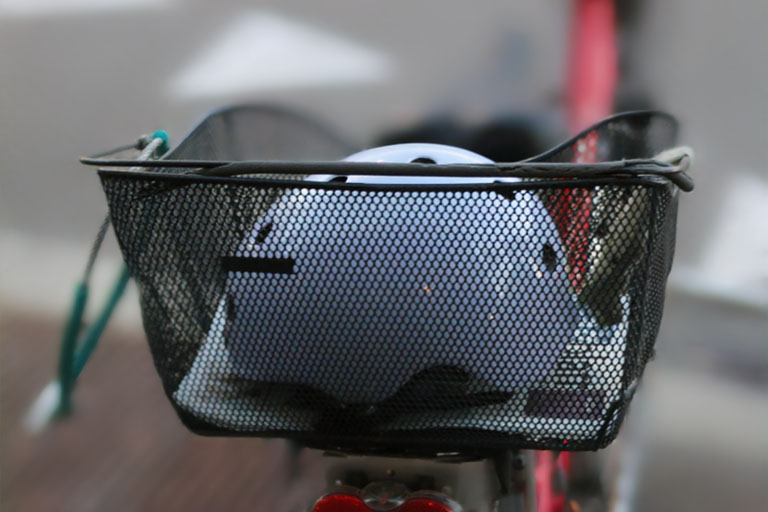}&
    \includegraphics[width=0.24\linewidth]{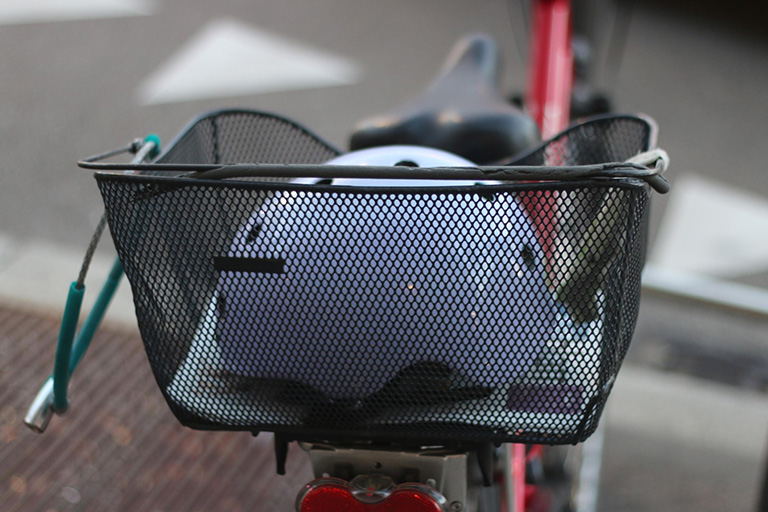}\\
    \includegraphics[width=0.24\linewidth]{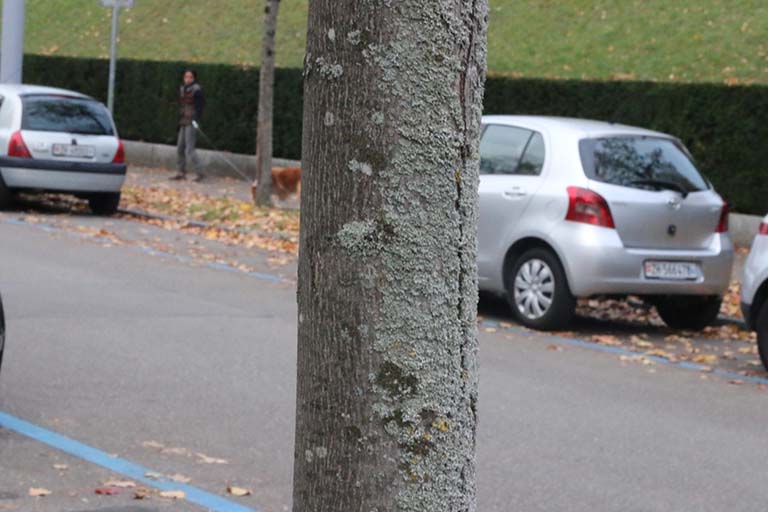}&
    \includegraphics[width=0.24\linewidth]{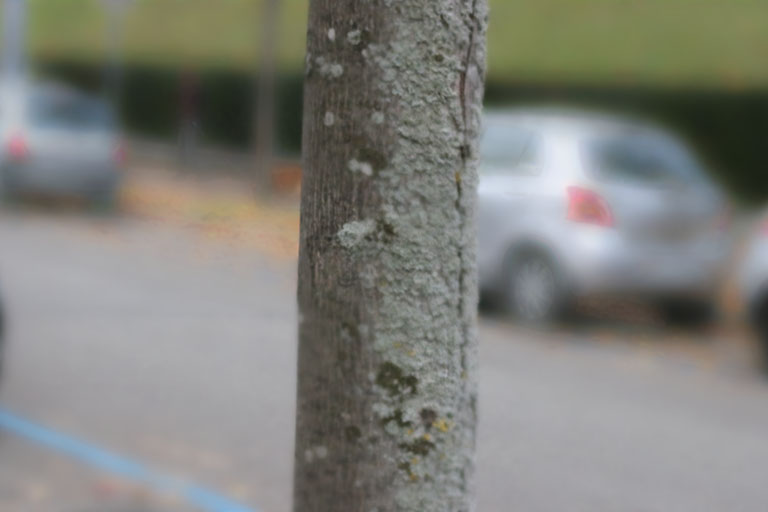}&
    \includegraphics[width=0.24\linewidth]{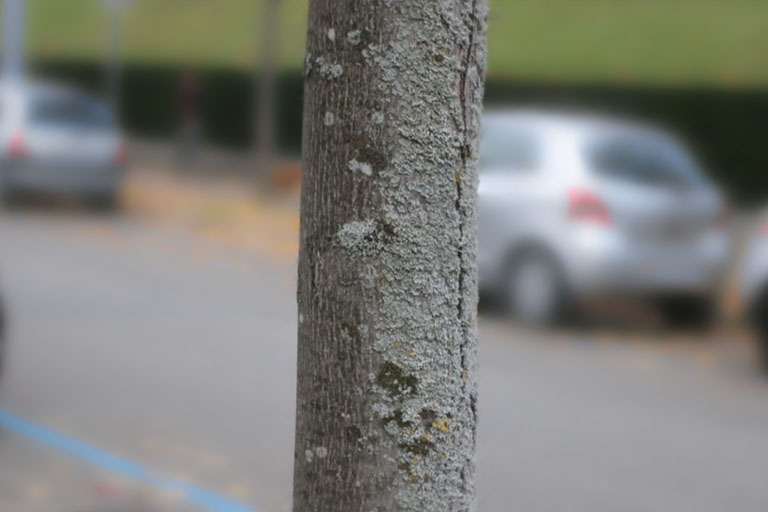}&
    \includegraphics[width=0.24\linewidth]{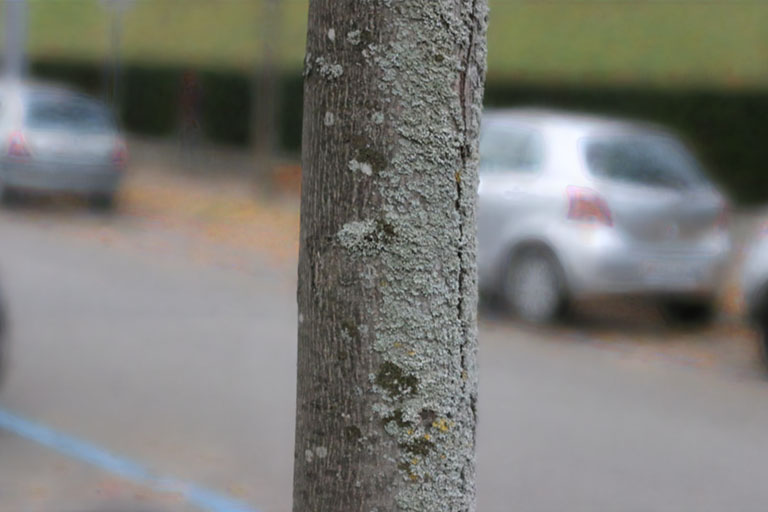}&
    \includegraphics[width=0.24\linewidth]{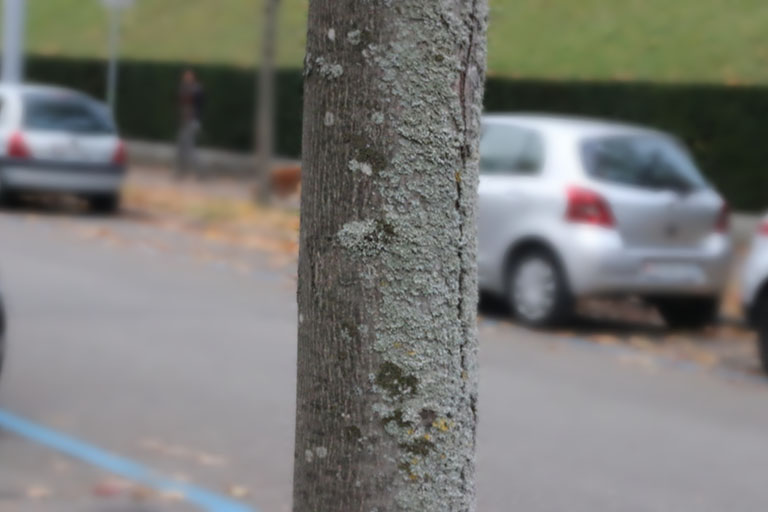}&
    \includegraphics[width=0.24\linewidth]{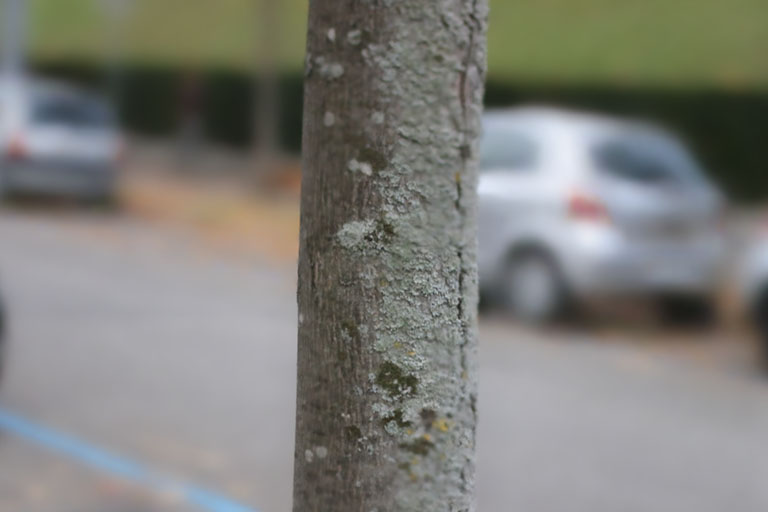}&
    \includegraphics[width=0.24\linewidth]{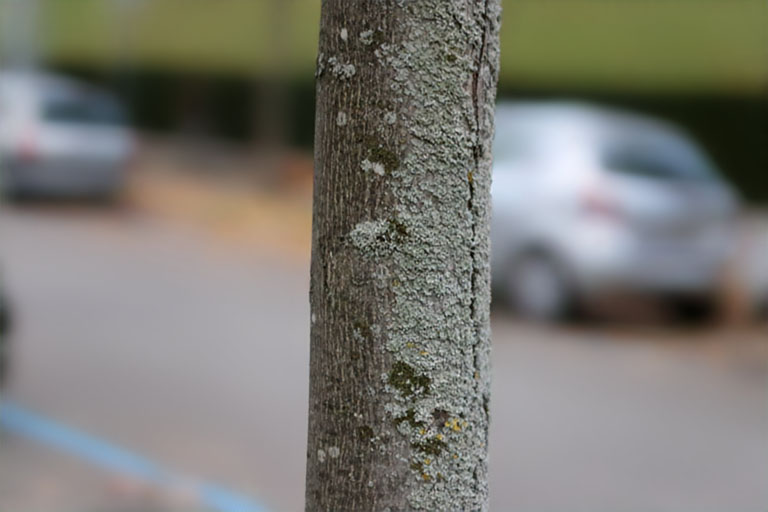}&
    \includegraphics[width=0.24\linewidth]{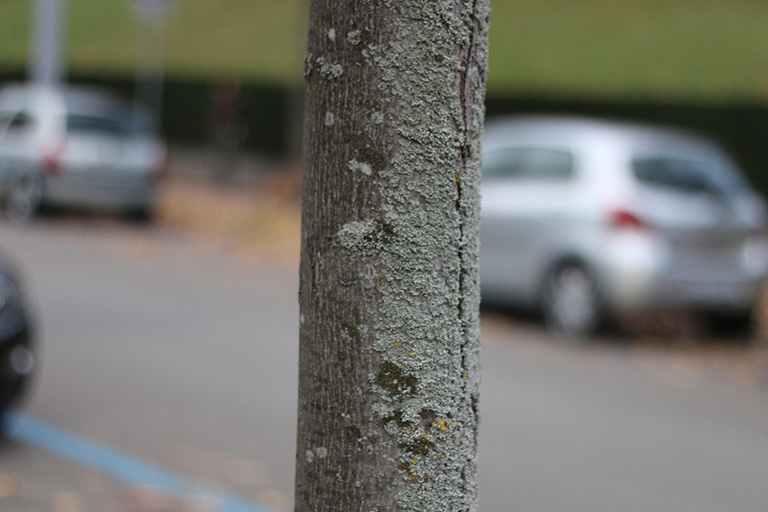}\\
    \includegraphics[width=0.24\linewidth]{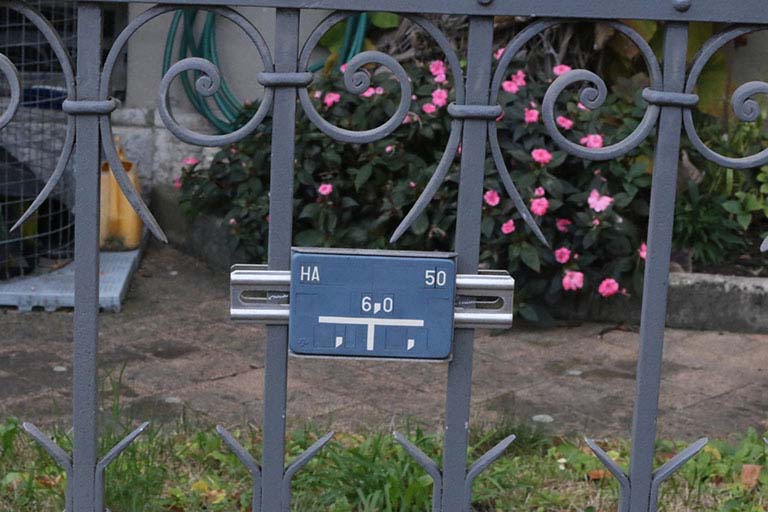}&
    \includegraphics[width=0.24\linewidth]{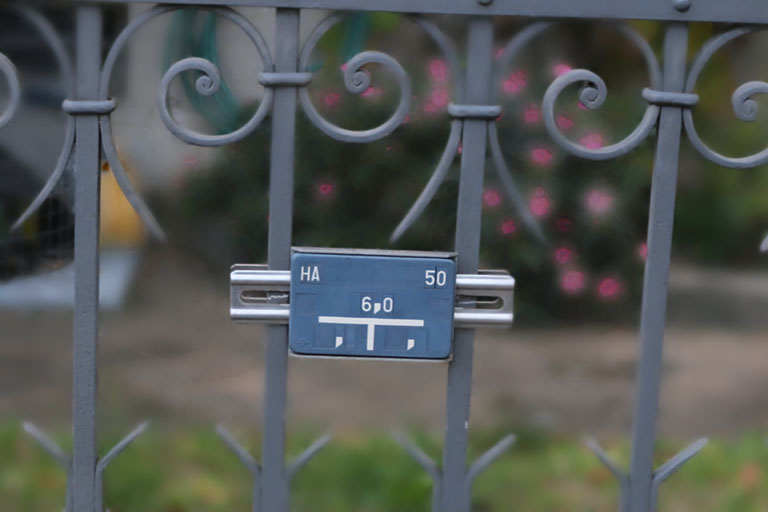}&
    \includegraphics[width=0.24\linewidth]{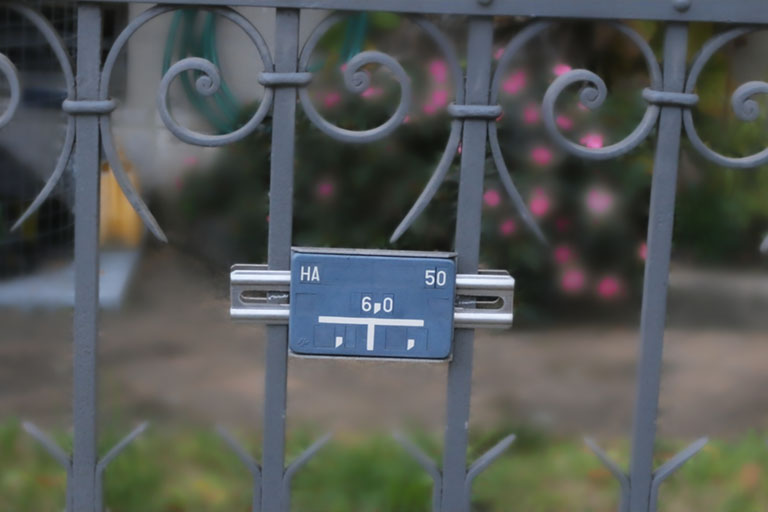}&
    \includegraphics[width=0.24\linewidth]{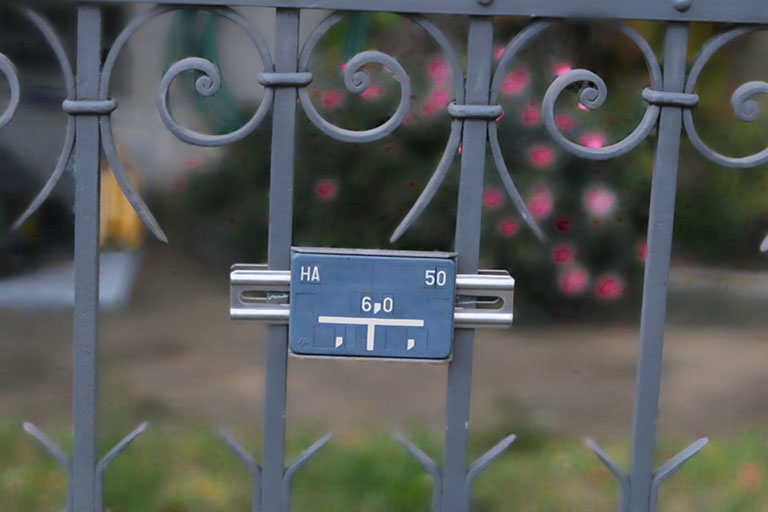}&
    \includegraphics[width=0.24\linewidth]{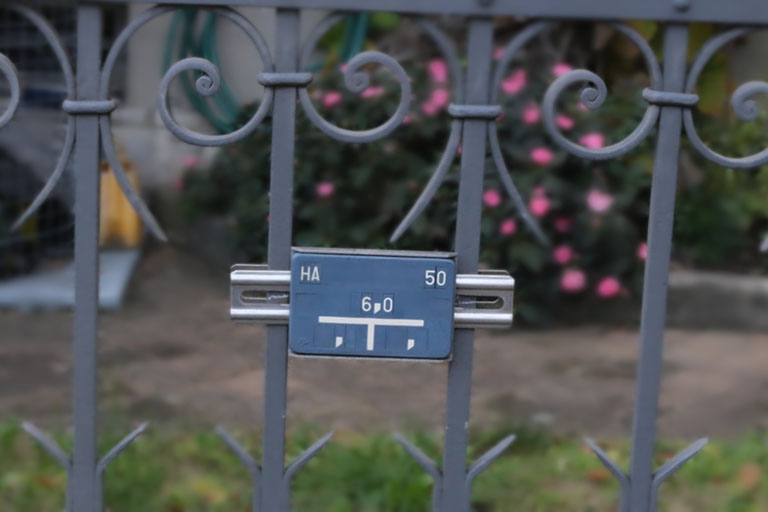}&
    \includegraphics[width=0.24\linewidth]{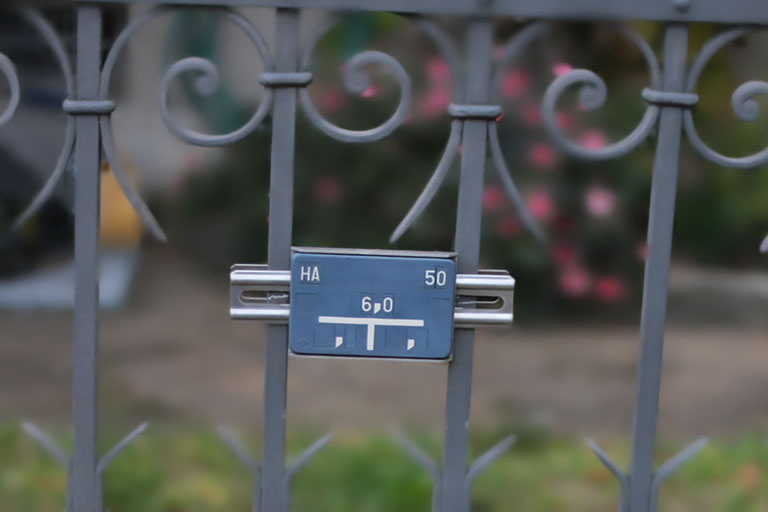}&
    \includegraphics[width=0.24\linewidth]{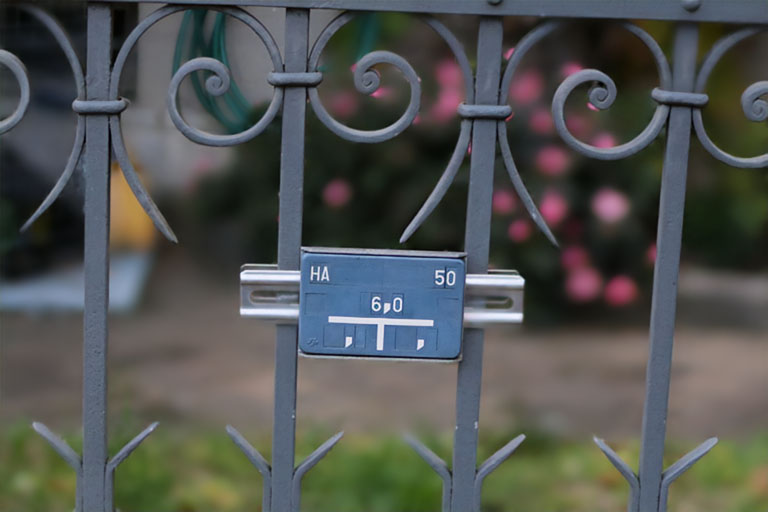}&
    \includegraphics[width=0.24\linewidth]{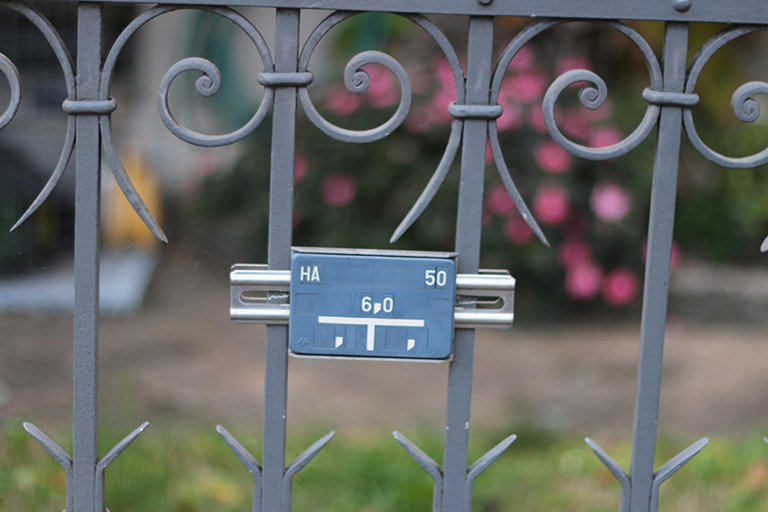}\\
    \includegraphics[width=0.24\linewidth]{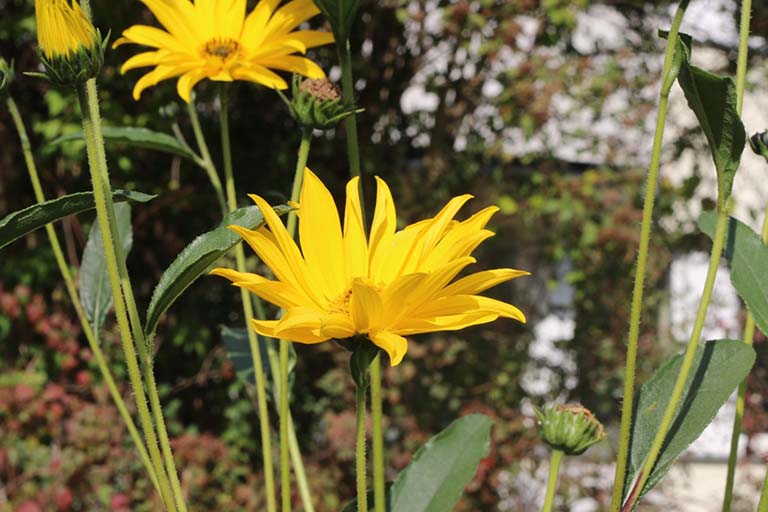}&
    \includegraphics[width=0.24\linewidth]{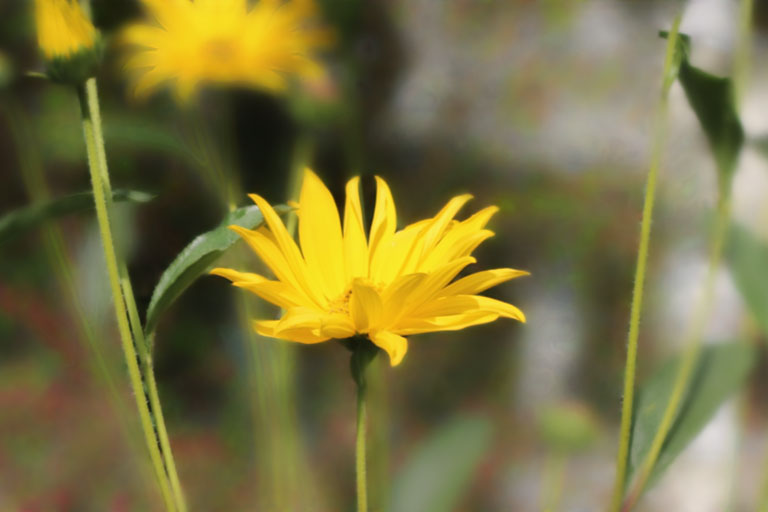}&
    \includegraphics[width=0.24\linewidth]{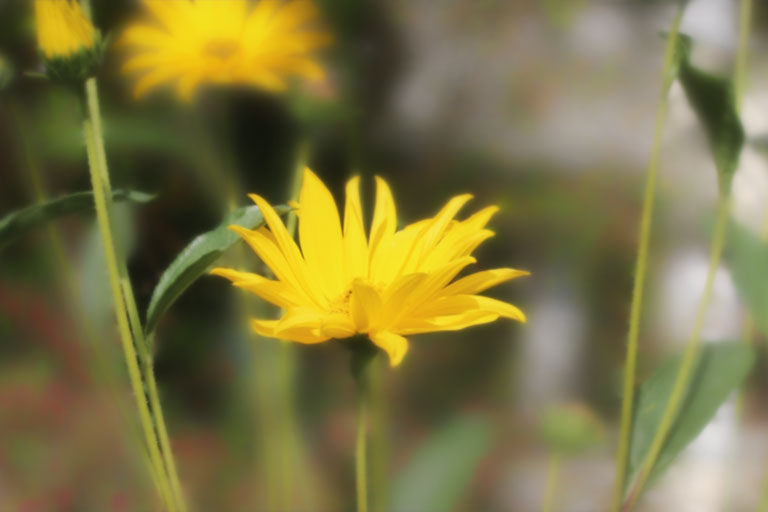}&
    \includegraphics[width=0.24\linewidth]{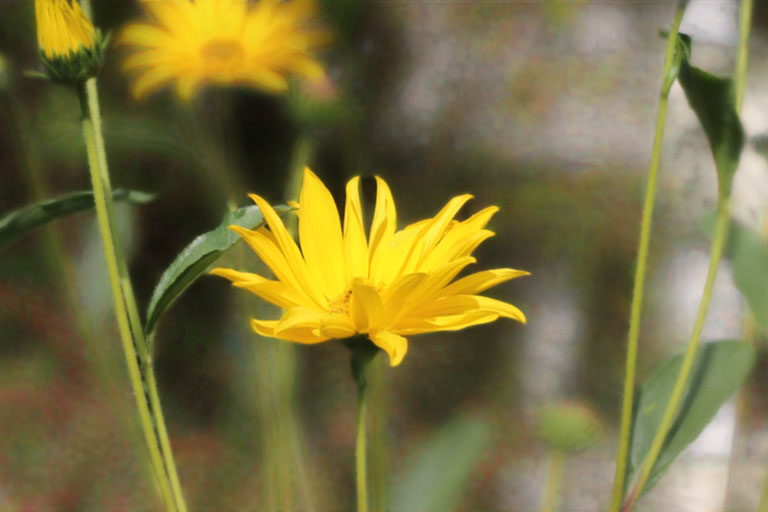}&
    \includegraphics[width=0.24\linewidth]{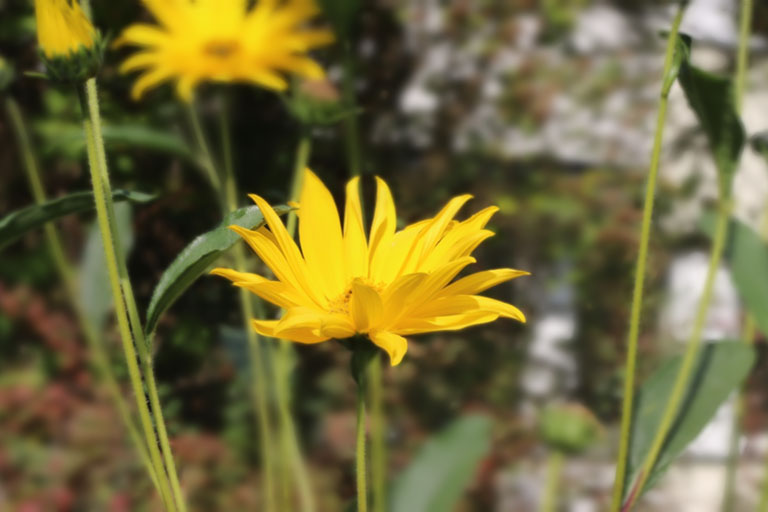}&
    \includegraphics[width=0.24\linewidth]{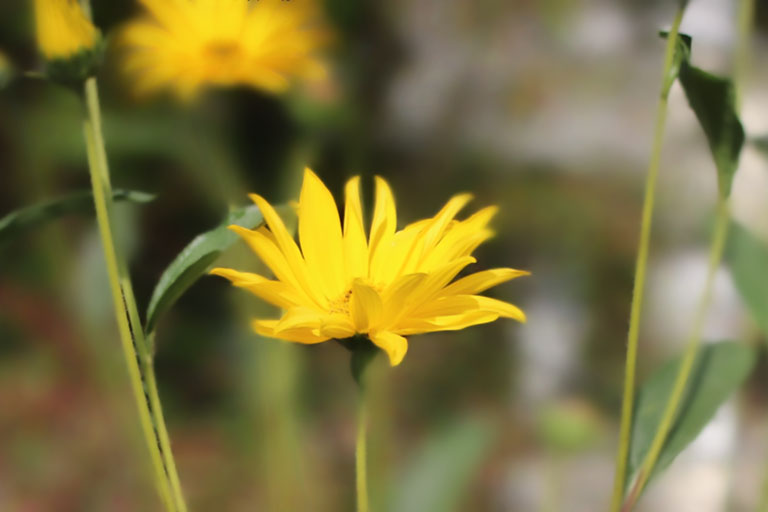}&
    \includegraphics[width=0.24\linewidth]{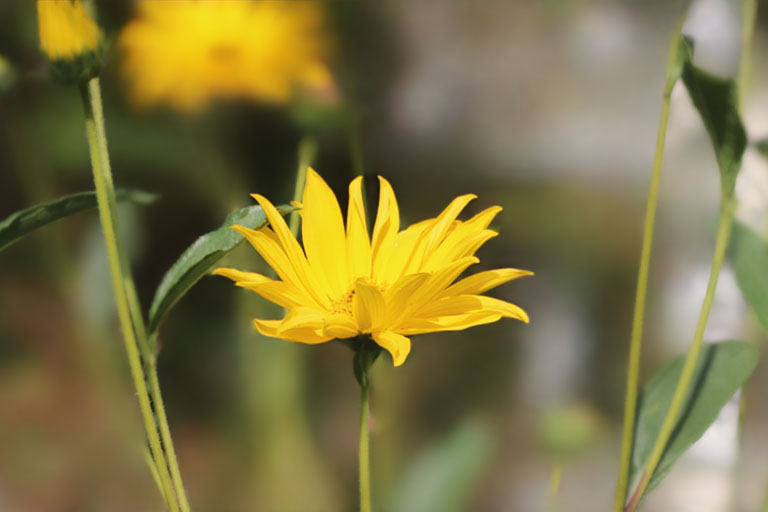}&
    \includegraphics[width=0.24\linewidth]{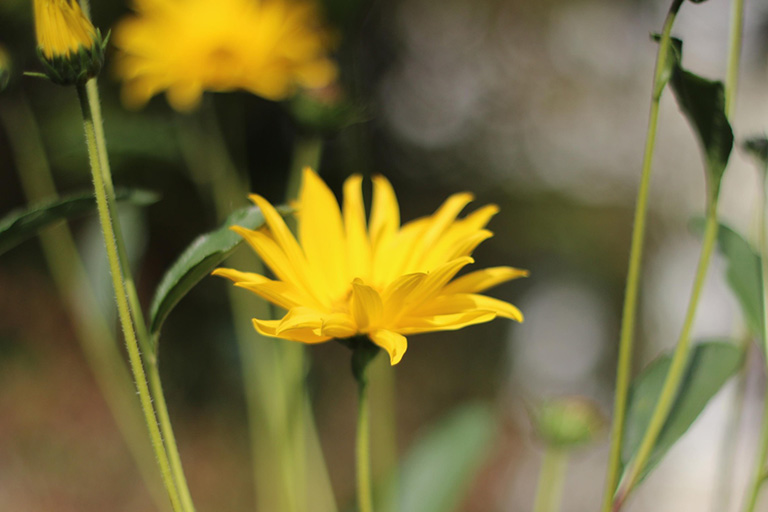}\\
    \includegraphics[width=0.24\linewidth]{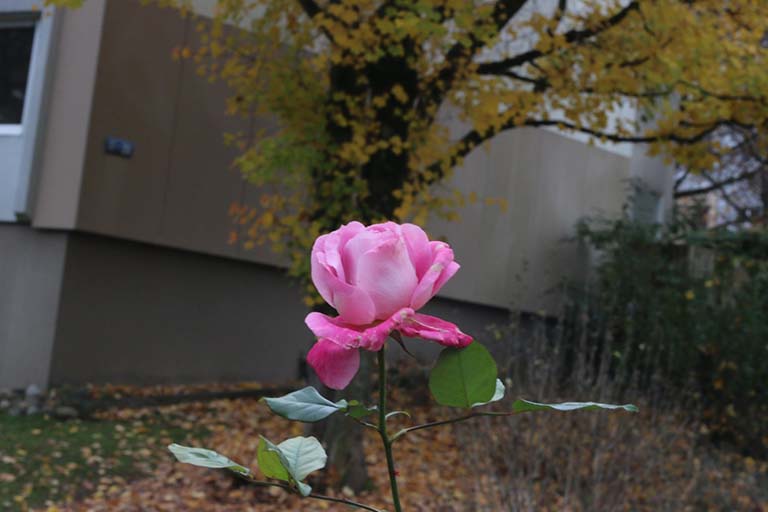}&
    \includegraphics[width=0.24\linewidth]{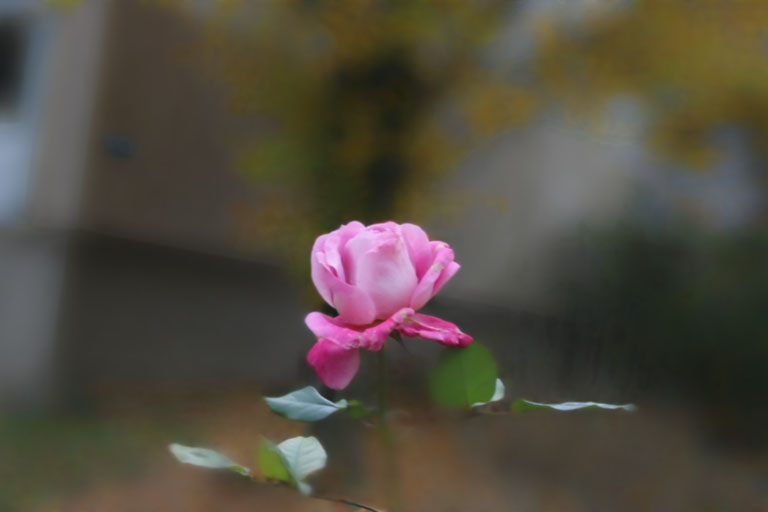}&
    \includegraphics[width=0.24\linewidth]{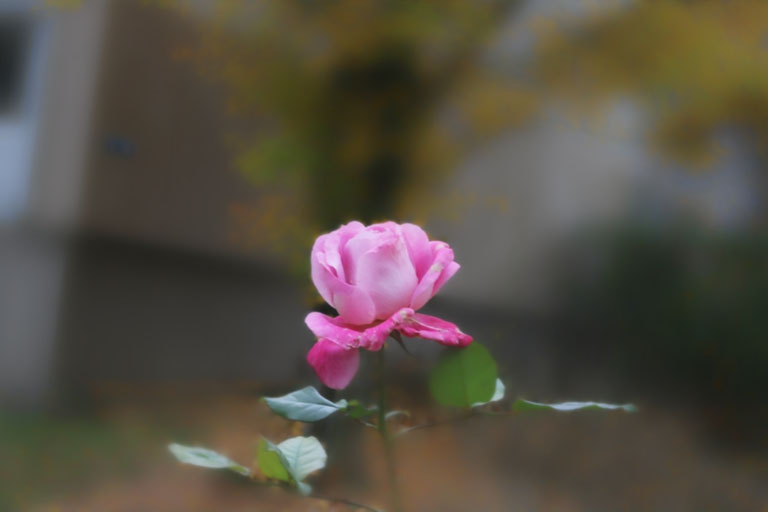}&
    \includegraphics[width=0.24\linewidth]{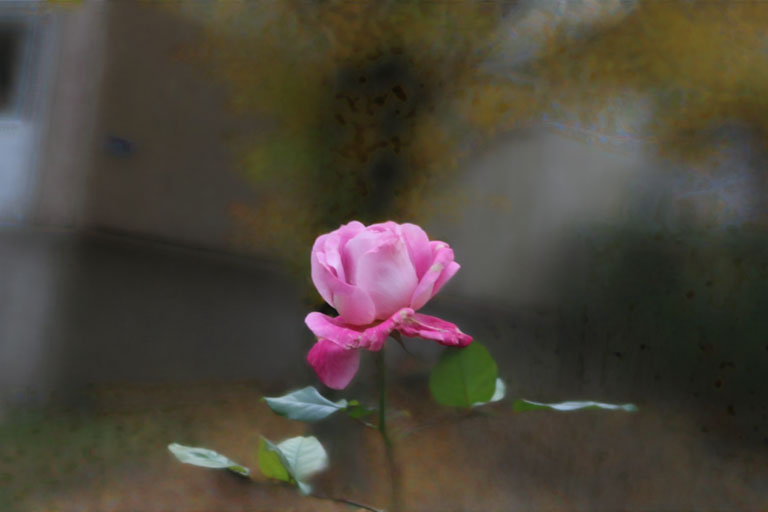}&
    \includegraphics[width=0.24\linewidth]{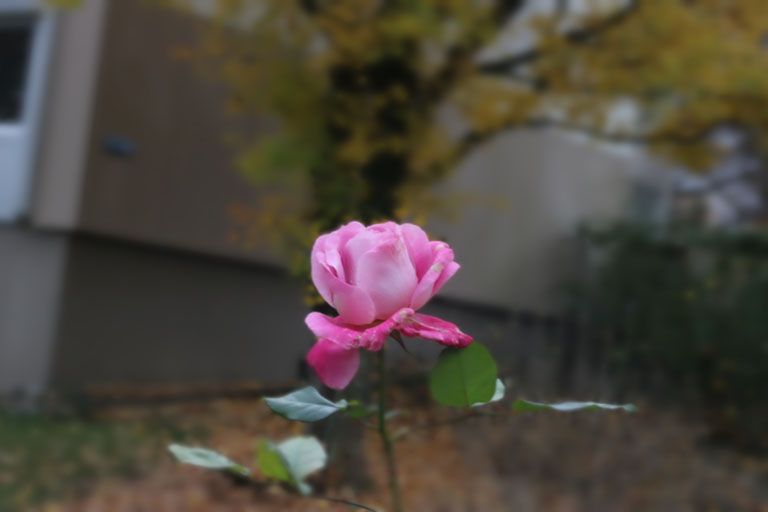}&
    \includegraphics[width=0.24\linewidth]{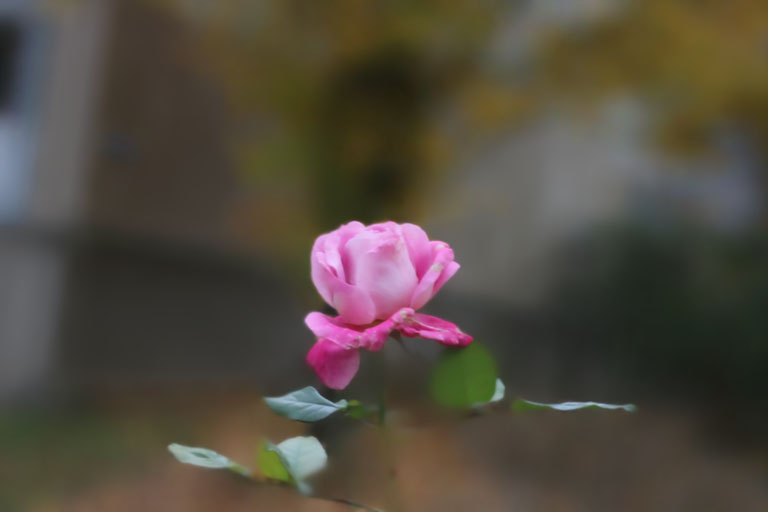}&
    \includegraphics[width=0.24\linewidth]{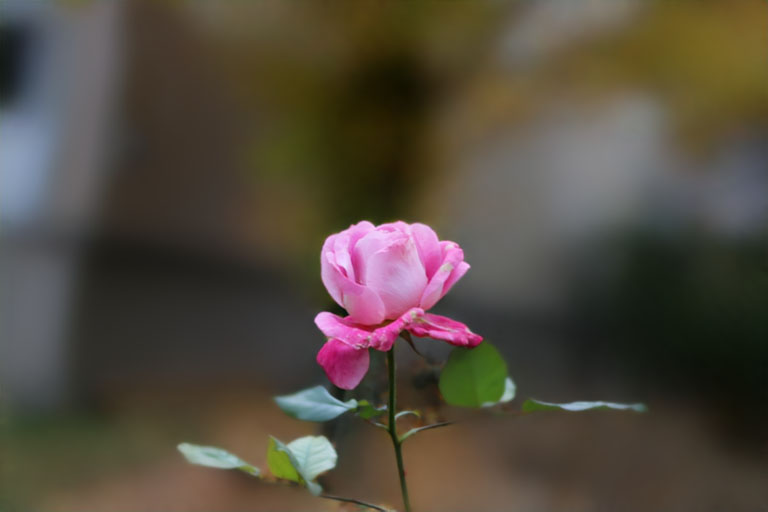}&
    \includegraphics[width=0.24\linewidth]{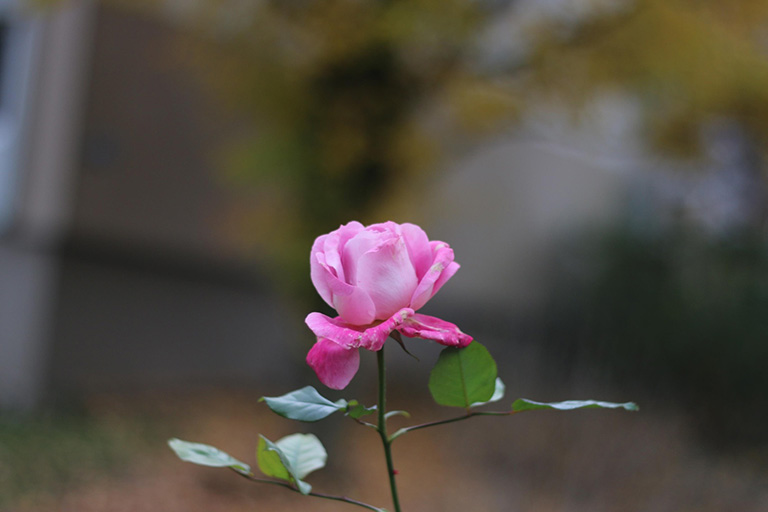}\\
    \includegraphics[width=0.24\linewidth]{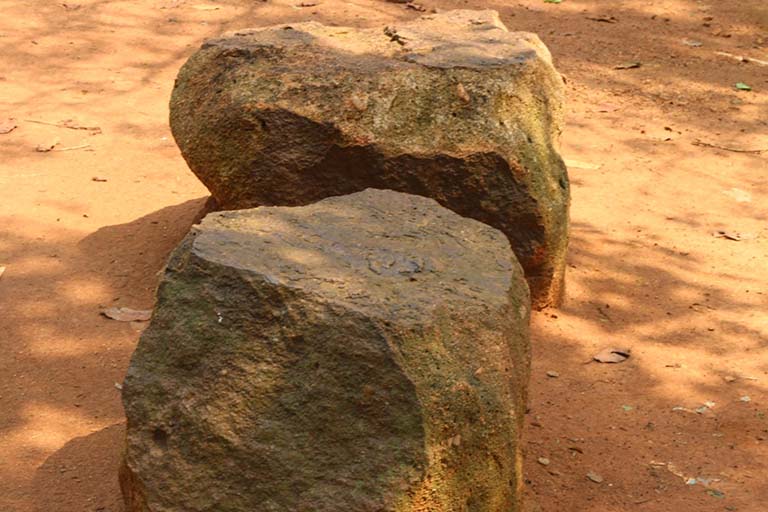}&
    \includegraphics[width=0.24\linewidth]{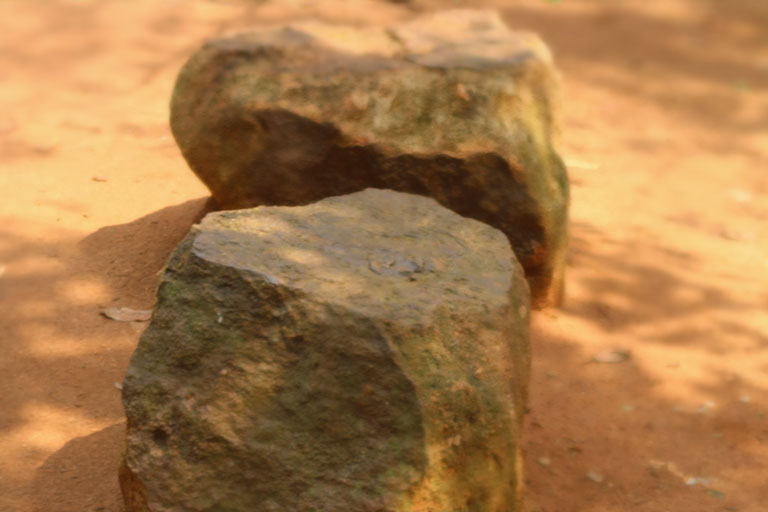}&
    \includegraphics[width=0.24\linewidth]{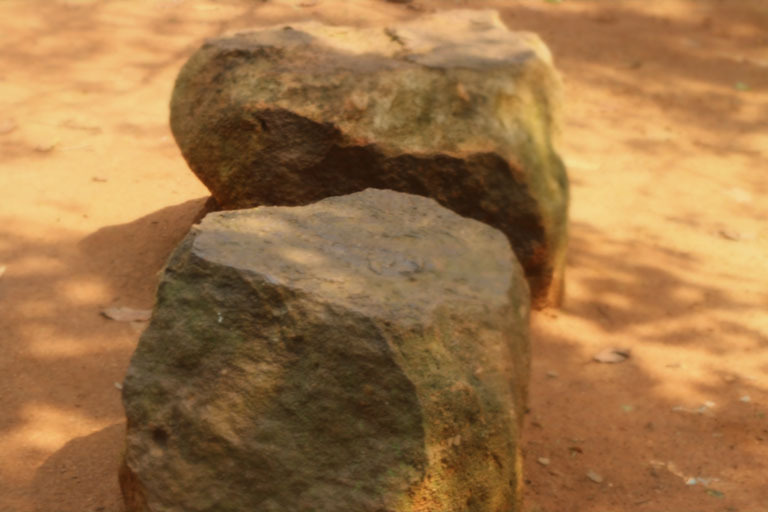}&
    \includegraphics[width=0.24\linewidth]{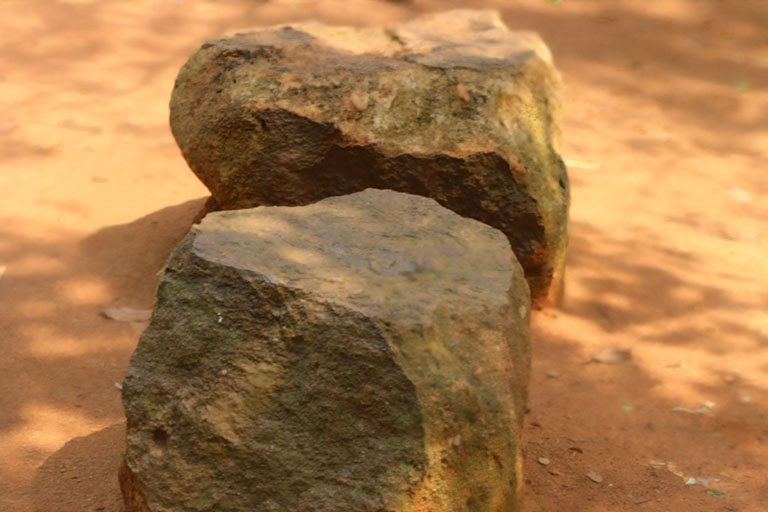}&
    \includegraphics[width=0.24\linewidth]{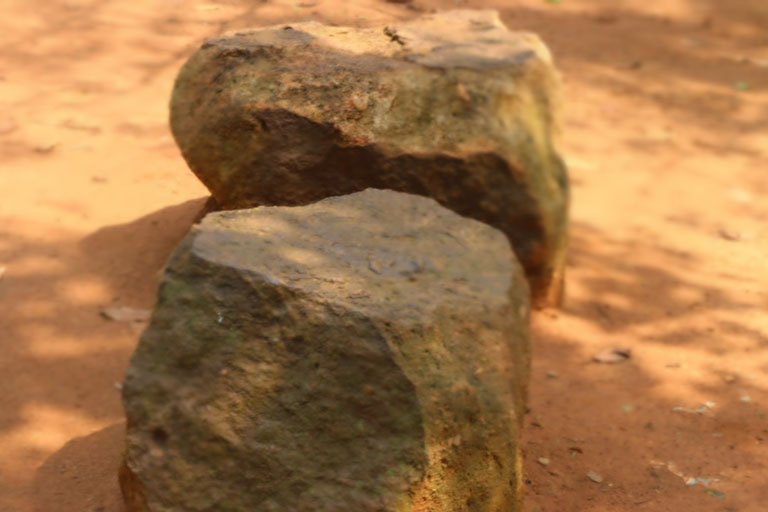}&
    \includegraphics[width=0.24\linewidth]{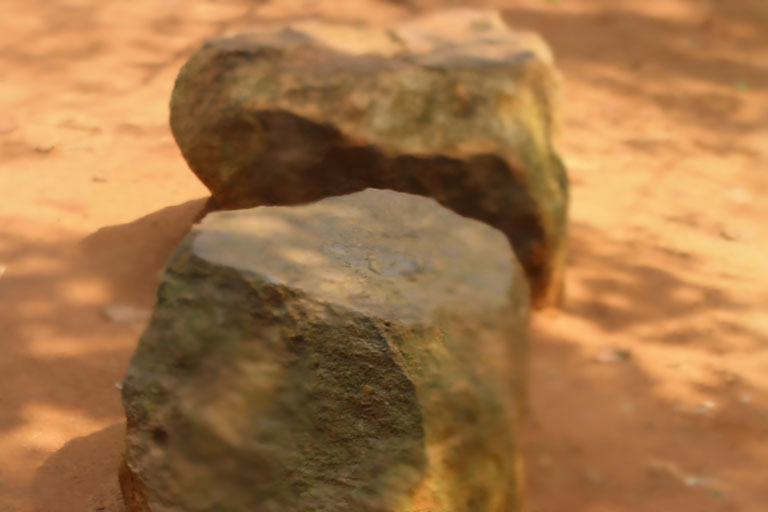}&
    \includegraphics[width=0.24\linewidth]{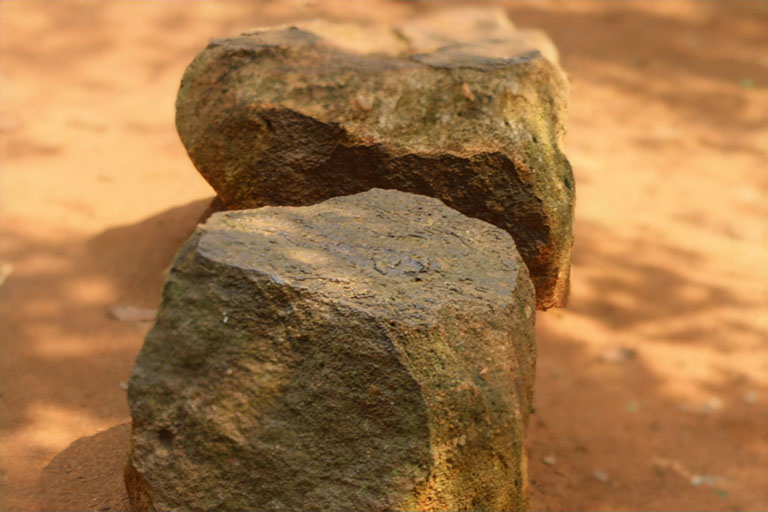}&
    \includegraphics[width=0.24\linewidth]{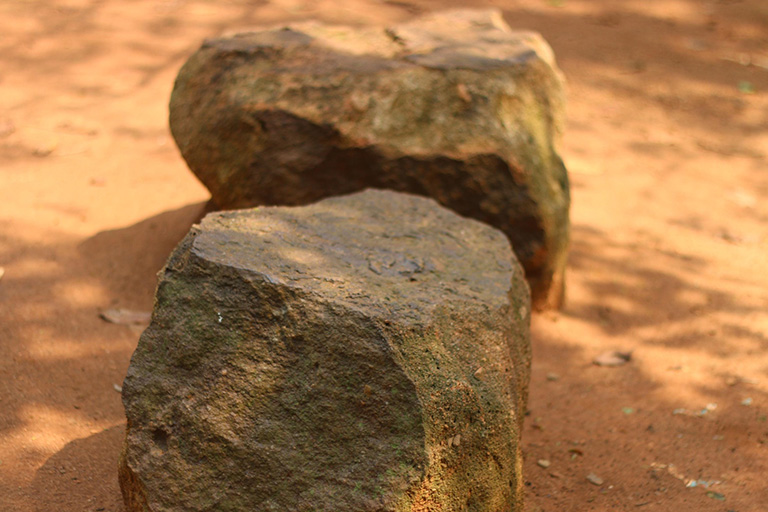}\\
    \includegraphics[width=0.24\linewidth]{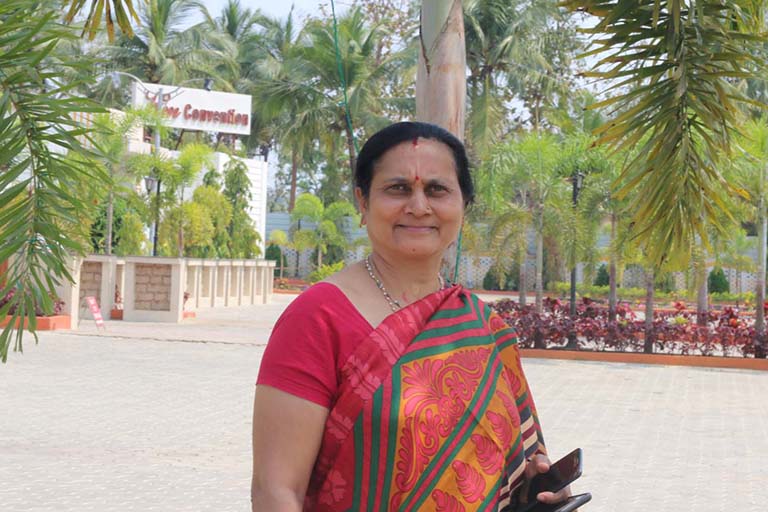}&
    \includegraphics[width=0.24\linewidth]{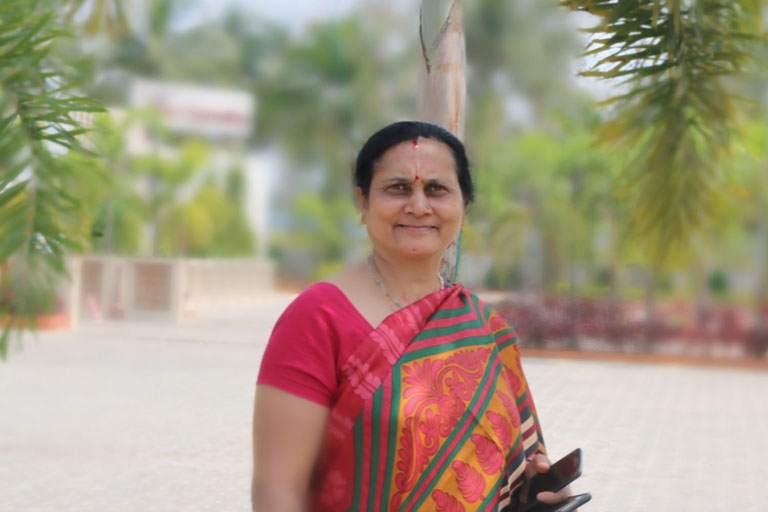}&
    \includegraphics[width=0.24\linewidth]{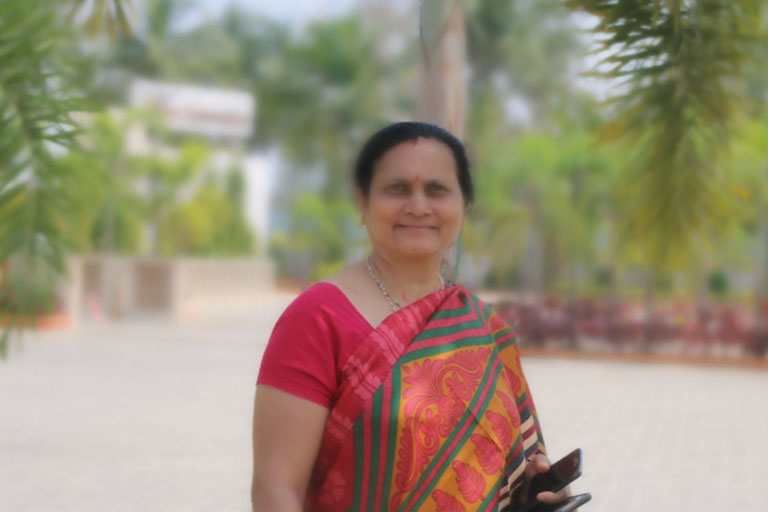}&
    \includegraphics[width=0.24\linewidth]{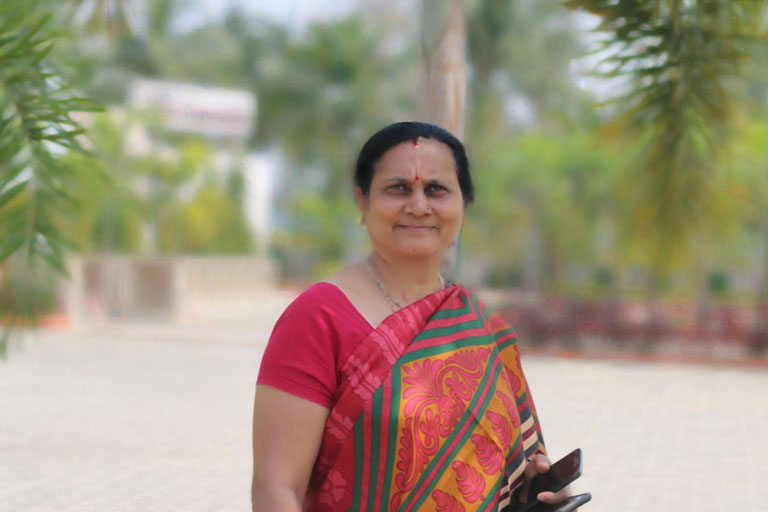}&
    \includegraphics[width=0.24\linewidth]{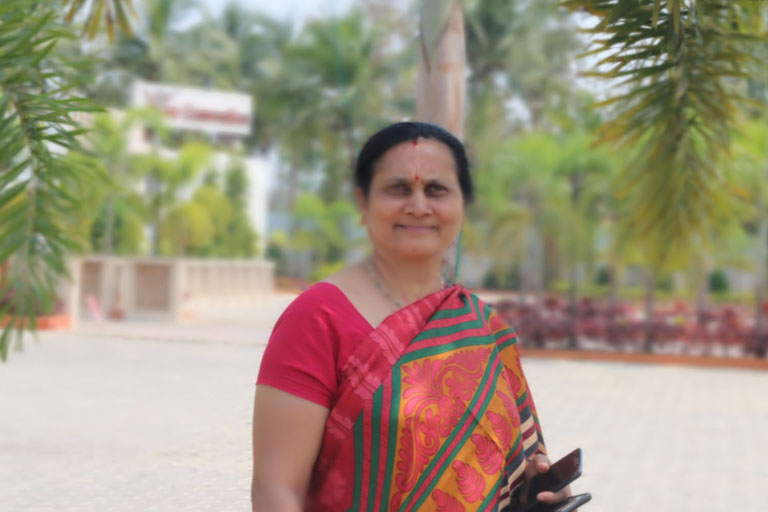}&
    \includegraphics[width=0.24\linewidth]{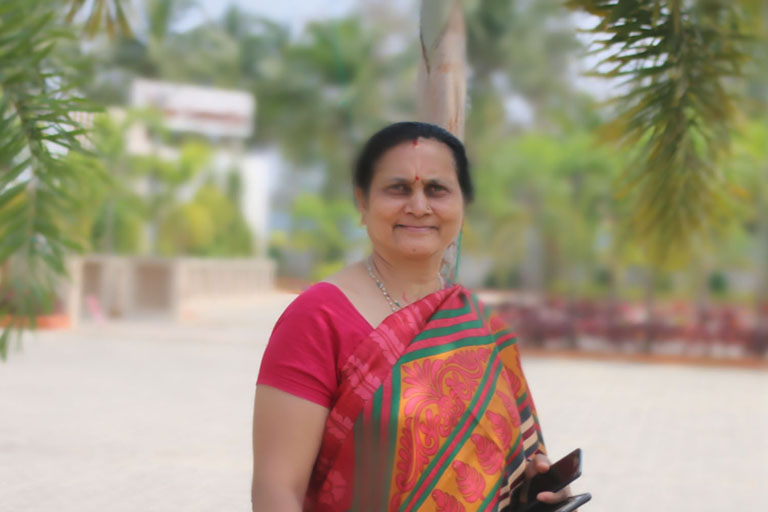}&
    \includegraphics[width=0.24\linewidth]{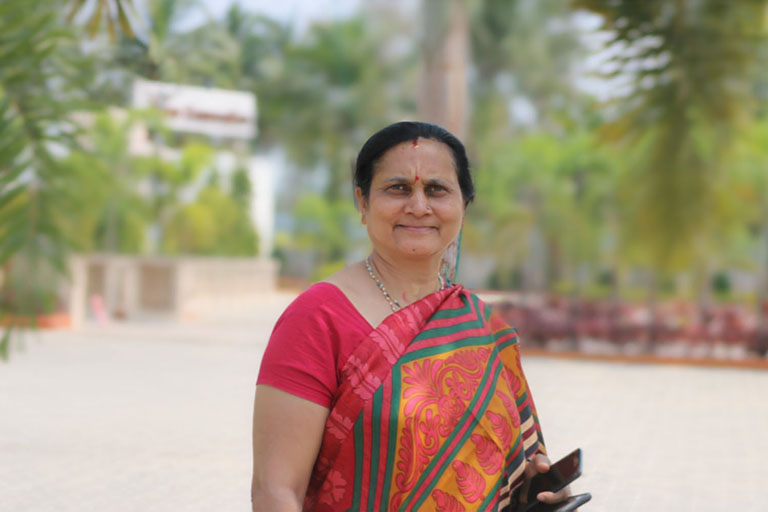}&
    \includegraphics[width=0.24\linewidth]{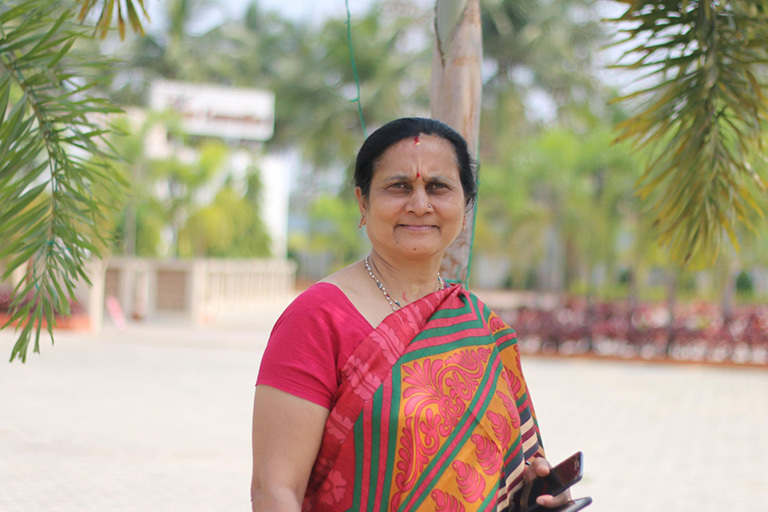}\\
    \includegraphics[width=0.24\linewidth]{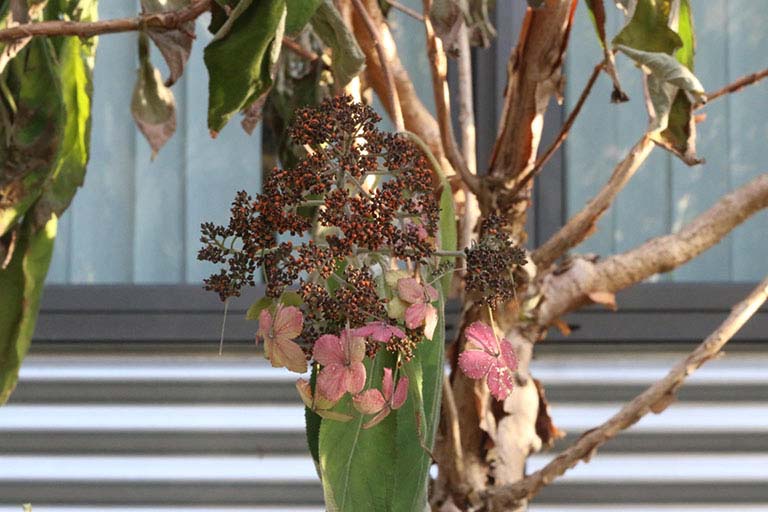}&
    \includegraphics[width=0.24\linewidth]{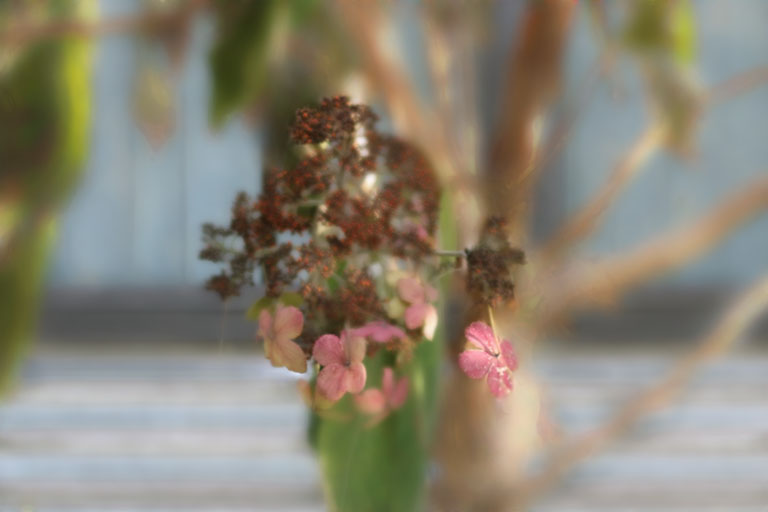}&
    \includegraphics[width=0.24\linewidth]{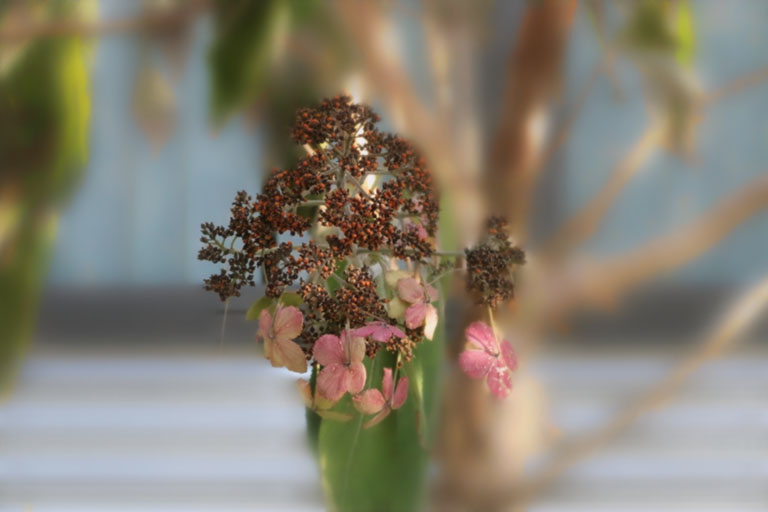}&
    \includegraphics[width=0.24\linewidth]{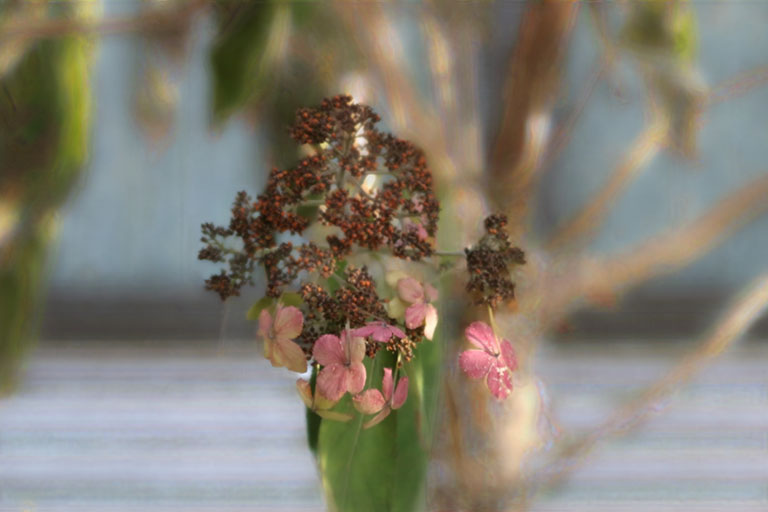}&
    \includegraphics[width=0.24\linewidth]{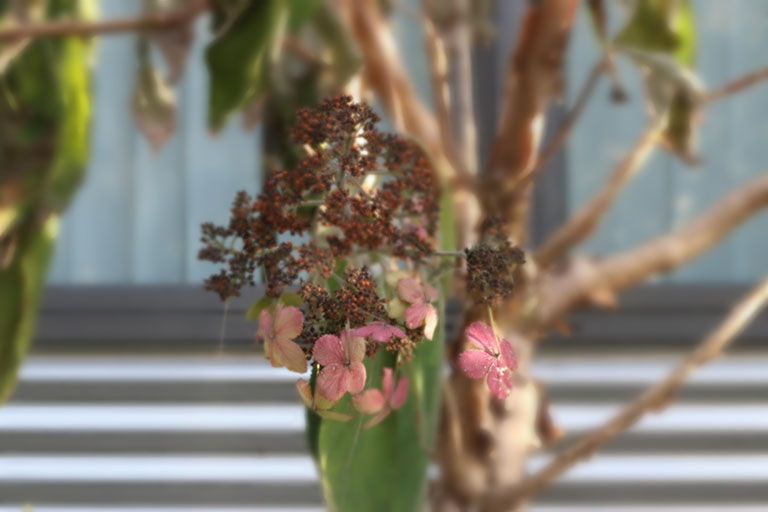}&
    \includegraphics[width=0.24\linewidth]{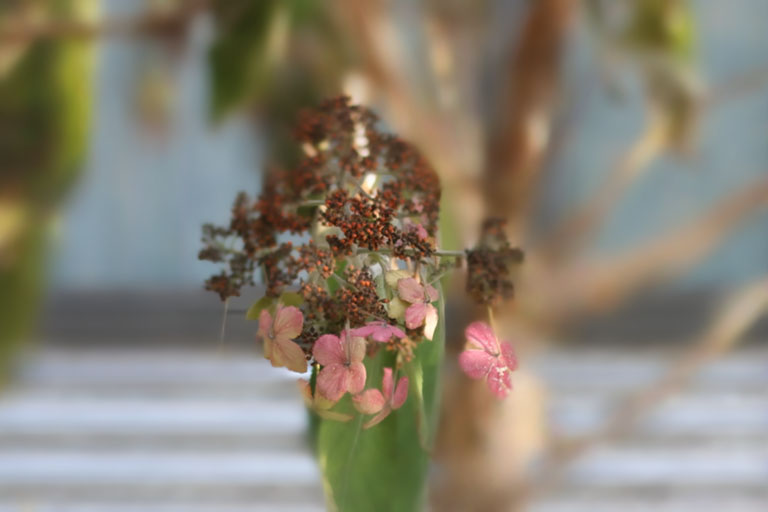}&
    \includegraphics[width=0.24\linewidth]{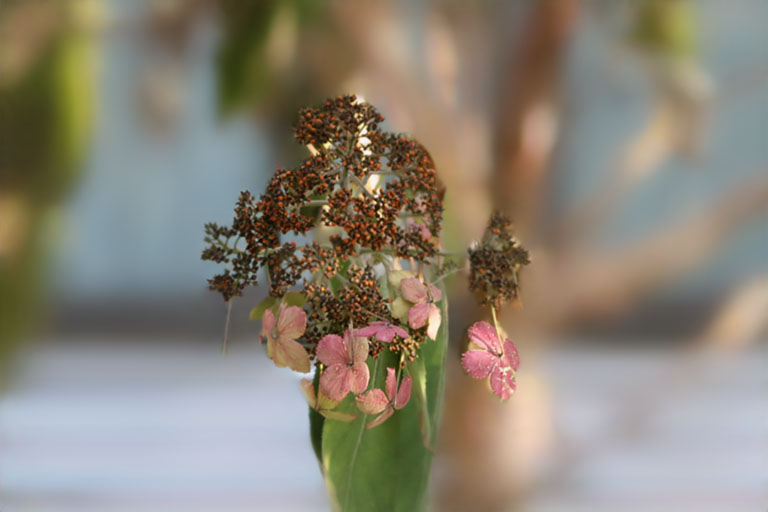}&
    \includegraphics[width=0.24\linewidth]{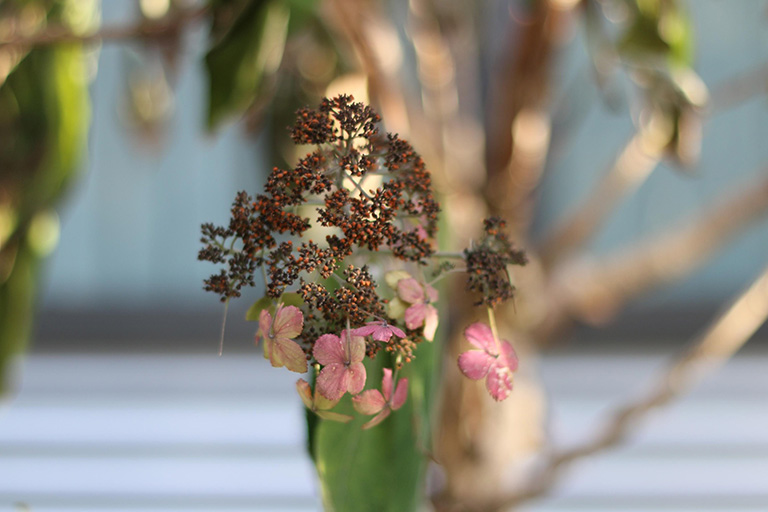}\\
Original Image & Yang~\etal~\cite{ignatov2019aim} & Xiong ~\etal~\cite{ignatov2019aim} & Purohit~\etal~\cite{purohit2019depth} & Dutta~\etal~\cite{ignatov2019aim} & Zheng~\etal~\cite{ignatov2019aim} & PyNET (Ours) & Canon Photo \\
\end{tabular}
}
\vspace{0.5mm}
\caption{\small{Visual results obtained with 6 different methods. From left to right, top to bottom: the original wide depth-of-field image, Yang~\etal\cite{ignatov2019aim}, Xiong ~\etal\cite{ignatov2019aim}, Purohit~\etal\cite{purohit2019depth}, Dutta~\etal\cite{ignatov2019aim}, Zheng~\etal\cite{ignatov2019aim}, our PyNET-based solution and the target Canon photo.}}
\label{fig:method_comparison}
\vspace{-0.7mm}
\end{figure*}

\begin{figure*}[t!]
\centering
\setlength{\tabcolsep}{1pt}
\resizebox{1.0\linewidth}{!}
{
\begin{tabular}{ccccc}
    \includegraphics[width=0.24\linewidth]{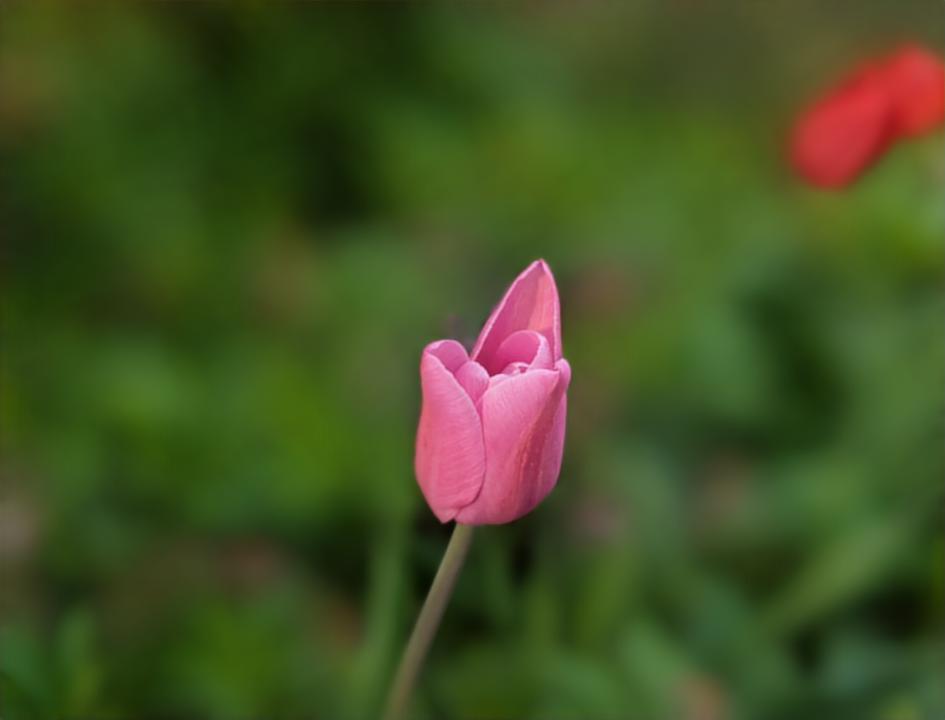}&
    \includegraphics[width=0.24\linewidth]{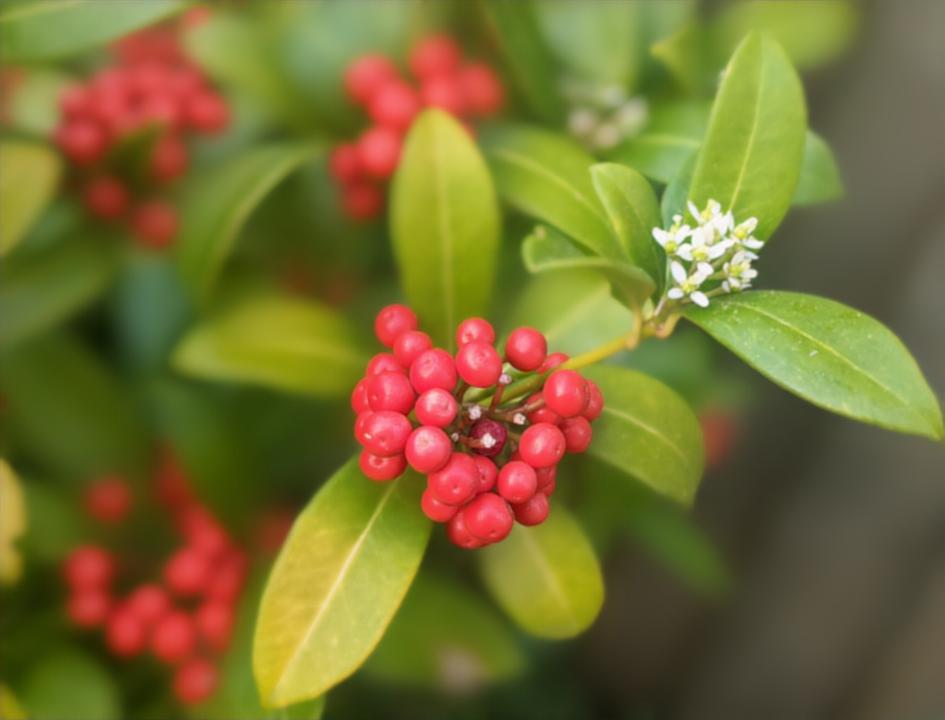}&
    \includegraphics[width=0.24\linewidth]{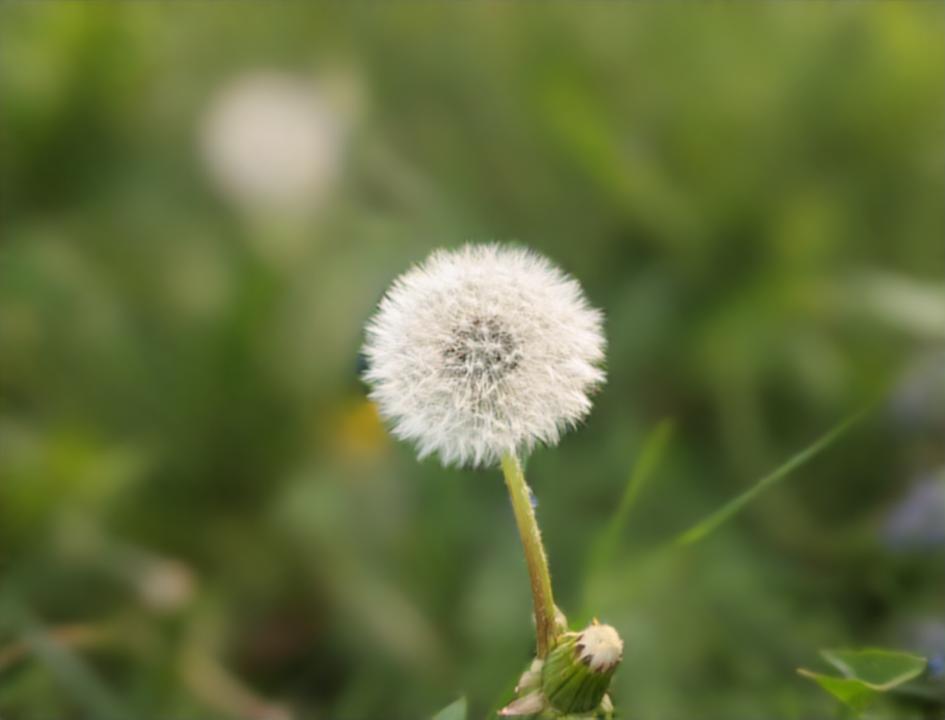}&
    \includegraphics[width=0.24\linewidth]{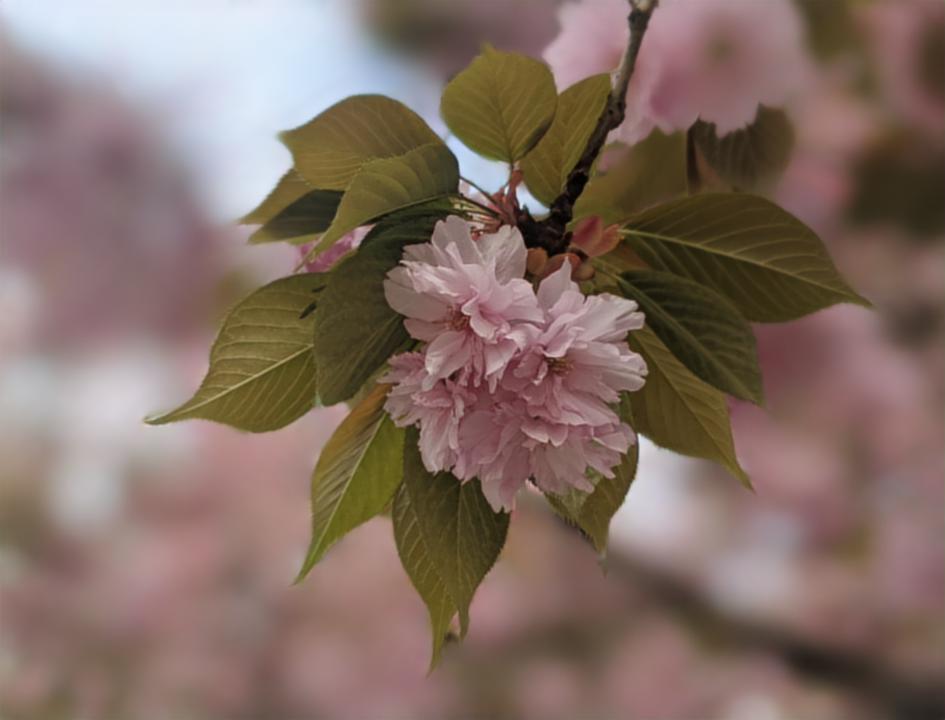}&
    \includegraphics[width=0.24\linewidth]{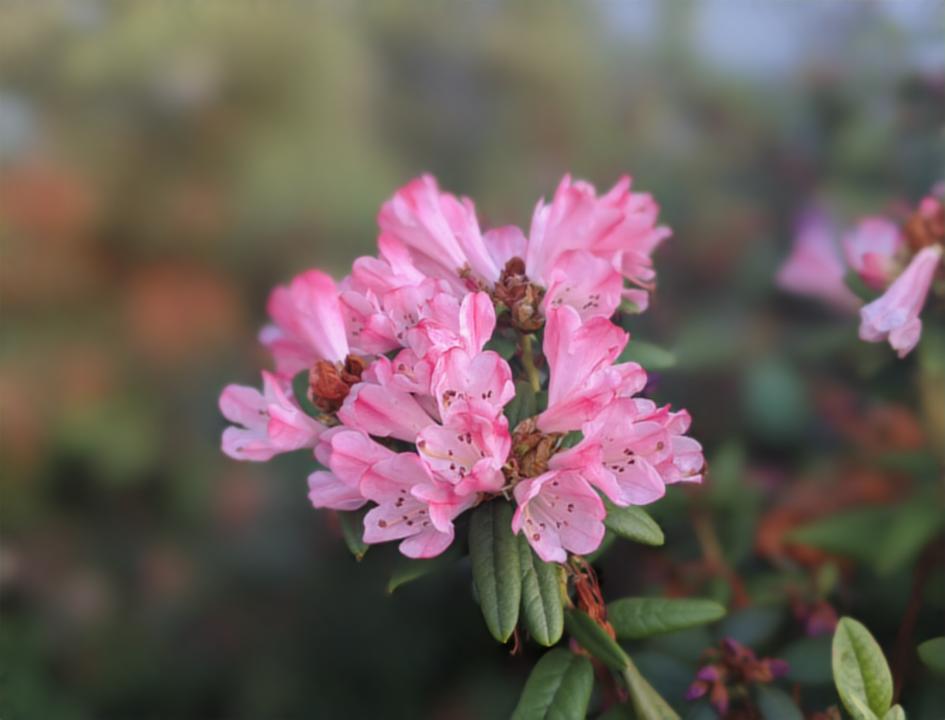}\\
    \includegraphics[width=0.24\linewidth]{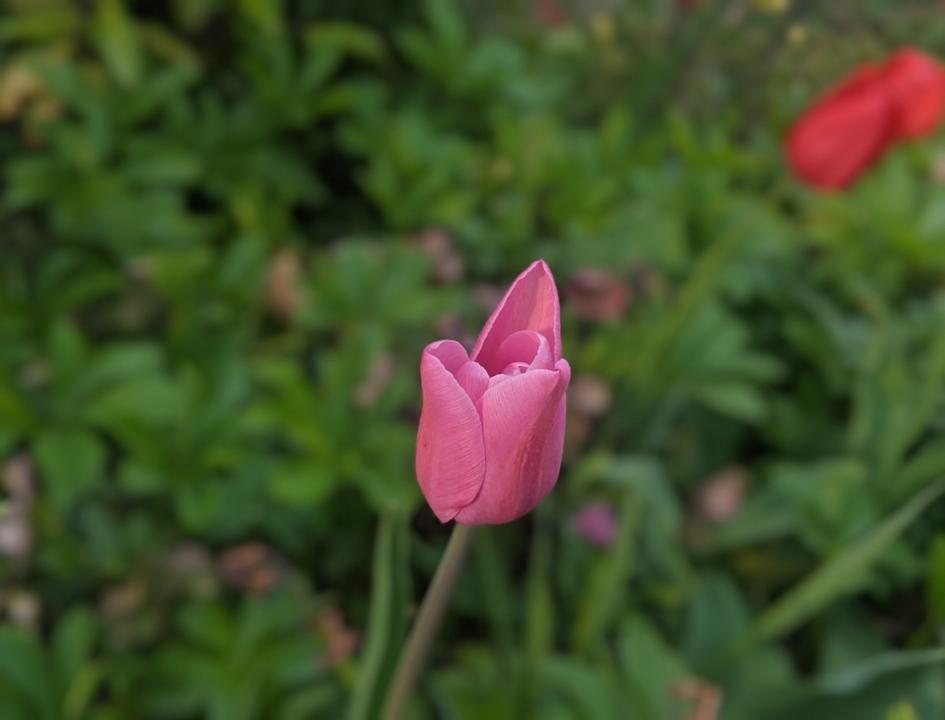}&
    \includegraphics[width=0.24\linewidth]{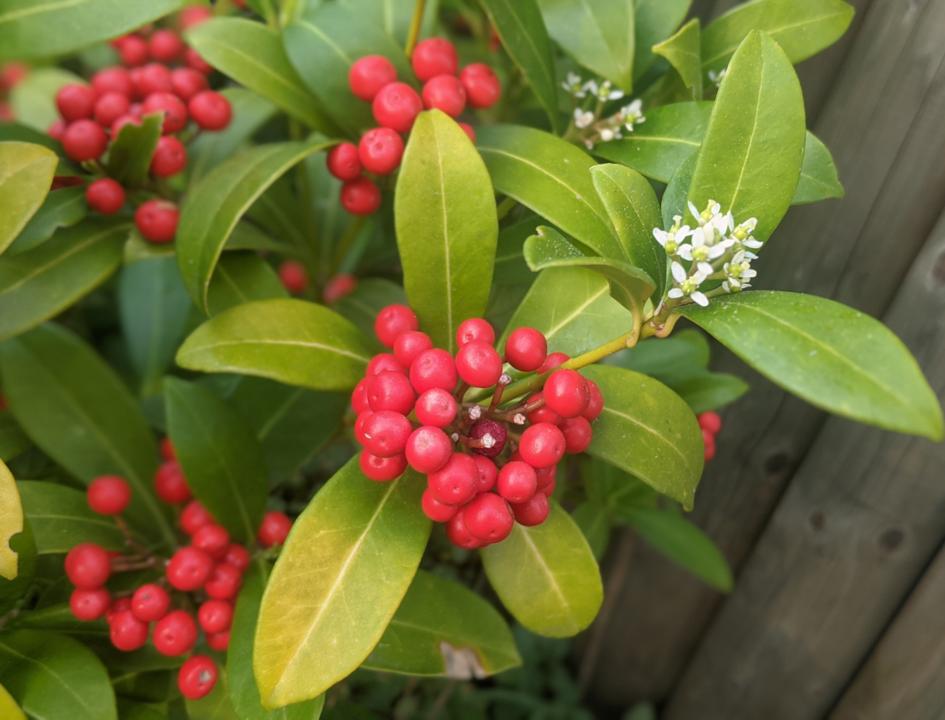}&
    \includegraphics[width=0.24\linewidth]{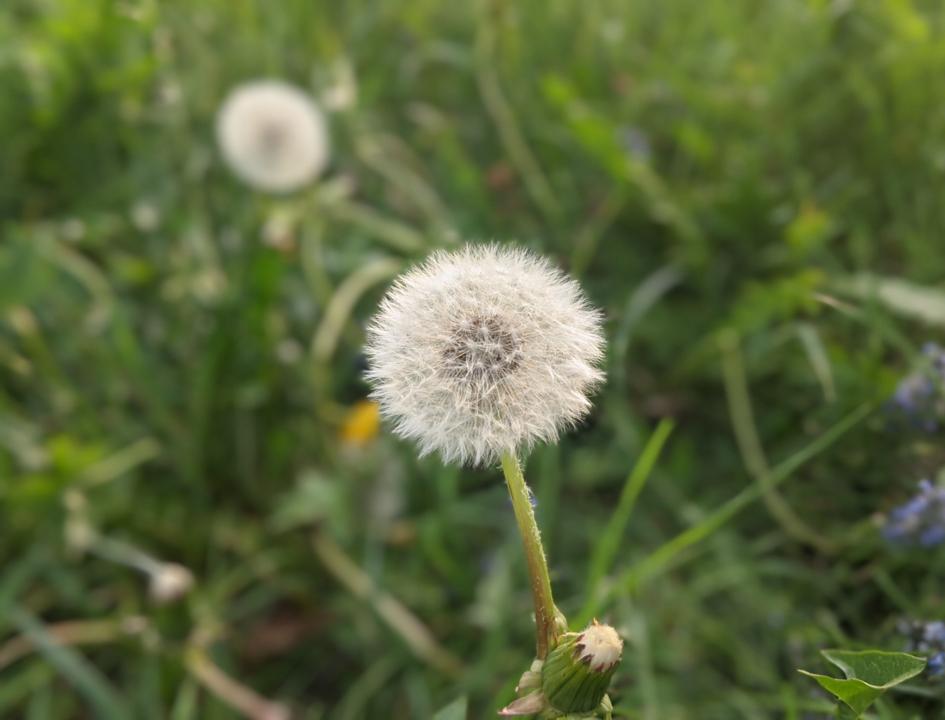}&
    \includegraphics[width=0.24\linewidth]{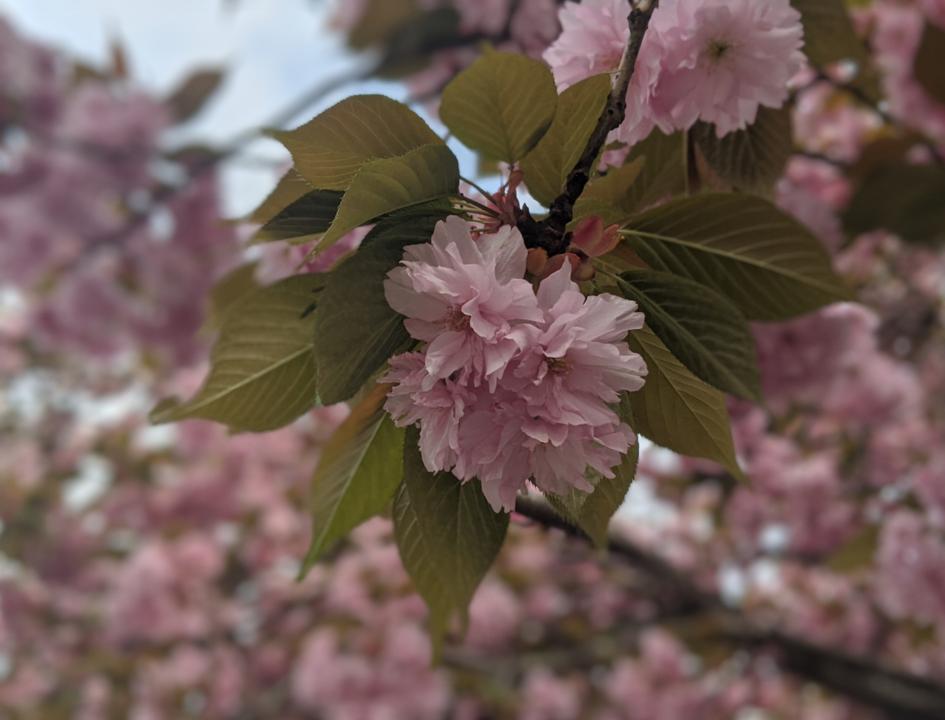}& \vspace{2mm}
    \includegraphics[width=0.24\linewidth]{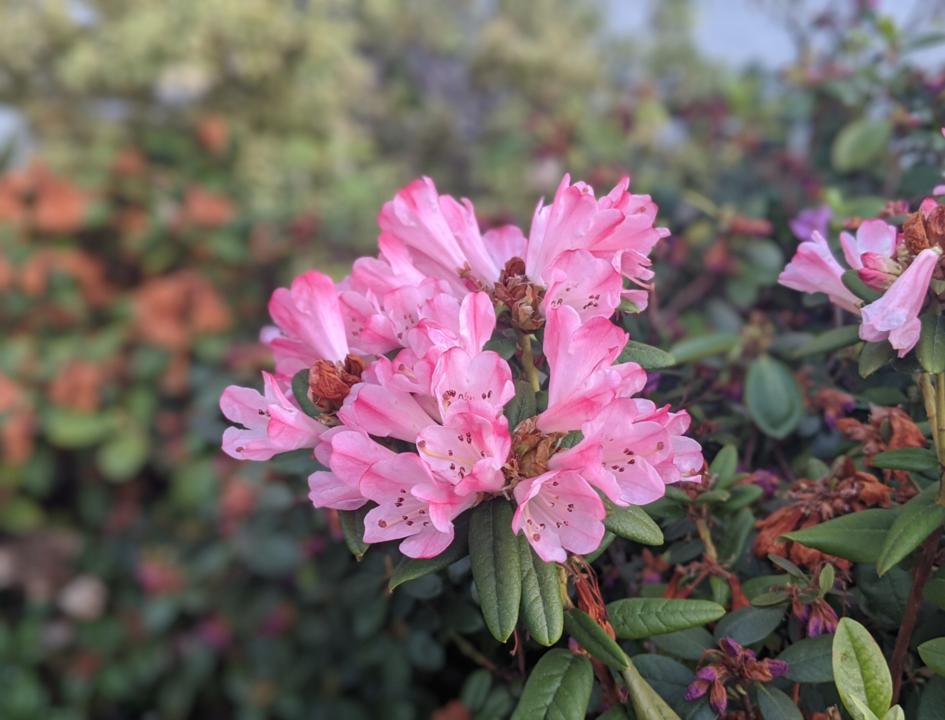}\\
    \includegraphics[width=0.24\linewidth]{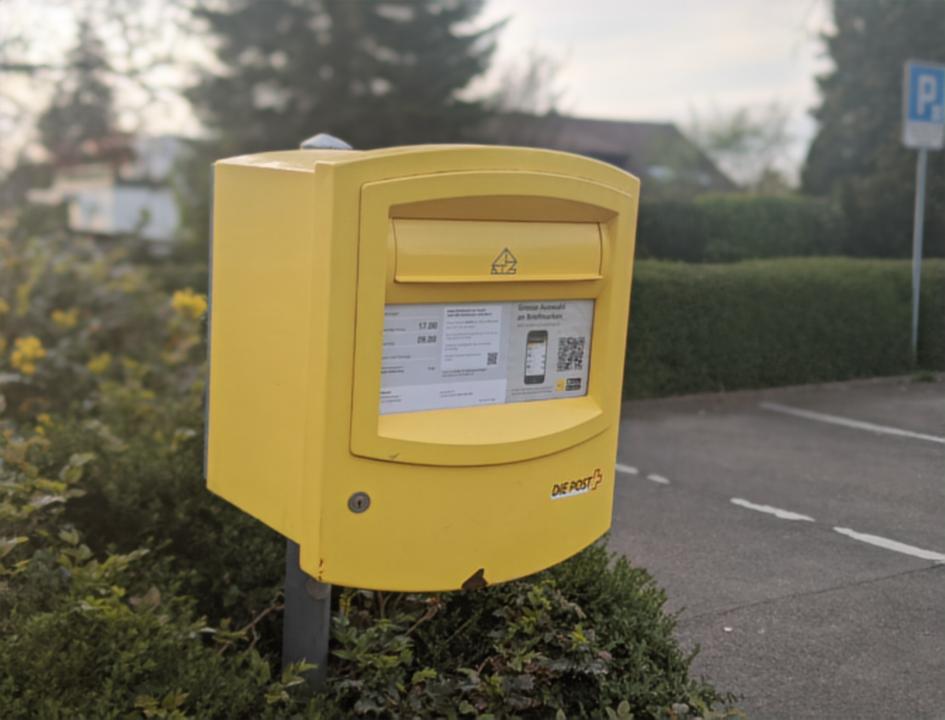}&
    \includegraphics[width=0.24\linewidth]{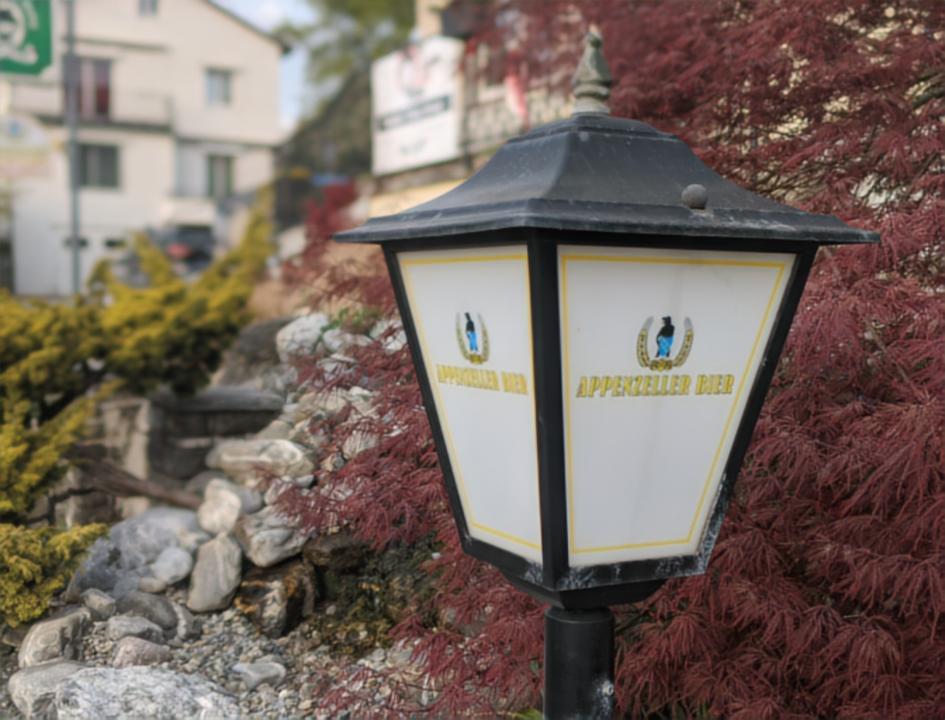}&
    \includegraphics[width=0.24\linewidth]{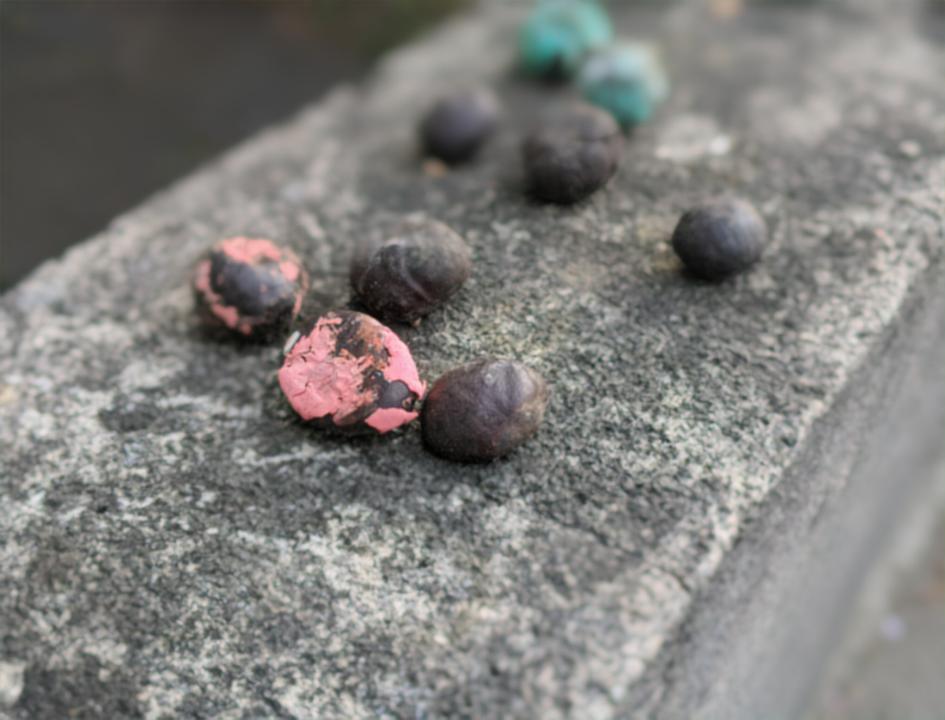}&
    \includegraphics[width=0.24\linewidth]{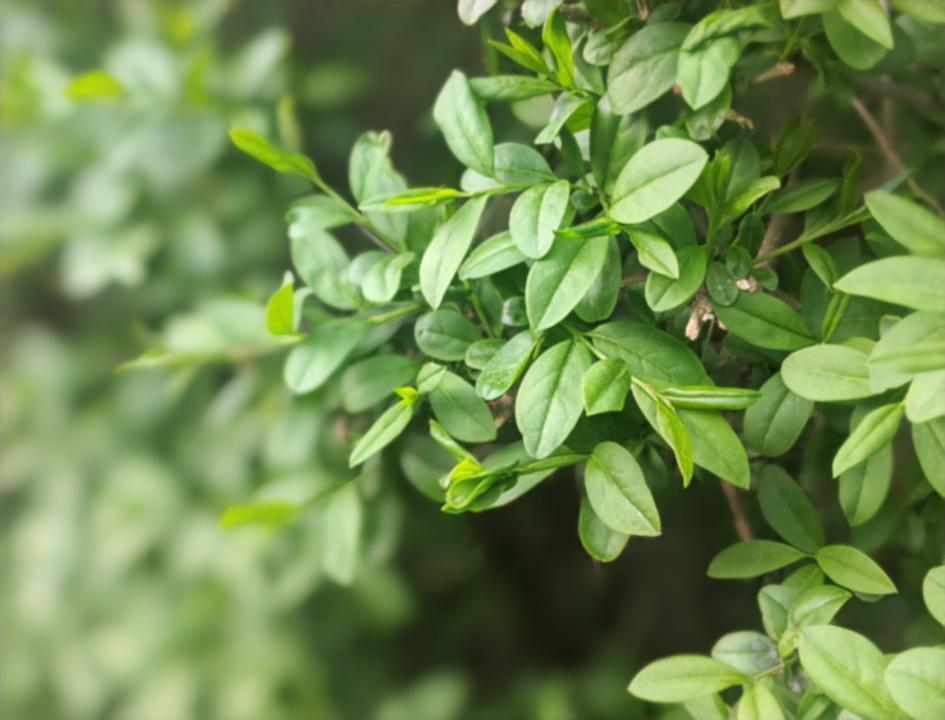}&
    \includegraphics[width=0.24\linewidth]{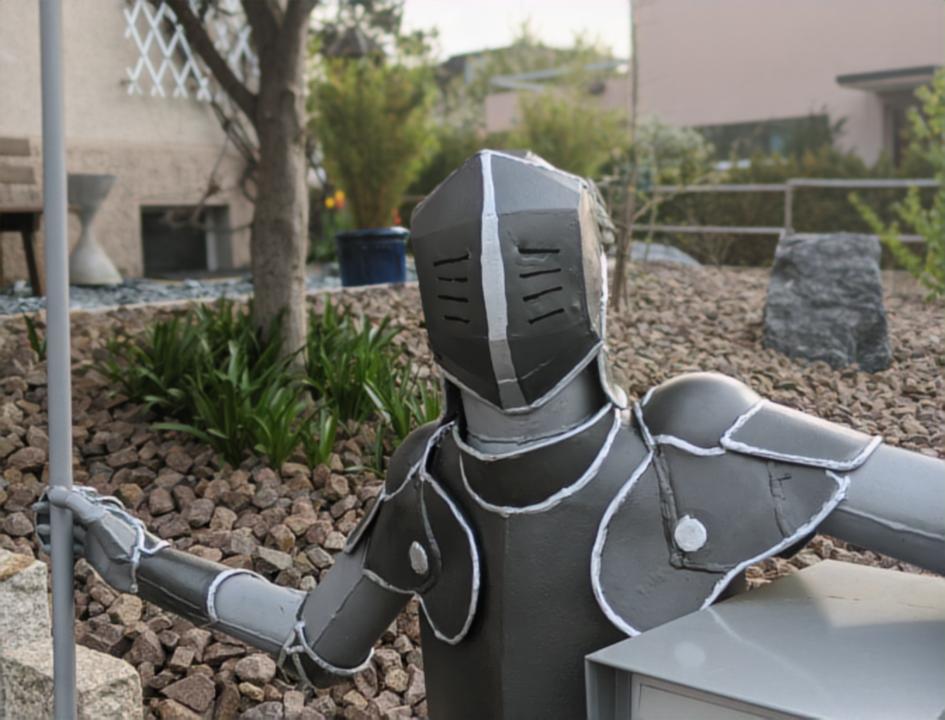}\\
    \includegraphics[width=0.24\linewidth]{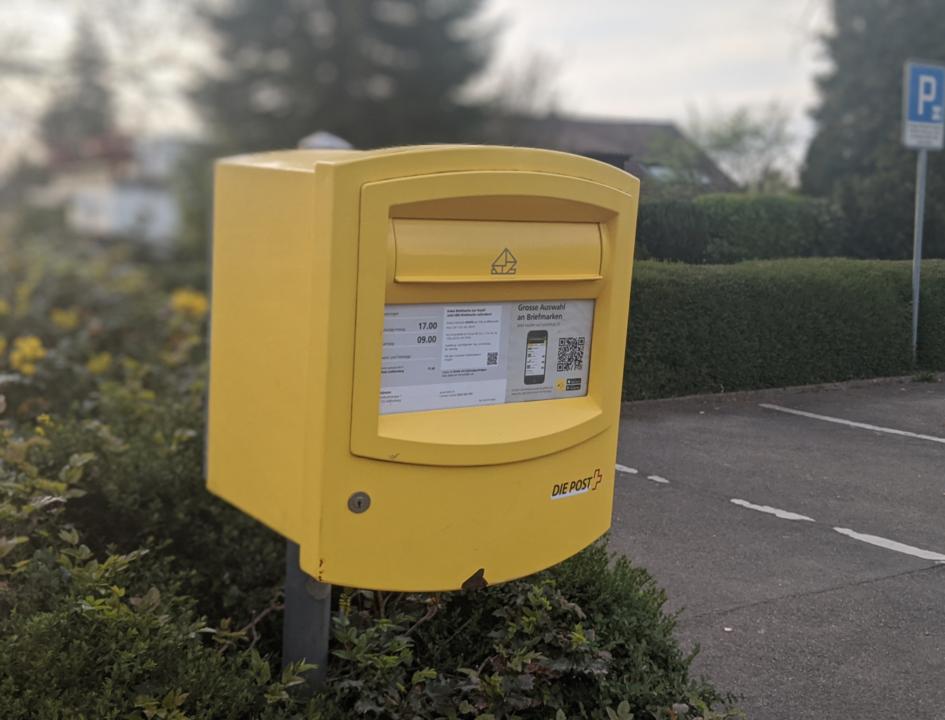}&
    \includegraphics[width=0.24\linewidth]{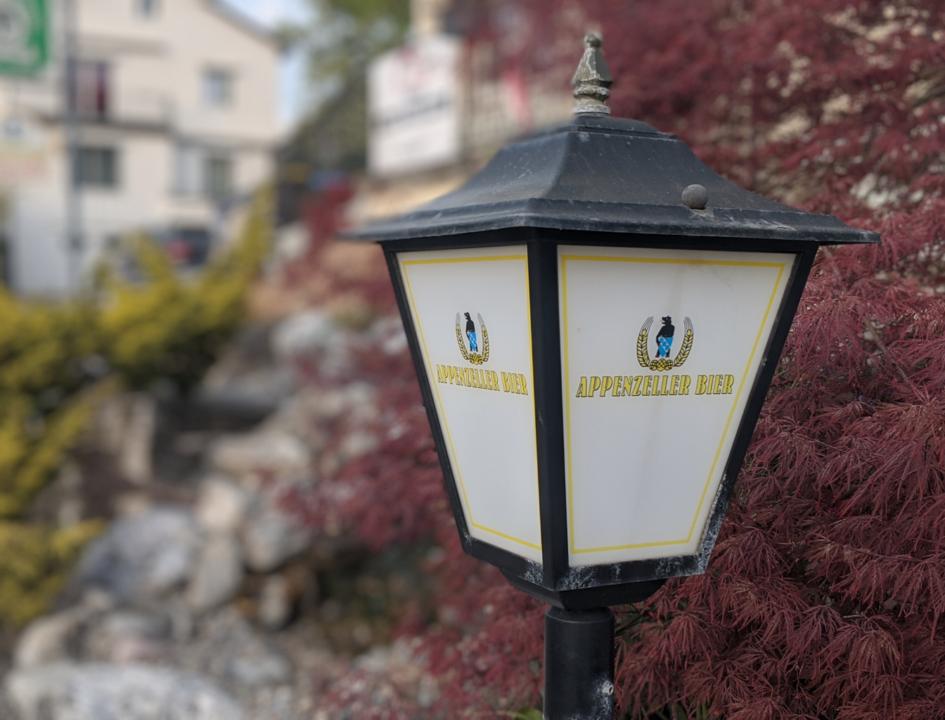}&
    \includegraphics[width=0.24\linewidth]{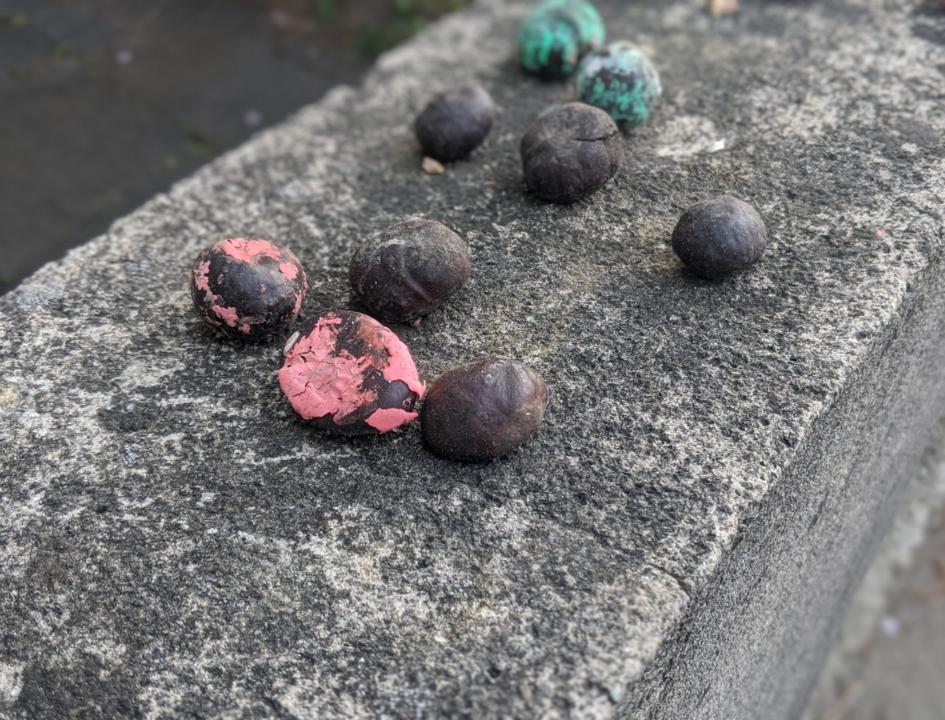}&
    \includegraphics[width=0.24\linewidth]{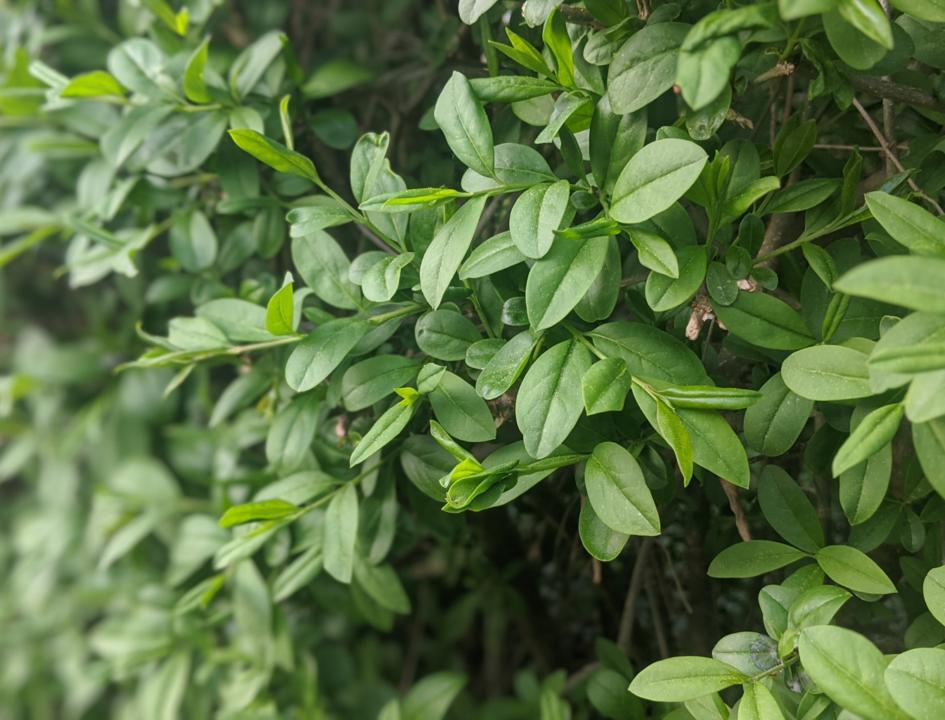}& \vspace{2mm}
    \includegraphics[width=0.24\linewidth]{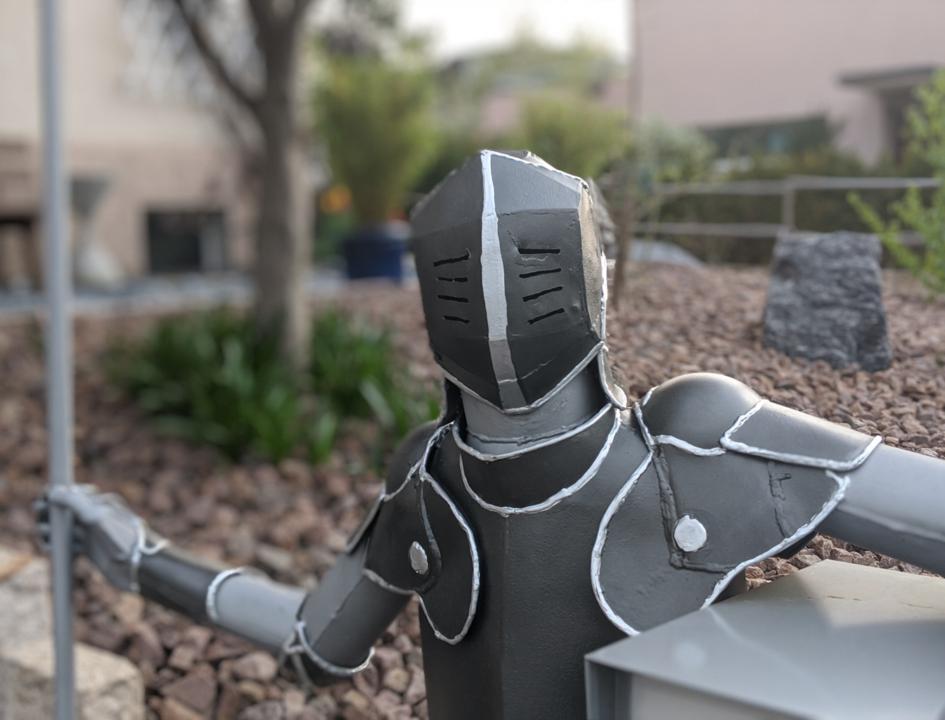}\\
    \includegraphics[width=0.24\linewidth]{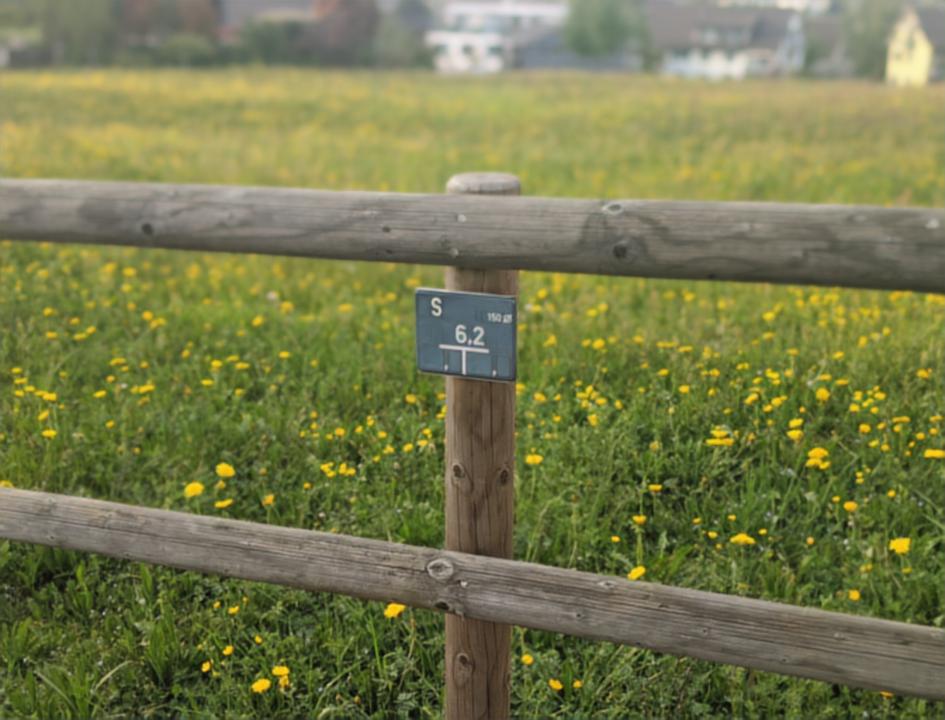}&
    \includegraphics[width=0.24\linewidth]{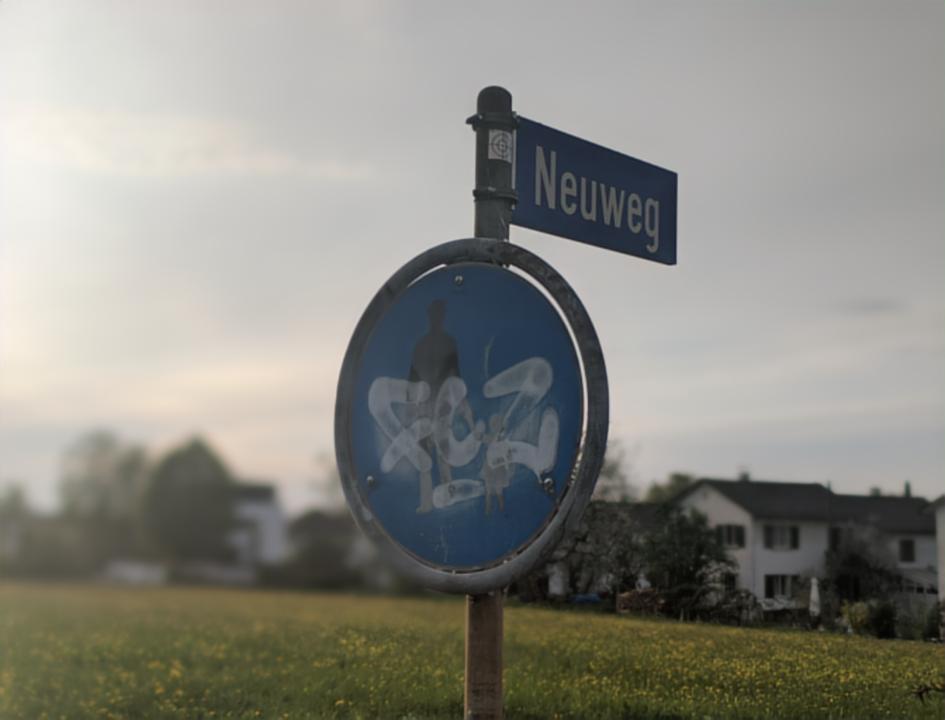}&
    \includegraphics[width=0.24\linewidth]{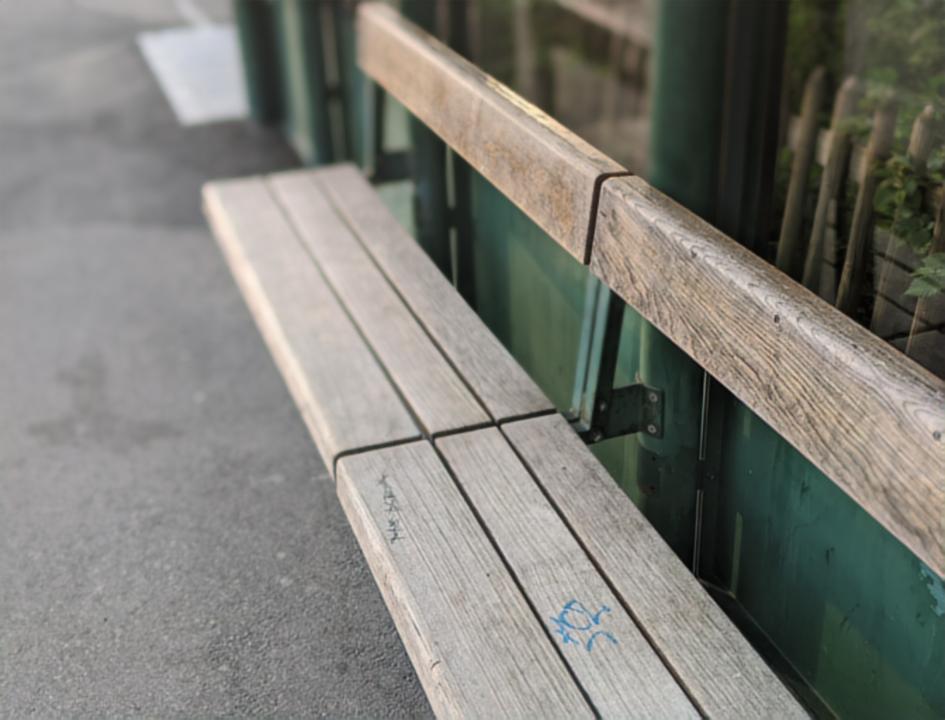}&
    \includegraphics[width=0.24\linewidth]{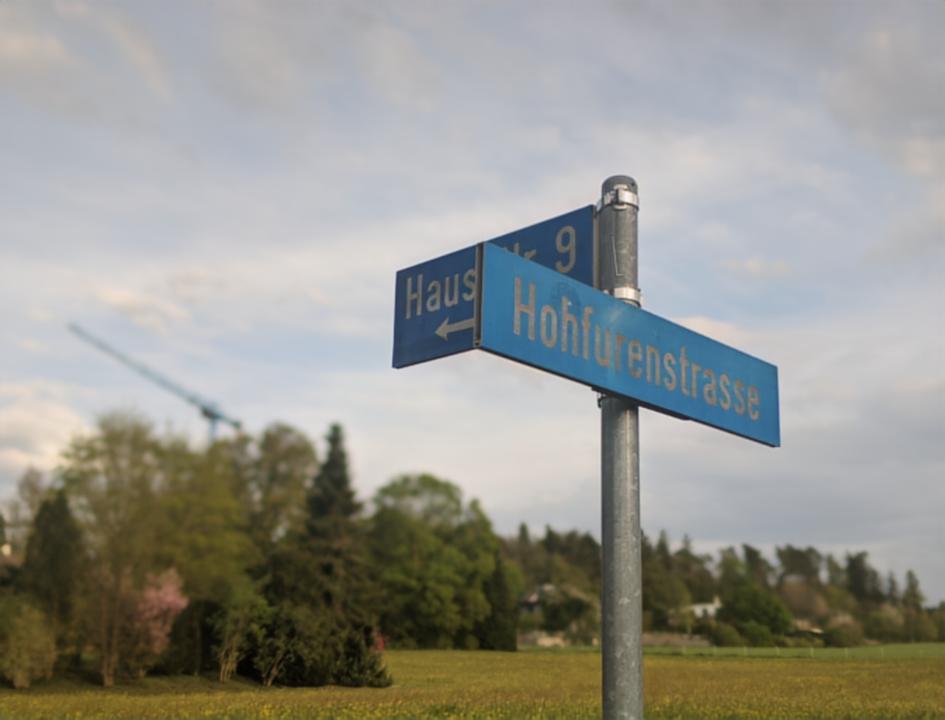}&
    \includegraphics[width=0.24\linewidth]{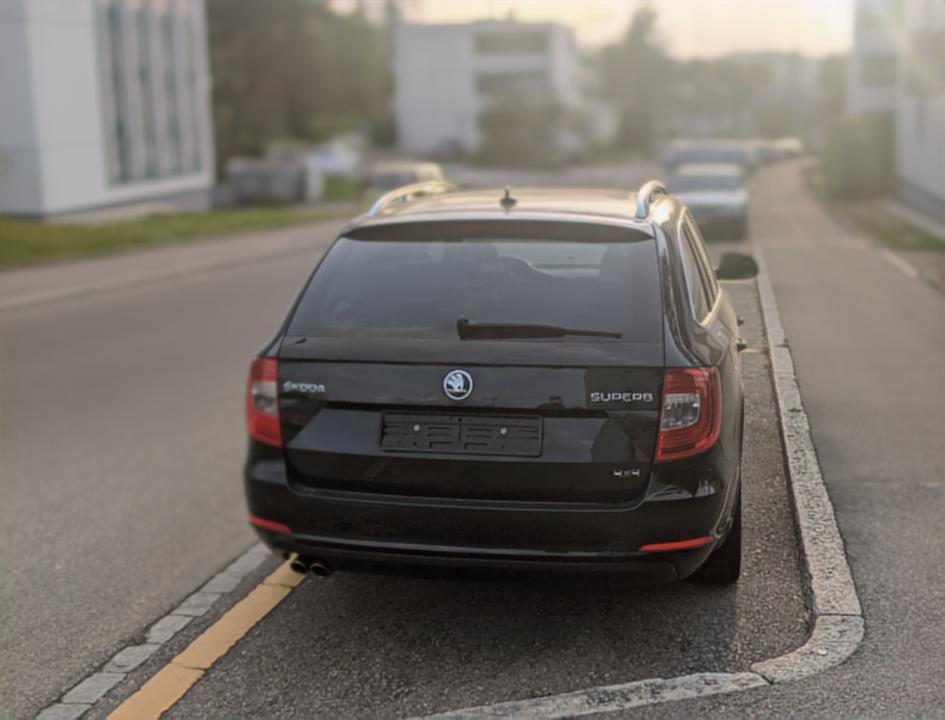}\\
    \includegraphics[width=0.24\linewidth]{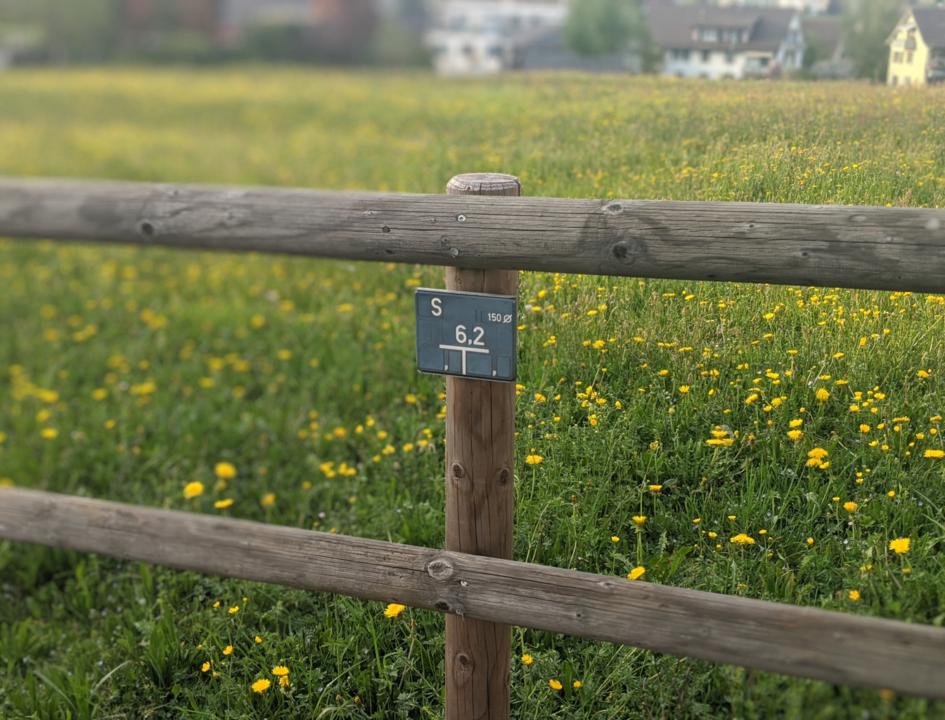}&
    \includegraphics[width=0.24\linewidth]{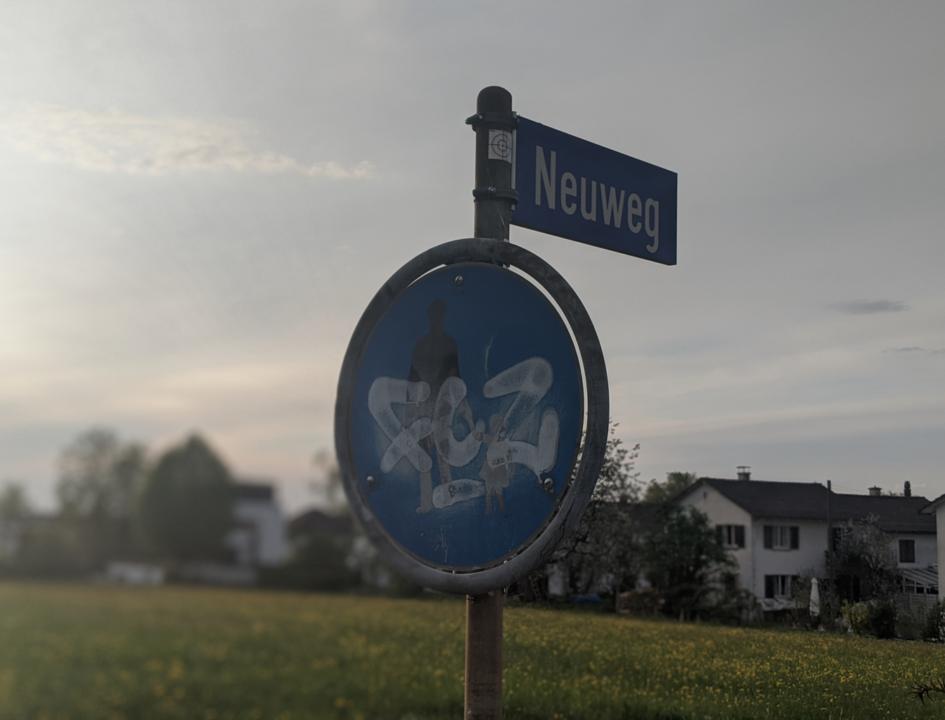}&
    \includegraphics[width=0.24\linewidth]{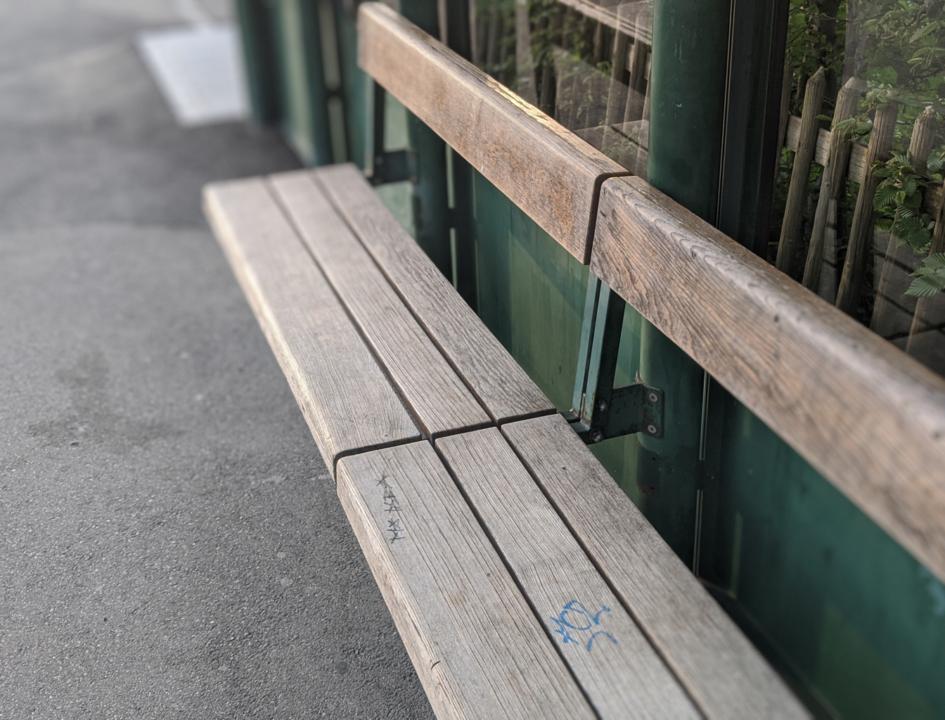}&
    \includegraphics[width=0.24\linewidth]{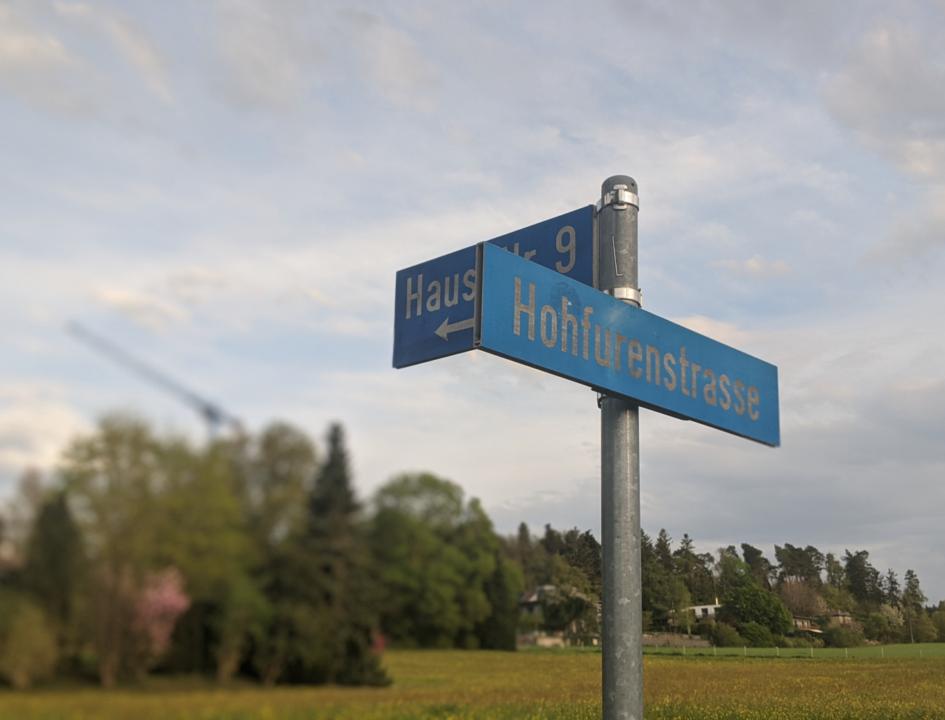}& \vspace{2mm}
    \includegraphics[width=0.24\linewidth]{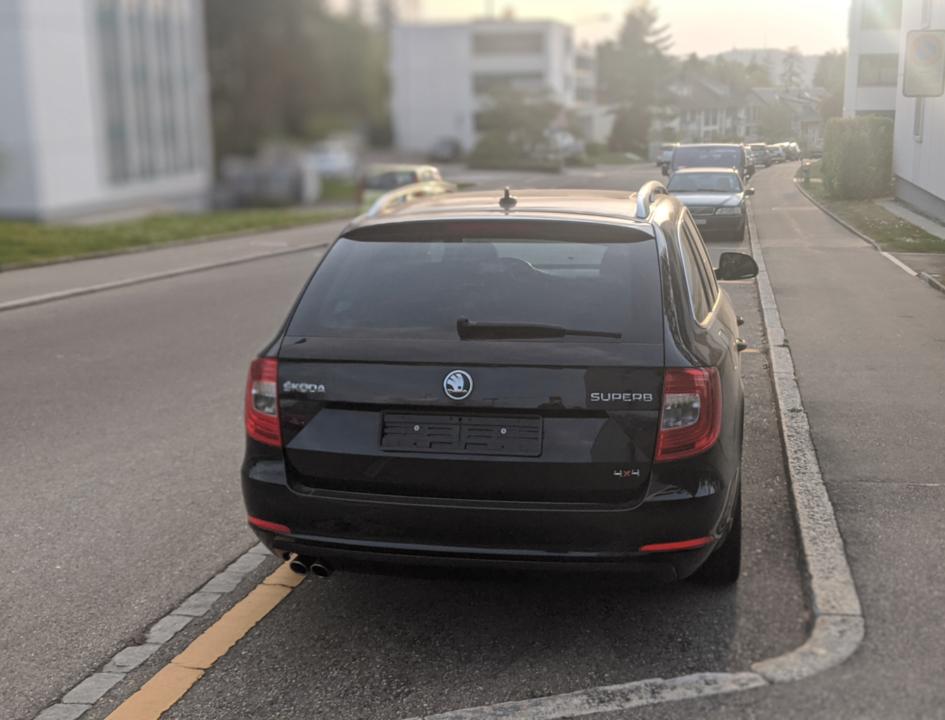}
\end{tabular}
}
\vspace{0.5mm}
\caption{\small{Visual results obtained using the proposed PyNET model (top) and the Google Pixel Camera application (bottom). Both approaches were applied to the same photos, Pixel Camera was additionally using dual-pixel auto-focus hardware. Best zoomed on screen.}}
\vspace{-1.2mm}
\label{fig:pixel_camera}
\end{figure*}

In this section, we compare our solution to the current state-of-the-art architectures that were designed and tuned specifically for the considered problem. The following models are used in the next experiments:
\smallskip

\noindent 1. Zheng~\etal\cite{ignatov2019aim}: a multiscale predictive filter CNN with the gate fusion and the constrained predictive filter blocks;

\noindent 2. Dutta~\etal\cite{ignatov2019aim}: a multiscale CNN trained to combine several images blurred with different Gaussian filters;

\noindent 3. Purohit~\etal\cite{purohit2019depth}: a modified U-Net based architecture;

\noindent 4. Xiong~\etal\cite{ignatov2019aim}: an ensemble of five U-Net based models with residual attention mechanism;

\noindent 5. Yang~\etal\cite{ignatov2019aim}: two stacked bokehNet CNN models with additional memory blocks.

\smallskip
All models were trained on the same EBB! dataset. For each network, we computed PSRN, SSIM~\cite{wang2003multiscale} and LPIPS~\cite{zhang2018unreasonable} metrics on the test subset of the dataset. Besides that, we conducted a user study involving several hundreds of participants (using Amazon's MTurk platform) evaluating the visual results of all considered methods. The users were asked to rate the full resolution bokeh images rendered with each method by selecting one of the five quality levels (0 - almost identical, 4 - mostly different) in comparison with the corresponding target Canon photos. The expressed preferences were then averaged per each approach to obtain the final Mean Opinion Scores (MOS). The numerical results of these experiments are presented in Table~\ref{tab:results}, sample visual results of all methods are shown in Fig.~\ref{fig:method_comparison}.

The first thing that should be noted is that, contrary to our expectations, neither of the numerical metrics work on the considered bokeh effect rendering problem. In particular, the best LPIPS results were obtained by Purohit~\etal, however, this solution is producing images with very strong artifacts in the bokeh area. The best PSNR and SSIM results were achieved by Xiong~\etal, though this approach is unfortunately also blurring in-focus objects and has significant issues with rendering the boundaries between the in and out of focus areas. Therefore, we had to rely on the actual visual results produced by all methods and the user study. According to both criteria, the PyNET model was able to significantly outperform the rest of the solutions. Compared to the second best approach, it is rendering the borders of in-focus objects considerably more accurately, especially in complex image scenes, and produces a slightly stronger bokeh effect. We should also highlight that the PyNET model is taking the input images downscaled by a factor of 2, thus also learning to perform image upscaling. Though training on the original resolution images might lead to slightly better numerical results, this will also significantly increase the training and inference times, while we did not observe any notable difference in the produced visual results.

\subsection{The Latency on Mobile Devices}
While the deployment of the proposed solution on mobile devices is not the main focus of this paper, we still decided to perform the corresponding experiment to show the feasibility of running the PyNET on smartphones. For this, we chose four mainstream high-end mobile chipsets: the Qualcomm Snapdragon 855, the Samsung Exynos 9820, the HiSilicon Kirin 980 all released at the end of 2018, and the Qualcomm Snapdragon 845 presented one year earlier. To run the pre-trained PyNET model, it was first converted to the TensorFlow Lite format~\cite{TensorFlowLite2018}, and then launched on the above chipsets using the corresponding Android library. TensorFlow Lite GPU delegate~\cite{lee2019device} was chosen to accelerate the model on smartphone GPUs: since it is independent of the vendors' NN drivers, it can run on any mobile device with OpenCL support. We used the PRO mode of the publicly available AI Benchmark application\footnote{\,\url{http://ai-benchmark.com}} that allows to run any custom TFLite model with various acceleration options, and adhered to the same settings as in~\cite{ignatov2019ai} to measure the runtime. The obtained results are presented in Table~\ref{tab:latency}.


When running the model without any modifications, it took almost 13 and 17 seconds on the Kirin 980 and the Exynos 9820 SoCs, respectively, while on both Qualcomm chipsets the model failed with an OOM error. The reason for this is that instance normalization layers are still not supported adequately by the TensorFlow Lite framework, and the corresponding ops are computed on the CPU, thus increasing the inference time and memory consumption dramatically due to additional CPU-GPU synchronization. After disabling instance normalization, the model was able to process one 1024$\times$1536 pixel photo in less than 5.5 seconds on all chipsets. The results on the current flagship mobile SoCs would be even better and should generally not exceed 3 seconds per image, making the solution ready to be deployed on mobile devices. We should also note that with several small modifications of the current architecture, one can achieve huge speed-up at the cost of a slightly degraded accuracy, however this topic is out of scope of this paper.

\begin{table*}
\centering
\resizebox{1.8\columnwidth}{!}
{
\begin{tabular}{l|c|c|c|c}
Mobile Chipset & Snapdragon 855 & \hspace{0.6mm} Exynos 9820 \hspace{0.6mm} & \hspace{2.6mm} Kirin 980 \hspace{2.6mm} & Snapdragon 845 \\
\hline
GPU Model & Adreno 640 & Mali-G76 & Mali-G76 & Adreno 630 \\
GPU Cores & - & 12 & 10 & - \\
GPU Frequency &  600 MHz & 600 MHz & 750 MHz & 710 MHz \\
\hline
\hline
PyNET with Instance Norm, seconds\, & - & 17.3 & 13.4 & - \\
PyNET w/o Instance Norm, seconds\, & 4.1 & 3.6 & 4.3 & 5.4 \\
\end{tabular}
}
\vspace{1.2mm}
\caption{\small{Average processing time for images of resolution 1024$\times$1536 pixels obtained on several mainstream high-end mobile SoCs. In each case, the model was running directly on the corresponding GPU using OpenCL-based TensorFlow Lite GPU delegate~\cite{lee2019device}.
\label{tab:latency}}}
\end{table*}

\vspace{1.6mm}

\subsection{Comparison to the Google Pixel Camera}

In the last section of the paper, we compare bokeh images rendered using the PyNET model with the photos obtained using the Google Pixel Camera capturing images in the ``Portrait Mode''~\cite{wadhwa2018synthetic}. The latter solution is utilizing camera's dual-pixel auto-focus system for an accurate estimation of the depth map that is then used to render bokeh effect on the photos. To compare these two approaches, we captured a large number of photos with the Google Pixel Camera installed on the Google Pixel 2 smartphone: it was saving both the original wide depth-of-field image and the rendered bokeh photo, the first image from each pair was then processed with the PyNET model trained on the EBB! dataset. The visual results produced by both solutions are presented in Figure~\ref{fig:pixel_camera}.

When considering the close-up photos (Fig.~\ref{fig:pixel_camera}, 1st row), PyNET is able to render images with much stronger and more pleasing bokeh effect than the Google Pixel Camera. For more distant objects, Pixel Camera is often producing better results due to the more accurate estimation of the distance to the objects and their boundaries (2nd row). When shooting more complex distant scenes, both the PyNET and the Google Pixel camera are often unable to provide good results (3rd row), though the latter one is producing unnaturally blurred photos more often. The conducted user study considering all collected photos demonstrated that, when rescaled to the same resolution, the images produced by the PyNET were preferred over the results of the Google Pixel camera in 53.7\% of the cases. We should note that the above comparison is not really fair since the Google Pixel Camera is able to process much larger images, though is also using additional camera hardware and was specifically tuned for the considered Pixel phone. However, the results indicate that the PyNET model is potentially able to outperform the existing commercial solutions when modifying its architecture for processing larger images, especially if using an accurate depth map that can be estimated using data from multiple cameras installed in the majority of modern mobile devices. We leave this challenge for the future work that can be done taking into account the above conditions.

\section{Conclusions}

In this paper, we presented a novel approach for the realistic bokeh effect rendering task. To simulate natural camera bokeh effect, we proposed to learn it directly from the real photos captured by a high-end DSLR camera. For this, we first collected a large-scale EBB! dataset consisting of aligned wide and shallow depth-of-field image pairs captured using the Canon 7D camera and 50mm f/1.8 fast lens. Then, we trained the proposed PyNET-based solution on the considered data and achieved significant qualitative improvements over the existing deep learning solutions tuned for this problem as confirmed by the conducted user study. We demonstrated that the proposed solution requires less than 5 seconds for processing one 1024$\times$1536 pixel image on all mobile high-end chipsets, and additionally shown that the rendered bokeh photos are comparable to the results of the Google Pixel Camera application when comparing both methods on images of the same resolution. We conclude that the results show the viability of our approach of an end-to-end bokeh effect rendering model, though further study is required to make it process high-resolution photos under the constraints imposed by mobile hardware.

\section*{Acknowledgements}
This work was partly supported by ETH Zurich General Fund (OK) and by Amazon AWS and Nvidia grants.

{\small
\bibliographystyle{ieee_fullname}

}

\end{document}